\documentclass{article}

\usepackage[final,nonatbib]{neurips_2021} 

\usepackage[utf8]{inputenc} 
\usepackage[T1]{fontenc}    
\usepackage{hyperref}       
\usepackage{url}            
\usepackage{booktabs}       
\usepackage{amsfonts}       
\usepackage{nicefrac}       
\usepackage{microtype}      
\usepackage{xcolor}         
\usepackage{bm}
\usepackage{amsmath}
\usepackage{booktabs,arydshln}
\makeatletter
\def\adl@drawiv#1#2#3{%
        \hskip.5\tabcolsep
        \xleaders#3{#2.5\@tempdimb #1{1}#2.5\@tempdimb}%
                #2\z@ plus1fil minus1fil\relax
        \hskip.5\tabcolsep}
\newcommand{\cdashlinelr}[1]{%
  \noalign{\vskip\aboverulesep
           \global\let\@dashdrawstore\adl@draw
           \global\let\adl@draw\adl@drawiv}
  \cdashline{#1}
  \noalign{\global\let\adl@draw\@dashdrawstore
           \vskip\belowrulesep}}
\makeatother
\usepackage[most]{tcolorbox}

\usepackage{graphicx} 
\usepackage{wrapfig}
\usepackage{multirow}
\usepackage{tabularx}
\usepackage{caption}
\usepackage{algorithm,algorithmicx,algpseudocode}
\usepackage{amssymb}

\newtheorem{observation}{Observation}
\usepackage[maxfloats=256]{morefloats}
\usepackage[textsize=scriptsize]{todonotes}

\title{
PARP: Prune, Adjust and Re-Prune\\
for Self-Supervised Speech Recognition
}

\author{%
  Cheng-I Jeff Lai\textsuperscript{1}, \hspace{2.25mm} 
  Yang Zhang\textsuperscript{2}\thanks{Equal contribution.}, \hspace{2.25mm} 
  Alexander H. Liu\textsuperscript{1}\footnotemark[1], \hspace{2.25mm} 
  Shiyu Chang\textsuperscript{2, 4}\footnotemark[1]\\
  \textbf{Yi-Lun Liao\textsuperscript{1}}, \hspace{1mm} 
  \textbf{Yung-Sung Chuang\textsuperscript{1, 3}}, \hspace{1mm} 
  \textbf{Kaizhi Qian\textsuperscript{2}}, \hspace{1mm} 
  \textbf{Sameer Khurana\textsuperscript{1}}\\
  \textbf{David Cox\textsuperscript{2}}, \hspace{3mm} 
  \textbf{James Glass\textsuperscript{1}}\\
  {\textsuperscript{1}MIT CSAIL, \textsuperscript{2}MIT-IBM Watson AI Lab, \textsuperscript{3}National Taiwan University, \textsuperscript{4}UC Santa Barbara}
  \vspace{2mm}
  \\
  \small{\texttt{clai24@mit.edu}}
}

\begin{document}

\maketitle

\vspace{-5mm}
\begin{abstract}
\vspace{-2mm}
Self-supervised speech representation learning (speech SSL) has demonstrated the benefit of scale in learning rich representations for Automatic Speech Recognition (ASR) with limited paired data, such as wav2vec 2.0. 
We investigate the existence of sparse subnetworks in pre-trained speech SSL models that achieve even better low-resource ASR results. 
However, directly applying widely adopted pruning methods such as the Lottery Ticket Hypothesis (LTH) is suboptimal in the computational cost needed.
Moreover, we show that the discovered subnetworks yield minimal performance gain compared to the original dense network. 

We present Prune-Adjust-Re-Prune ({\tt PARP}), which discovers and finetunes subnetworks for much better performance, while only requiring a \textit{single} downstream ASR finetuning run.
{\tt PARP} is inspired by our surprising observation that subnetworks pruned for pre-training tasks need merely a slight adjustment to achieve a sizeable performance boost in downstream ASR tasks.
Extensive experiments on low-resource ASR verify (1) sparse subnetworks exist in mono-lingual/multi-lingual pre-trained speech SSL, and (2) the computational advantage and performance gain of {\tt PARP} over baseline pruning methods.

In particular, on the 10min Librispeech split without LM decoding, {\tt PARP} discovers subnetworks from wav2vec 2.0 with an absolute 10.9\%/12.6\% WER decrease compared to the full model.  
We further demonstrate the effectiveness of {\tt PARP} via: cross-lingual pruning without any phone recognition degradation, the discovery of a multi-lingual subnetwork for 10 spoken languages in 1 finetuning run, and its applicability to pre-trained BERT/XLNet for natural language tasks\footnote[1]{Project webpage: \url{https://people.csail.mit.edu/clai24/parp/}}.

\end{abstract}

\vspace{-4mm}
\section{Introduction}
\label{sec:intro}
\vspace{-2mm}
For many low-resource spoken languages in the world, collecting large-scale transcribed corpora is very costly and sometimes infeasible.
Inspired by efforts such as the IARPA BABEL program, Automatic Speech Recognition (ASR) trained without sufficient transcribed speech data has been a critical yet challenging research agenda in speech processing~\cite{cui2013developing,cui2014improving,gales2014speech,cui2015multilingual,cho2018multilingual}. 
Recently, Self-Supervised Speech Representation Learning (speech SSL) has emerged as a promising pathway toward solving low-resource ASR~\cite{oord2018representation,chung2019unsupervised,wang2020unsupervised,baevski2020wav2vec,conneau2020unsupervised,zhang2020pushing,hsu2021hubert,chung2021w2v}.
Speech SSL involves pre-training a speech representation module on large-scale \emph{unlabelled} data with a self-supervised learning objective, followed by finetuning on a small amount of supervised transcriptions.
Many recent studies have demonstrated the empirical successes of speech SSL on low-resource English and multi-lingual ASR, matching systems trained on fully-supervised settings~\cite{baevski2020wav2vec,conneau2020unsupervised,zhang2020pushing,baevski2021unsupervised,zhang2021bigssl}. 
Prior research attempts, however, focus on pre-training objectives~\cite{oord2018representation,chung2019unsupervised,wang2020unsupervised,liu2020non,jiang2020speech,liu2020mockingjay,ling2020decoar,liu2021tera,hsu2021hubert,chorowski2021aligned,chung2021w2v,chen2021unispeech,zhu2021wav2vec}, scaling up speech representation modules~\cite{baevski2019vq,baevski2020wav2vec,hsu2021hubertlarge}, pre-training data selections~\cite{wang2021unispeech,hsu2021robust,wang2021unispeechscale,wang2021wav2vec,meng2021don}, or applications of pre-trained speech representations~\cite{chung2018unsupervised,lai2019contrastive,riviere2020unsupervised,chung2020splat,lai2021semi,conneau2020unsupervised,maekaku2021speech,yang2021superb,lakhotia2021generative,xu2021simple,wiesner2021injecting,gao2021zero,baevski2021unsupervised,polyak2021speech,kharitonov2021text,lee2021direct,ao2021speecht5,huang2021s3prl,tseng2021mandarin,chang2021exploration,cooper2021generalization,chen2021speech}.
In this work, we aim to develop an orthogonal approach that is complementary to these existing speech SSL studies, that achieves 1) lower architectural complexity and 2) higher performance (lower WER) under the same low-resource ASR settings. 

Neural network pruning~\cite{lecun1990optimal,hassibi1993second,han2015learning,li2016pruning}, as well as the more recently proposed Lottery Ticket Hypothesis (LTH)~\cite{frankle2018lottery}, provide a potential solution that accomplishes both objectives. 
According to LTH, there exists sparse subnetworks that can achieve the same or \textit{even better} accuracy than the original dense network. 
Such phenomena have been successfully observed in various domains: Natural Language Processing (NLP)~\cite{yu2019playing,chen2020lottery,prasanna2020bert,movva2020dissecting}, Computer Vision (CV)~\cite{chen2020lotteryCV,girish2020lottery}, and many others. 
All finding sparse subnetworks with comparable or better performance than the dense network. 
Given the lack of similar studies on pruning self-supervised ASR, we intend to fill this gap by finding sparse subnetworks \textit{within a pre-trained} speech SSL that can achieve superior performance to the full pre-trained model on downstream ASR tasks.

\begin{wrapfigure}{r}{0.3\textwidth}
\vspace{-1.5em}\centering\strut
\includegraphics[width=0.3\textwidth]{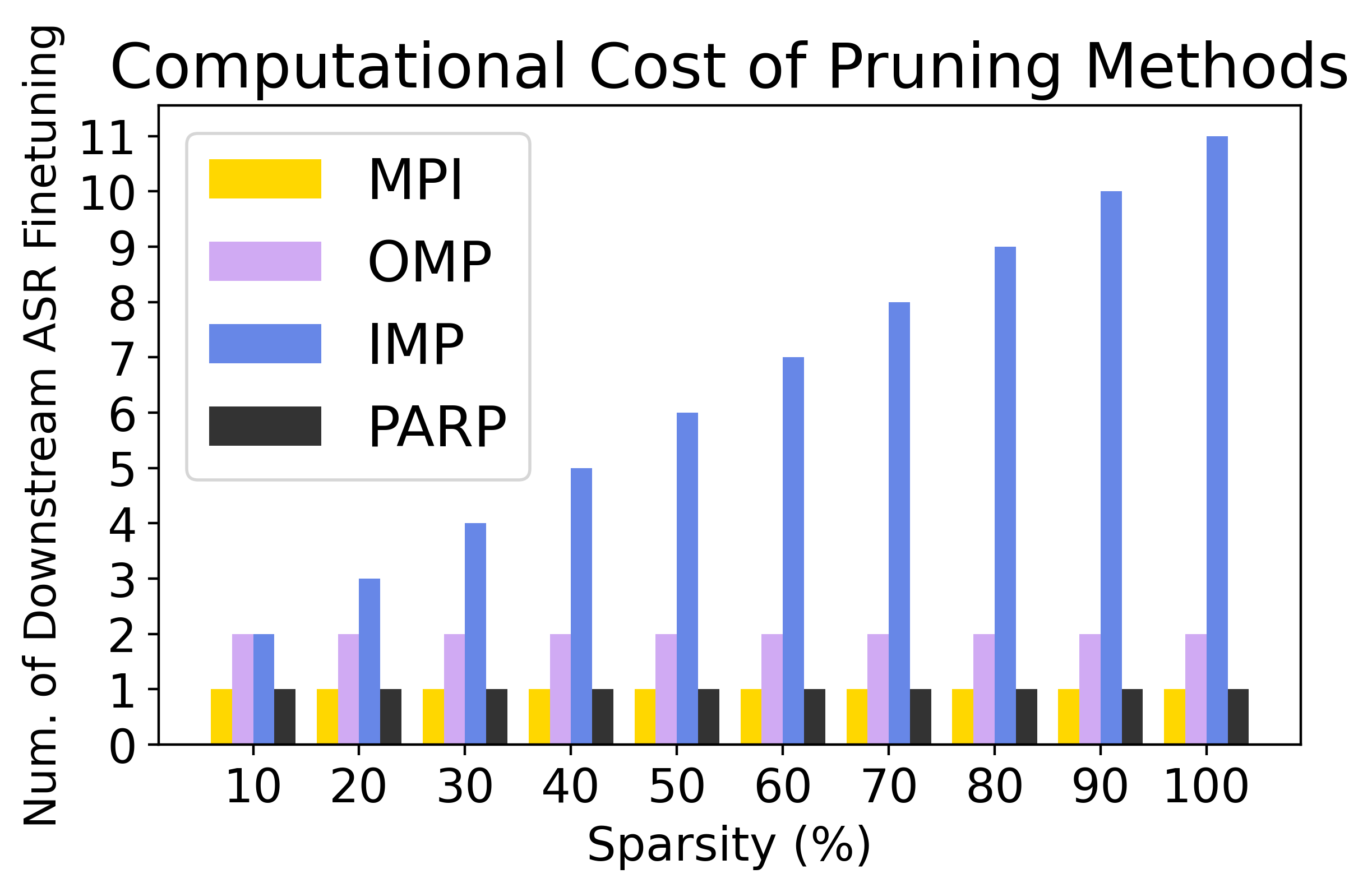}
\vspace{-1em}
\caption{
Number of ASR finetuning iterations needed (y-axis) versus target sparsities (x-axis) for \textit{each} downstream task/language.
Cross-referencing Figure~\ref{fig:main_ler_results} indicates that {\tt IMP} requires linearly more compute to match the performance (either sparsity/WER) of {\tt PARP}.}
\vspace{-1.5em}
\label{fig:teaser}
\end{wrapfigure}

However, directly applying widely-adopted pruning methods, such as One-Shot Magnitude Pruning ({\tt OMP}) and Iterative Magnitude Pruning ({\tt IMP})~\cite{han2015learning,frankle2018lottery}, to pre-trained speech SSL suffers from two challenges. 
First, adopting these methods in the conventional pruning framework is extremely time-consuming for SOTA speech SSL models.  
{\tt OMP} and {\tt IMP} involve more than one round of finetuning on downstream tasks (c.f. Figure~\ref{fig:teaser}), and finetuning for ASR is time-consuming and computationally demanding\footnote{Standard wav2vec 2.0 finetuning setup~\cite{baevski2020wav2vec} on any Librispeech/Libri-light splits requires at least 50$\sim$100 V100 hours, which is more than 50 times the computation cost for finetuning a pre-trained BERT on GLUE~\cite{wang2018glue}.}.
The second challenge is that we do not observe \textit{any} performance improvement of the subnetworks over the original dense network with {\tt OMP} or {\tt IMP}.
Figure~\ref{fig:main_ler_results} shows the WER under low-resource scenarios of the subnetworks identified by {\tt OMP} (purple line) and {\tt IMP} (blue dashed line) at different sparsity levels. 
None of the sparsity levels achieves a visible drop in WER compared to the zero sparsity case, corresponding to the original dense network. 
These two challenges have prompted us to ask -- do there exist sparse subnetworks within pre-trained speech SSL with improved performance on low-resource ASR? How can we discover them efficiently in a \textit{single} downstream finetuning run?

We propose a magnitude-based unstructured pruning method~\cite{gale2019state,blalock2020state}, termed Prune-Adjust-Re-Prune ({\tt PARP}), for discovering sparse subnetworks within pre-trained speech SSL. 
{\tt PARP} consists of the following two steps:
\begin{enumerate}
    \vspace{-2.5mm}
    \item Directly prune the SSL pre-trained model at target sparsity, and obtain an initial subnetwork and an initial pruning mask. 
    \vspace{-1.25mm}
    \item Finetune the initial subnetwork on target downstream task/language. During finetuning, zero out the pruned weights specified by the pruning mask, but allow the weights be updated by gradient descent during backpropogation. After a few number of model updates, re-prune the updated subnetwork at target sparsity again.
    \vspace{-2.5mm}
\end{enumerate}
Step 1 provides an initial subnetwork that is agnostic to the downstream task, and Step 2 makes learnable adjustments by reviving pruned out weights. 
A formal and generalized description and its extension are introduced in Section~\ref{sec:method}.
Different from pruning methods in~\cite{han2015learning,frankle2018lottery}, {\tt PARP} allows pruned-out weights to be revived during finetuning. 
Although such a high-level idea was introduced in~\cite{guo2016dynamic}, we provide an alternative insight: despite its flexibility, Step 2 only makes \textbf{minimal adjustment} to the initial subnetwork, and obtaining a good initial subnetwork in Step 1 is the key. 
We empirically show in Section~\ref{sec:method} that \textit{any} task-agnostic subnetwork surprisingly provides a good basis for Step 2, suggesting that the initial subnetwork can be cheaply obtained either from a readily available task/language or directly pruning the pre-trained SSL model itself. 
In addition, this observation allows us to perform cross-lingual pruning (mask transfer) experiments, where the initial subnetwork is obtained via a different language other than the target language. 


\textbf{Our Contributions.} We conduct extensive {\tt PARP} and baseline ({\tt OMP} and {\tt IMP}) pruning experiments on low-resource ASR with mono-lingual (pre-trained wav2vec 2.0~\cite{baevski2020wav2vec}) and cross-lingual (pre-trained XLSR-53~\cite{conneau2020unsupervised}) transfer. 
{\tt PARP} finds significantly superior speech SSL subnetworks for low-resource ASR, while only requiring a single pass of downstream ASR finetuning. 
Due to its simplicity, {\tt PARP} adds minimal computation overhead to existing SSL downstream finetuning.
    \begin{itemize}
    \vspace{-2mm}
        \item We show that sparse subnetworks exist in pre-trained speech SSL when finetuned for low-resource ASR. 
        In addition, {\tt PARP} achieves superior results to {\tt OMP} and {\tt IMP} across all sparsities, amount of finetuning supervision, pre-trained model scale, and downstream spoken languages.
        Specifically, on Librispeech 10min without LM decoding, {\tt PARP} discovers subnetworks from wav2vec 2.0 with an absolute 10.9\%/12.6\% WER decrease compared to the full model, without modifying the finetuning hyper-parameters or objective (Section~\ref{subsec:exp_main_result}). 
        \item Ablation studies on demonstrating the importance of {\tt PARP}'s initial subnetwork (Section~\ref{subsec:exp_ablation}). 
        \item {\tt PARP} minimizes phone recognition error increases in cross-lingual mask transfer, where a subnetwork pruned for ASR in one spoken language is adapted for ASR in another language (Section~\ref{subsec:mask_transfer}). 
        {\tt PARP} can also be applied to efficient multi-lingual subnetwork discovery for 10 spoken languages (Section~\ref{subsec:joint}). 
        \item Last but not least, we demonstrate {\tt PARP}'s effectiveness on pre-trained BERT/XLNet, mitigating the cross-task performance degradation reported in BERT-Ticket~\cite{chen2020lottery} (Section~\ref{subsec:exp_bert}).
    \vspace{-2mm}
    \end{itemize}

\vspace{-3mm}
\textbf{Significance.} Findings of this work not only complement and advance current and future speech SSL for low-resource ASR, but also provide new insights for the rich body of pruning work.

\vspace{-2mm}
\section{Preliminaries}
\label{sec:prelim}

\vspace{-2mm}
\subsection{Problem Formulation}
\label{subsec:prelim_formulation}
    \vspace{-2mm}
    Consider the low-resource ASR problem, where there is only a small transcribed training set $(x, y) \in \mathcal{D}_l$. 
    Here $x$ represents input audio, and $y$ represents output transcription. 
    Subscript $l \in \{1, 2, \cdots\}$ represents the downstream spoken language identity.
    Because of the small dataset size, empirical risk minimization generally does not yield good results.
    Speech SSL instead assumes there is a much larger unannotated dataset $x \in \mathcal{D}_0$. 
    SSL pre-trains a neural network $f(x; \theta)$, where $\theta \in \mathcal{R}^d$ represents the network parameters and $d$ represents the number of parameters, on some self-supervised objective, and obtains the pre-trained weights $\theta_{0}$. 
    $f(x; \theta_{0})$ is then finetuned on downstream ASR tasks specified by a downstream loss $\mathcal{L}_l(\theta)$, such as CTC, and evaluated on target dataset $\mathcal{D}_l$.
    
    Our goal is to discover a subnetwork that minimizes downstream ASR WER on $\mathcal{D}_l$. 
    Formally, denote $m \in \{0, 1\}^d$, as a binary pruning mask for the pre-trained weights $\theta_{0}$, and $\theta^l$ as the finetuned weights on $\mathcal{D}_l$. 
    The ideal pruning method should learn $(m, \theta^l)$, such that the subnetwork $f(x; m \odot \theta^l)$ (where $\odot$ is element-wise product) achieves minimal finetuning $\mathcal{L}_l(\theta)$ loss on $\mathcal{D}_l$.
    

\vspace{-2mm}
\subsection{Pruning Targets and Settings}
\label{subsec:prelim_targets}
    \vspace{-2mm}
    We adopted pre-trained speech SSL {\tt wav2vec2} and {\tt xlsr} for the pre-trained initialization $\theta_{0}$. 
    
    \textbf{wav2vec 2.0} We took wav2vec 2.0 base ({\tt wav2vec2-base}) and large ({\tt wav2vec2-large}) pre-trained on Librispeech 960 hours~\cite{baevski2020wav2vec}.
    During finetuning, a task specific linear layer is added on top of {\tt wav2vec2} and jointly finetuned with CTC loss. 
    More details can be found in Appendix~\ref{app:model_details}.

    
    \textbf{XLSR-53} ({\tt xlsr})
    shares the same architecture, pre-training and finetuning objectives as {\tt wav2vec2-large}. 
    {\tt xlsr} is pre-trained on 53 languages sampled from CommonVoice, BABEL, and Multilingual LibriSpeech, totaling for 56k hours of multi-lingual speech data.

    We consider three settings where {\tt wav2vec2} and {\tt xlsr} are used as the basis for low-resource ASR:
    
    \textbf{LSR: Low-Resource English ASR.}
    Mono-lingual pre-training and finetuning -- an English pre-trained speech SSL such as {\tt wav2vec2} is finetuned for low-resource English ASR. 
    
    \textbf{H2L: High-to-Low Resource Transfer for Multi-lingual ASR.}
    Mono-lingual pre-training and multi-lingual finetuning -- a speech SSL pre-trained on a high-resource language such as English is finetuned for low-resource multi-lingual ASR. 
    
    \textbf{CSR: Cross-lingual Transfer for Multi-lingual ASR.}
    Multi-lingual pre-training and finetuning -- a cross-lingual pretrained speech SSL such as {\tt xlsr} is finetuned for low-resource multi-lingual ASR. 
    
\vspace{-2mm}
\subsection{Subnetwork Discovery in Pre-trained SSL}
\label{subsec:prelim_lth}
    \vspace{-2mm}
    One obvious solution to the aforementioned problem in Section~\ref{subsec:prelim_formulation} is to directly apply pruning with rewinding to $\theta_{0}$, which has been successfully applied to pre-trained BERT~\cite{chen2020lottery} and SimCLR~\cite{chen2020lotteryCV}.
    All pruning methods, including our proposed {\tt PARP}, are based on Unstructured Magnitude Pruning ({\tt UMP})~\cite{frankle2018lottery,gale2019state}, where weights of the lowest magnitudes are pruned out regardless of the network structure to meet the target sparsity level. 
    We introduce four pruning baselines below, and we also provide results with Random Pruning ({\tt RP})~\cite{frankle2018lottery,gale2019state,chen2020lottery}, where weights in $\theta_{0}$ are randomly eliminated. 
    
    
    
    \textbf{Task-Aware Subnetwork Discovery} is pruning with target dataset $D_{l}$ seen in advance, including One-Shot Magnitude Pruning ({\tt OMP}) and Iterative Magnitude Pruning ({\tt IMP}).
    {\tt OMP} is summarized as: 
    \vspace{-2mm}
    \begin{enumerate}
        \vspace{0mm}
        \item Finetune pretrained weights $\theta_{0}$ on target dataset $\mathcal{D}_l$ to get the finetuned weights $\theta^l$.
        \vspace{-.5mm}
        \item Apply {\tt UMP} on $\theta^l$ and retrieve pruning mask $m$.
        \vspace{-2mm}
    \end{enumerate}
    {\tt IMP} breaks down the above subnetwork discovery phase into multiple iterations -- in our case multiple downstream ASR finetunings. 
    Each iteration itself is an {\tt OMP} with a fraction of the target sparsity pruned.
    We follow the {\tt IMP} implementation described in BERT-Ticket~\cite{chen2020lottery}, where each iteration prunes out 10\% of the \textit{remaining} weights.
    The main bottleneck for {\tt OMP} and {\tt IMP} is the computational cost, since multiple rounds of finetunings are required for subnetwork discovery. 
    
    \textbf{Task-Agnostic Subnetwork Discovery} refers to pruning without having seen $D_{l}$ nor $l$ in advance. 
    One instance is applying {\tt UMP} directly on $\theta_{0}$ without any downstream finetuning to retrieve $m$, referred to as Magnitude Pruning at Pre-trained Initailizations ({\tt MPI}). 
    Another case is pruning weights finetuned for a different language $t$, \textit{i.e.} applying {\tt UMP} on $\theta^t$ for the target language $l$; in our study, we refer to this as cross-lingual mask transfer.
    While these approaches do not require target task finetuning, the discovered subnetworks generally have worse performance than those from {\tt OMP} or {\tt IMP}.
    
    The above methods are only for subnetwork discovery via applying pruning mask $m$ on $\theta_{0}$.
    The discovered subnetwork $f(x; m \odot \theta_{0})$ needs another downstream finetuning to recover the pruning loss\footnote{This step is referred to as subnetwork finetuning/re-training in the pruning literature~\cite{liu2018rethinking,renda2020comparing,blalock2020state}.}, \textit{i.e.} finetune $f(x; m \odot \theta_{0})$ on $D_{l}$.
    
    

\vspace{-2mm}
\section{Method}
\label{sec:method}
\vspace{-2mm}

In this section, we highlight our proposed pruning method, {\tt PARP} (Section~\ref{subsec:exp_alg}), its underlying intuition (Section~\ref{subsec:exp_intuition}), and an extension termed {\tt PARP-P} (Section~\ref{subsec:exp_parpp}).
\vspace{-2mm}
\subsection{Algorithm}
\label{subsec:exp_alg}
\vspace{-2mm}
We formally describe {\tt PARP} with the notations from Section~\ref{sec:prelim}. 
A visual overview of {\tt PARP} is Figure~\ref{fig:overview}. 
\vspace{-5mm}
{\begin{algorithm}[H]
    \small
    \caption{Prune-Adjust-Re-Prune (PARP) to target sparsity $s$}
    \begin{algorithmic}[1]
        \State Assume there are $N$ model updates in target task/language $l$'s downstream finetuning. 
        \State Take a pre-trained SSL $f(x; \theta_{0})$ model. 
        Apply task-agnostic subnetwork discovery, such as {\tt MPI}\footnotemark, at target sparsity $s$ to obtain initial subnetwork $f(x; m_{0} \odot \theta_{0})$. Set $m = m_{0}$ and variable $n_{1}=0$ .
        \Repeat
        \State Zero-out masked-out weights in $\theta_{n1}$ given by $m$. Lift up $m$ such that whole $\theta_{n1}$ is updatable. 
        \State Train $f(x; \theta_{n1})$ for $n$ model updates and obtain $f(x; \theta_{n2})$. 
        \State Apply {\tt UMP} on $f(x; \theta_{n2})$ and adjust $m$ accordingly. The adjusted subnetwork is $f(x; m \odot \theta_{n2})$. Set variable $n_{1}=n_{2}$.
        \Until{total model updates reach $N$}. 
        \State Return finetuned subnetwork $f(x; m \odot \theta_{N})$.
    \end{algorithmic}
    \label{alg:parp}
\end{algorithm}}
\addtocounter{footnote}{0}
\footnotetext{By default, {\tt MPI} is used for obtaining the initial subnetwork for {\tt PARP} and {\tt PARP-P} unless specified otherwise.}
\vspace{-4mm}

Empirically, we found the choice of $n$ has little impact. 
In contrast to {\tt OMP}/{\tt IMP}/{\tt MPI}, {\tt PARP} allows the pruned-out weights to take gradient descent updates. 
A side benefit of {\tt PARP} is it jointly discovers and finetunes subnetwork in a single pass, instead of two or more in {\tt OMP} and {\tt IMP}.

\vspace{-2mm}
\subsection{Obtaining and Adjusting the Initial Subnetwork}
\label{subsec:exp_intuition}
\vspace{-2mm}
{\tt PARP} achieves superior or comparable pruning results as task-aware subnetwork discovery, while inducing similar computational cost as task-agnostic subnetwork discovery.
How does it get the best of both worlds? 
The key is the discovered subnetworks from task-aware and task-agnostic prunings have high, non-trivial overlaps in LSR, H2L, and CSR. 
We first define Intersection over Union ({\tt IOU}) for quantifying subnetworks' (represented by their pruning masks $m^{a}$ and $m^{b}$) similarity: 
\begin{equation}
    {\tt IOU}(m^{a}, m^{b})\triangleq \frac{|(m^{a} = 1) \cap (m^{b} = 1)|}{|(m^{a} = 1) \cup (m^{b} = 1)|} 
\label{eq:IOU}
\end{equation}

\begin{figure} [h]
\includegraphics[width=0.45\linewidth]{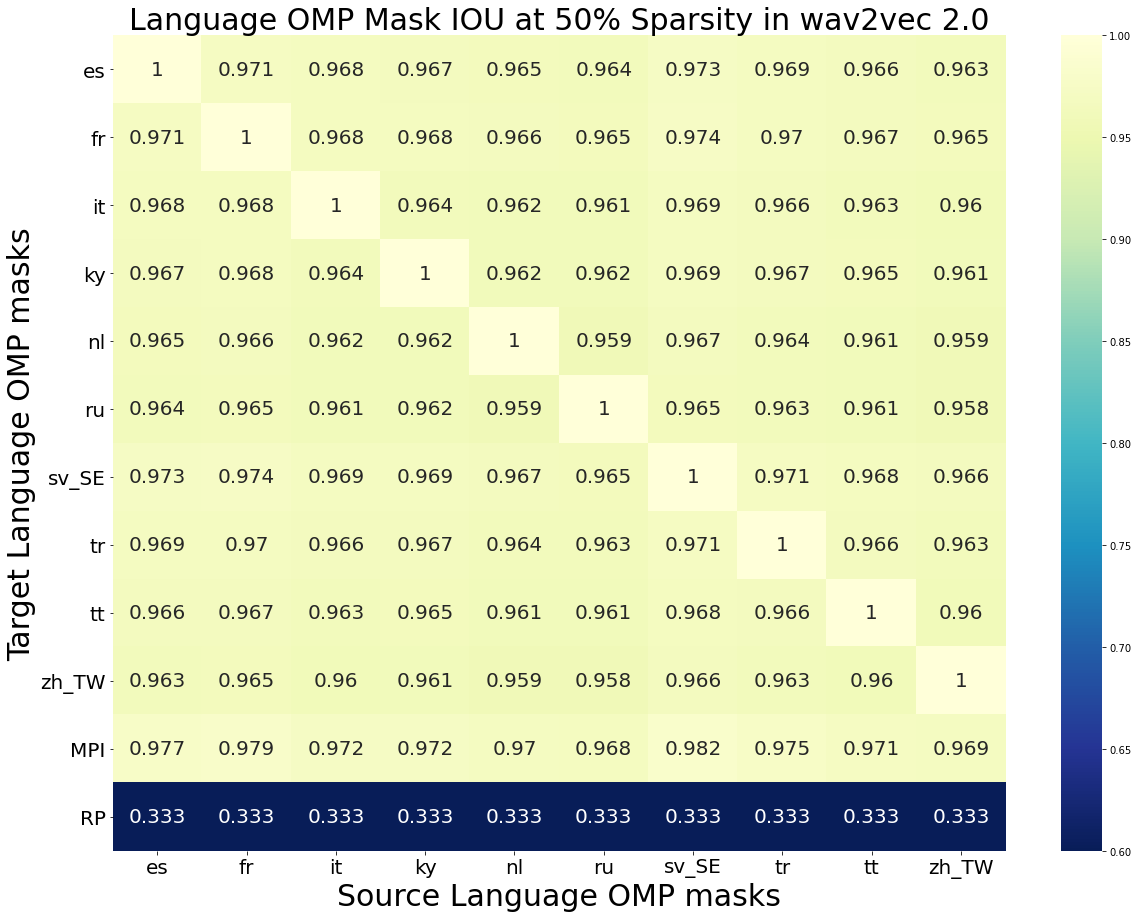}
\hspace{.5cm}
\includegraphics[width=0.45\linewidth]{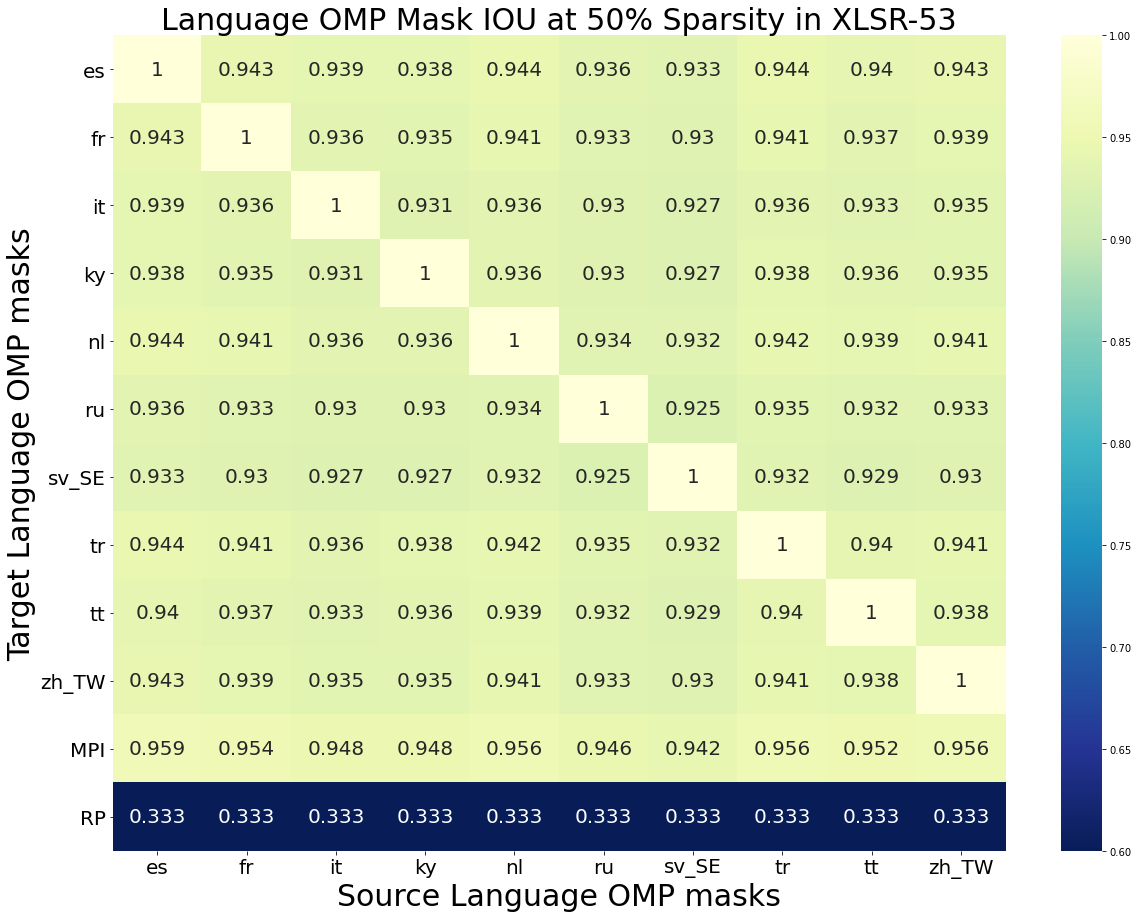}
\centering
\caption{{\tt IOU}s over all spoken language pairs' {\tt OMP} pruning masks on finetuned {\tt wav2vec2} and {\tt xlsr}. Second to last row is the {\tt IOU}s between {\tt OMP} masks and the {\tt MPI} masks from pre-trained {\tt wav2vec2} and {\tt xlsr}.
Here, we show the {\tt IOU}s at 50\% sparsity, and the rest can be found in Appendix~\ref{app:mask_overlap_matrices}. 
Surprisingly at any sparsities, there is a high, non-trivial (c.f. {\tt RP} in the last row), similarity (>90\%) between all spoken language {\tt OMP} masks, as well as with the {\tt MPI} masks.
Language IDs are in Appendix~\ref{app:exp_setup}.}
\label{fig:lan_mask_overlap_matrix}
\vspace{-6mm}
\end{figure}

Take H2L and CSR for instance, Figure~\ref{fig:lan_mask_overlap_matrix} visualizes language pairs' {\tt OMP} pruning mask {\tt IOU}s on {\tt wav2vec2} and {\tt xlsr}.  
Observe the high overlaps across all pairs, but also the high {\tt IOU}s with the {\tt MPI} masks (second to last row).
We generalize these observations to the following: 
\begin{tcolorbox}
\begin{observation}
\vspace{-2mm}
For any sparsity, any amount of finetuning supervision, any pre-training model scale, and any downstream spoken languages, the non-zero ASR pruning masks obtained from task-agnostic subnetwork discovery has high {\tt IOU}s with those obtained from task-aware subnetwork discovery.
\label{observation:similarity}
\vspace{-2mm}
\end{observation}
\end{tcolorbox}

Observation~\ref{observation:similarity} suggests that \textit{any} task-agnostic subnetwork could sufficiently be a good initial subnetwork in {\tt PARP} due to the high similarities.
In the same instance for H2L and CSR, we could either take {\tt MPI} on {\tt wav2vec2} and {\tt xlsr}, or take {\tt OMP} on a different spoken language as the initial subnetworks.
Similarly in LSR, we take {\tt MPI} on {\tt wav2vec2} as the initial subnetwork.
The underlying message is -- the initial subnetwork can be obtained cheaply, without target task finetuning. 

Now, because of the high similarity, the initial subnetwork (represented by its pruning mask $m_{0}$) needed merely a slight adjustment for the target downstream task. 
While there are techniques such as dynamic mask adjustment~\cite{guo2016dynamic}, important weights pruning~\cite{molchanov2019importance}, and deep rewiring~\cite{bellec2017deep}, we provide an even simpler alternative suited for our setting. 
Instead of permanently removing the masked-out weights from the computation graph, {\tt PARP} merely zeroes them out. 
Weights that are important for the downstream task (the ``important weights'') should emerge with gradient updates; those that are relatively irrelevant should decrease in magnitude, and thus be zero-outed at the end. 
Doing so circumvents the need of straight-through estimation or additional sparsity loss, see Table 1 of~\cite{sanh2020movement}.
\vspace{-2mm}
\subsection{PARP-Progressive ({\tt PARP-P})} 
\label{subsec:exp_parpp}
\vspace{-2mm}
An extension to {\tt PARP} is {\tt PARP-P}, where the second {\tt P} stands for Progressive. 
In {\tt PARP-P}, the initial subnetwork starts at a lower sparsity, and progressively prune up to the target sparsity $s$ in Step 2. 
The intuition is that despite Observation~\ref{observation:similarity}, \textit{not any} subnetwork can be a good initial subnetwork, such as those obtained from {\tt RP}, or those obtained at very high sparsities in {\tt MPI}/{\tt OMP}/{\tt IMP}.
We show later that {\tt PARP-P} is especially effective in higher sparsity regions, e.g. 90\% for LSR.
Note that {\tt PARP-P} has the same computational cost as {\tt PARP}, and the only difference is the initial starting sparsity in Step 1. 

\vspace{-2mm}

\vspace{-0mm}
\section{Experiments and Analysis}
\label{sec:exp}
\vspace{-2mm}

\subsection{Comparing {\tt PARP}, {\tt OMP}, and {\tt IMP} on LSR, H2L, and CSR}
\label{subsec:exp_main_result}
\begin{figure} [h]
\includegraphics[width=.9\linewidth]{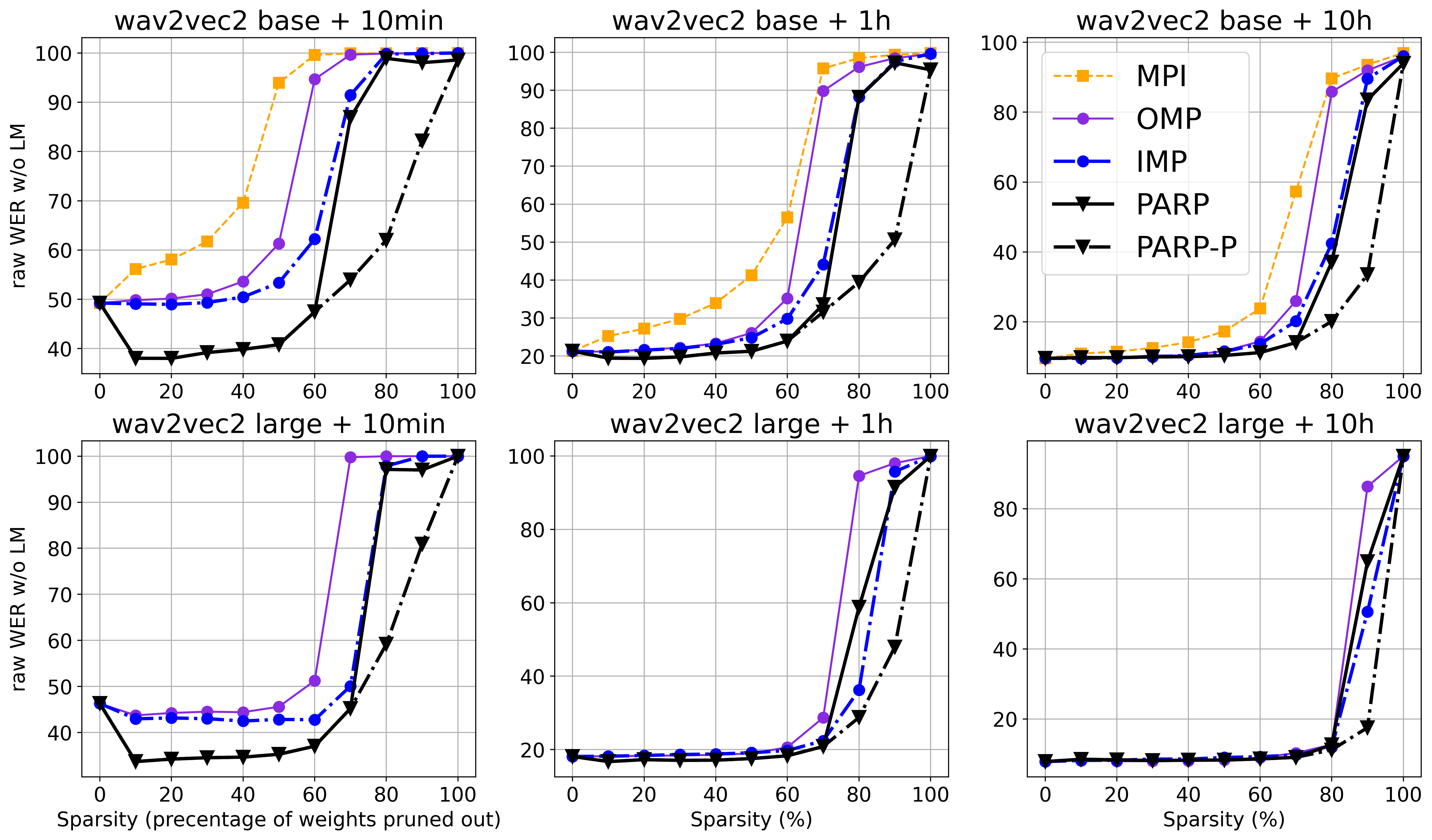}
\small
\centering
\vspace{-1mm}
\caption{Comparison of different pruning techniques on LSR ({\tt wav2vec2} with 10min/1h/10h Librispeech finetuning splits).
{\tt PARP} (black line) and {\tt PARP-P} (black dashed line) are especially effective under ultra-low data regime (e.g. 10min) and high-sparsity (70-100\%) regions.
}
\label{fig:main_ler_results}
\vspace{-4mm}
\end{figure}

\vspace{-2mm}
Our experimental setup can be found in Appendix~\ref{app:exp_setup}.
We first investigate the existence of sparse subnetworks in speech SSL. 
Figure~\ref{fig:main_ler_results} shows the pruning results on LSR. 
Observe that subnetworks discovered by {\tt PARP} and {\tt PARP-P} can achieve 60$\sim$80\% sparsities with minimal degradation to the full models.
The gap between {\tt PARP} and other pruning methods also widens as sparsities increase. 
For instance, Table~\ref{tab:wer_comparison1} compares {\tt PARP} and {\tt PARP-P} with {\tt OMP} and {\tt IMP} at 90\% sparsity, and {\tt PARP-P} has a 40\% absolute WER reduction.
In addition, observe the WER reduction with {\tt PARP} in the low sparsity regions on the 10min split in Figure~\ref{fig:main_ler_results}.
The same effect is not seen with {\tt OMP}, {\tt IMP}, nor {\tt MPI}.
Table~\ref{tab:wer_comparison2} compares the subnetworks discovered by {\tt PARP} with the full {\tt wav2vec2} and prior work on LSR under the same setting\footnote{We underscore again that LM decoding/self-training are not included to isolate the effect of pruning.}. 
Surprisingly, the discovered subnetwork attains an absolute 10.9\%/12.6\% WER reduction over the full {\tt wav2vec2-large}. 
We hypothesize that the performance gains are attributed to pruning out generic, unnecessary weights while preserving important weights, which facilitates training convergence. 
In other words, {\tt PARP} provides additional regularization effects to downstream finetuning. 
We also examined the effectiveness of {\tt IMP} with different rewinding starting points as studied in~\cite{frankle2020linear,renda2020comparing}, and found rewinding initializations bear minimal effect on downstream ASR. 
Full rewinding details are in Appendix~\ref{app:imp_rewinding}.
\vspace{-3mm}

\begin{figure}[ht]
    \begin{minipage}{0.47\linewidth}
        \small\centering
        \captionof{table}{WER comparison of pruning LSR: 
        {\tt wav2vec2-base} at 90\% sparsity with 10h finetuning on Librispeech without LM decoding.
        At 90\% sparsity, {\tt OMP}/{\tt IMP}/{\tt MPI} perform nearly as bad as {\tt RP}.
        sub-finetuning stands for subnetwork finetuning. 
        }
        \label{tab:wer_comparison1}
        \begin{center}
        \vspace{-3.5mm}
        \scalebox{0.9}{
        \begin{tabular}{lccc}
            \toprule
            \multirow{2}{*}{Method} & \# ASR & test & test \\
            & finetunings & clean & other \\
            \midrule 
            {\tt RP}  + sub-finetuning  & 1  & 94.5 & 96.4 \\
            {\tt MPI} + sub-finetuning & 1  & 93.6 & 96.1 \\
            {\tt OMP} + sub-finetuning & 2  & 92.0 & 95.3 \\
            {\tt IMP} + sub-finetuning & 10 & 89.6 & 93.9 \\
            \midrule
            {\tt PARP} ($90\%\rightarrow90\%$)   & 1 & 83.6 & 90.7 \\
            {\tt PARP-P} & & & \\
            \quad $70\%\rightarrow90\%$ & 1 & 51.9 & 69.1 \\
            \quad $60\%\rightarrow80\%\rightarrow90\%$ & 2 & 33.6 & 53.3 \\
            \bottomrule
        \end{tabular}}
        \end{center}
    \end{minipage}
    \hfill
    \begin{minipage}{0.47\linewidth}
        \centering\small
        \captionof{table}{WER comparison of {\tt PARP} for LSR with previous speech SSL results on Librispeech 10min.
        {\tt PARP} discovers sparse subnetworks within {\tt wav2vec2} with lower WER while adding minimal computational cost to the original ASR finetuning.}
        \label{tab:wer_comparison2}
        \begin{center}
        \vspace{-3mm}
        \scalebox{0.9}{
        \begin{tabular}{lcc}
            \toprule
            \multirow{2}{*}{Method} & test & test \\
            & clean & other \\
            \midrule 
            Continuous BERT~\cite{baevski2019effectiveness} + LM & 49.5 & 66.3 \\
            Discrete BERT~\cite{baevski2019effectiveness} + LM & 16.3 & 25.2 \\
            {\tt wav2vec2-base} reported~\cite{baevski2020wav2vec} & 46.9 & 50.9 \\
            {\tt wav2vec2-large} reported~\cite{baevski2020wav2vec} & 43.5 & 45.3 \\
            {\tt wav2vec2-base} replicated  & 49.3  & 53.2 \\
            {\tt wav2vec2-large} replicated & 46.3  & 48.1 \\
            \midrule
            {\tt wav2vec2-base} w/ 10\% {\tt PARP}  & 38.0 & 44.3 \\
            {\tt wav2vec2-large} w/ 10\% {\tt PARP} & 33.7 & 37.2 \\
            \bottomrule
        \end{tabular}} 
        \end{center}
    \end{minipage}
\end{figure}
\vspace{-2mm}

\begin{figure} [h]
\includegraphics[width=1\linewidth]{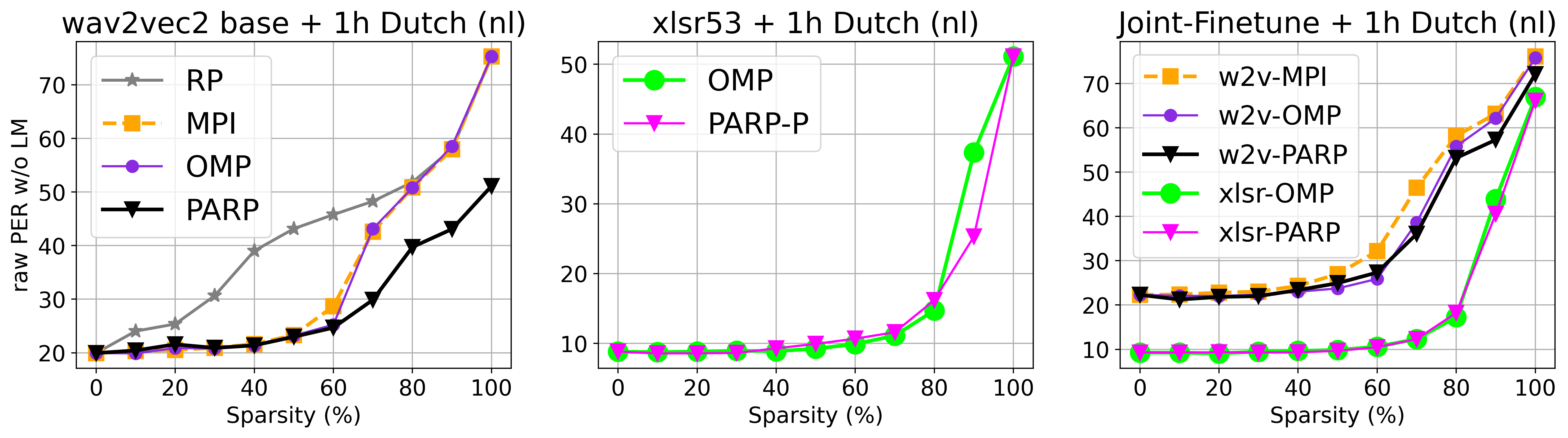}
\centering
\vspace{-3mm}
\caption{Comparison of pruning techniques on H2L \& CSR with 1h of Dutch (\textit{nl}) ASR finetuning. \textbf{(Left)} Pruning H2L ({\tt wav2vec2-base} + \textit{nl}). \textbf{(Center)} Pruning CSR ({\tt xlsr} + \textit{nl}). \textbf{(Right)} Pruning jointly-finetuned {\tt wav2vec2-base} and {\tt xlsr} on \textit{nl}.
Trend is consistent for other 9 spoken languages.}
\vspace{-3mm}
\label{fig:multilin_per_nl}
\end{figure}

\begin{figure}[ht]
    \begin{minipage}{0.6\linewidth}
        \small
        \captionof{table}{Comparing subnetworks discovered by {\tt OMP} and {\tt PARP} from {\tt wav2vec2-base} and {\tt xlsr} with prior work on H2L and CSR. PER is averaged over 10 languages.}
        \label{tab:wer_comparison3}
        \begin{center}
        \vspace{-2.5mm}
        \scalebox{0.9}{
        \begin{tabular}{lccccccccc}
            \toprule
            Method & Pre-training & Sparsity & avg. PER \\
            \midrule 
            Bottleneck~\cite{fer2017multilingually} & Babel-1070h & 0\% & 44.9 \\
            CPC~\cite{oord2018representation} & LS-100h & 0\% & 50.9 \\
            Modified CPC~\cite{riviere2020unsupervised} & LS-360h & 0\% & 44.5 \\
            \midrule 
            {\tt wav2vec2-base} & LS-960h & 0\% & 18.7 \\
            {\tt wav2vec2} + {\tt OMP} & LS-960h & 70\% & 41.3 \\
            {\tt wav2vec2} + {\tt PARP} & LS-960h & 90\% & 40.1 \\
            \midrule
            {\tt xlsr} reported~\cite{conneau2020unsupervised} & 56,000h & 0\% & 7.6 \\
            {\tt xlsr} replicated & 56,000h & 0\% & 9.9 \\
            {\tt xlsr} + {\tt OMP} & 56,000h & 90\% & 33.9 \\
            {\tt xlsr} + {\tt PARP-P} & 56,000h & 90\% & 22.9 \\
            \bottomrule
        \end{tabular}}
        \vspace{-4mm}
        \end{center}
    \end{minipage}
    \hfill
    \begin{minipage}{0.38\linewidth}
        \centering
        \includegraphics[width=\linewidth]{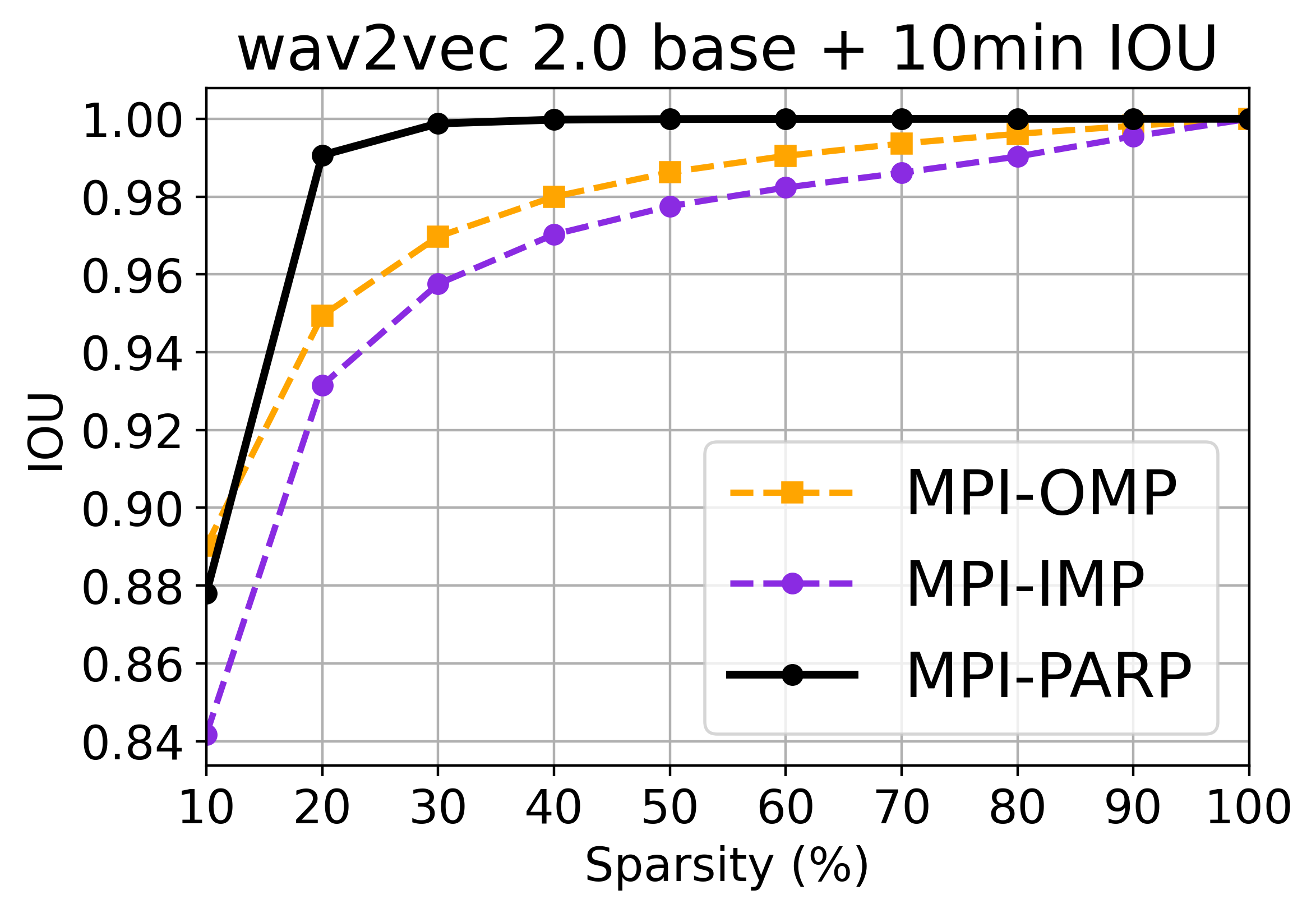}
        \vspace{-1.5em}
        \caption{{\tt PARP}'s final subnetwork and its initial {\tt MPI} subnetwork exceeds 99.99\% {\tt IOU} after 20\% sparsity (black line).}
        \label{fig:iou_change}
    \end{minipage}
\end{figure}

Next, we examine if the pruning results of LSR transfers to H2L and CSR.
Figure~\ref{fig:multilin_per_nl} is pruning H2L and CSR with 1h of Dutch (\textit{nl}) finetuning, and the same conclusion can be extended to other spoken languages. 
Comparing Figures~\ref{fig:main_ler_results} and~\ref{fig:multilin_per_nl}, we notice that shapes of their pruning curves are different, which can be attributed to the effect of character versus phone predictions. 
Comparing left and center of Figure~\ref{fig:multilin_per_nl}, we show that {\tt PARP} and {\tt OMP} reach 50\% sparsity on H2L and 70\% sparsity on CSR with minimal degradations. 
Furthermore, while {\tt PARP} is more effective than {\tt OMP} on H2L for all sparsities, such advantage is only visible in the higher sparsity regions on CSR.
Lastly, Table~\ref{tab:wer_comparison3} compares the subnetworks from H2L and CSR with prior work.
Even with as high as 90\% sparsities in either settings, subnetworks from {\tt PARP} and {\tt OMP} out-performs prior art. 

\vspace{-2mm}

\vspace{2mm}
\subsection{How Important is the Initial Subnetwork (Step 1) in {\tt PARP}?}
\label{subsec:exp_ablation}
\begin{figure} []
\includegraphics[width=1\linewidth]{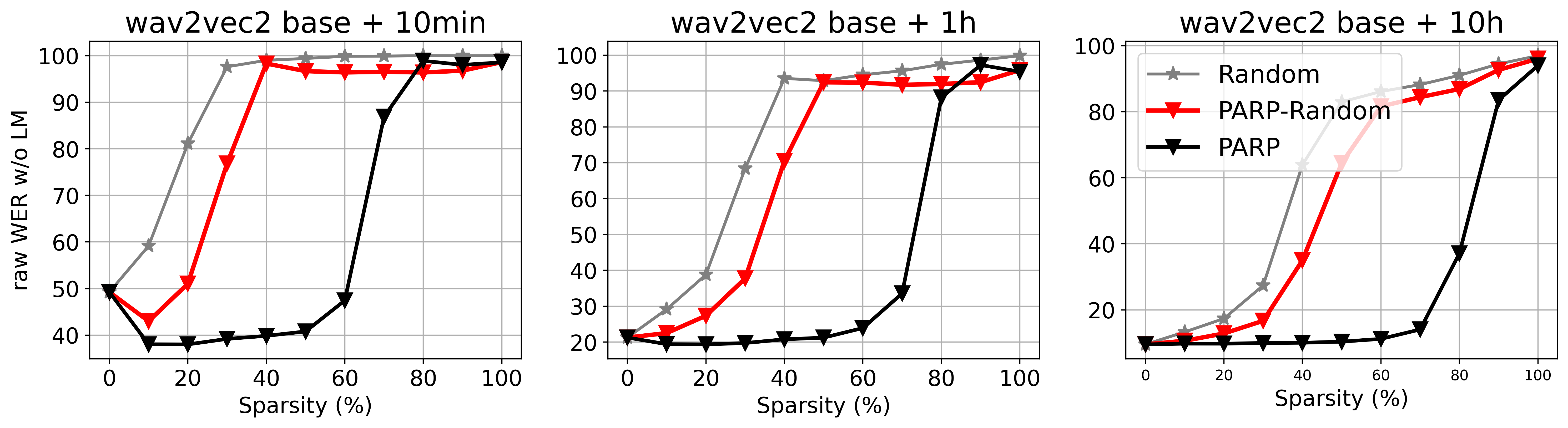}
\centering
\vspace{-2mm}
\caption{{\tt PARP} with random (red line) v.s. with {\tt MPI} (black line) initial subnetworks in LSR.
}
\label{fig:parp+random}
\end{figure}

\vspace{-2mm}
Obtaining a good initial subnetwork (Step 1) is critical for {\tt PARP}, as Adjust \& Re-Prune (Step 2) is operated on top of it. 
In this section, we isolate the effect of Step 1 from Step 2 and examine the role of the initial subnetwork in {\tt PARP}.
Figure~\ref{fig:parp+random} shows {\tt PARP} with a random subnetwork from {\tt RP}, instead of subnetwork from {\tt MPI}, as the initial subnetwork. 
{\tt PARP} with random initial subnetwork performs nearly as bad as {\tt RP} (grey line), signifying the importance of the initial subnetwork. 

Secondly, despite Observation~\ref{observation:similarity}, {\tt MPI} in high sparsity regions (e.g. 90\% in LSR) is not a good initial subnetwork, since the majority of the weights are already pruned out (thus is hard to be recovered from). 
From Figure~\ref{fig:main_ler_results}, {\tt PARP} performs only on par or even worse than {\tt IMP} in high sparsity regions.
In contrast, {\tt PARP-P} starts with a relatively lower sparsity (e.g. 60\% or 70\% {\tt MPI}), and progressively prunes up to the target sparsity. 
Doing so yields considerable performance gain (up to over 50\% absolute WER reduction). 
Third, as shown in Figure~\ref{fig:iou_change}, there is >99.99\% {\tt IOU} between the final ``adjusted'' subnetwork from {\tt PARP} and its initial {\tt MPI} subnetwork after $20\%$ sparsity, confirming Step 2 indeed only made minimal ``adjustment'' to the initial subnetwork.

\vspace{-3mm}

\vspace{0mm}
\subsection{Are Pruning Masks Transferrable across Spoken Languages?}
\label{subsec:mask_transfer}
\vspace{-2mm}

    
Is it possible to discover subnetworks with the wrong guidance, and how transferrable are such subnetworks?
More concretely, we investigate the transferability of {\tt OMP} pruning mask discovered from a source language by finetuning its subnetwork on another target language.
Such study should shed some insights on the underlying influence of spoken language structure on network pruning -- that similar language pairs should be transferrable. 
From a practical perspective, consider pruning for an unseen new language in H2L, we could deploy the readily available discovered subnetworks and thus save the additional finetuning and memory costs. 

\begin{figure} [t]
\includegraphics[width=0.45\linewidth]{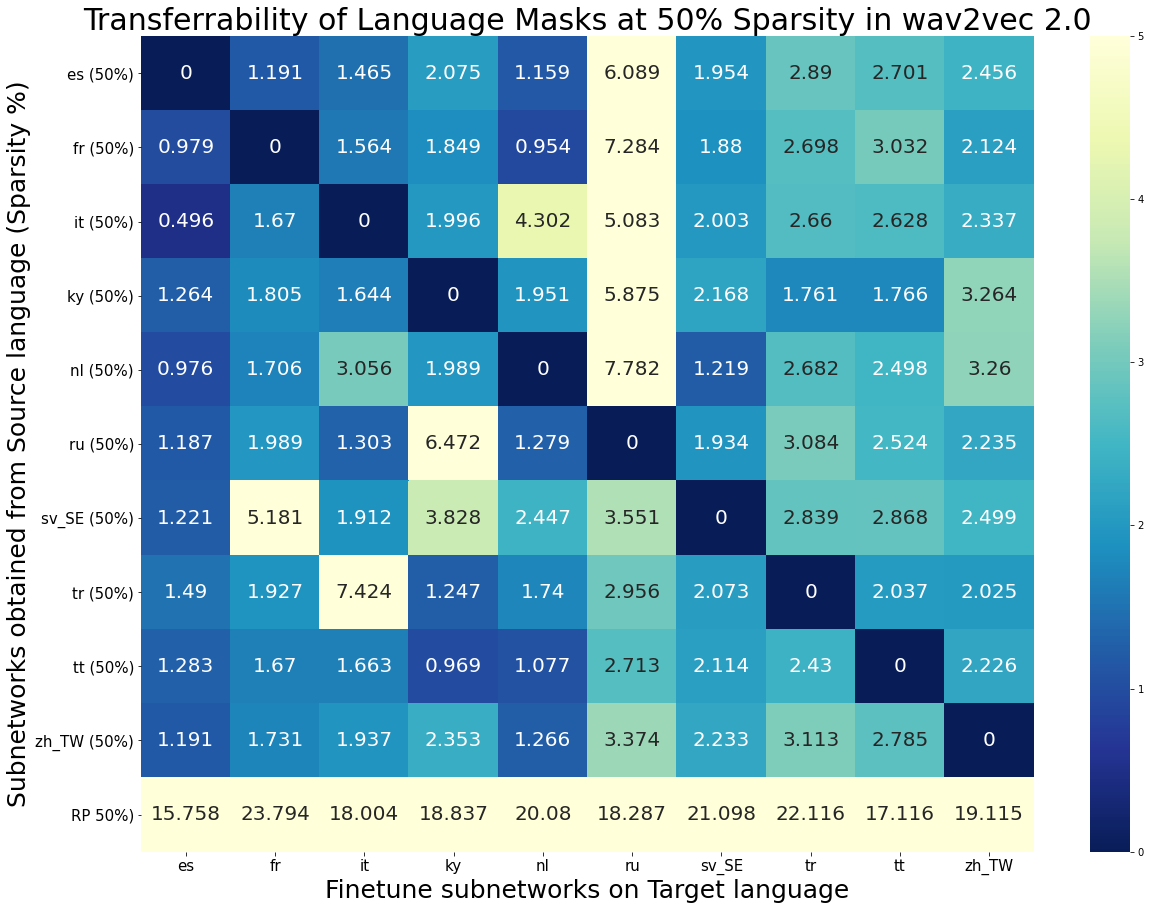}
\hspace{.5cm}
\includegraphics[width=0.46\linewidth]{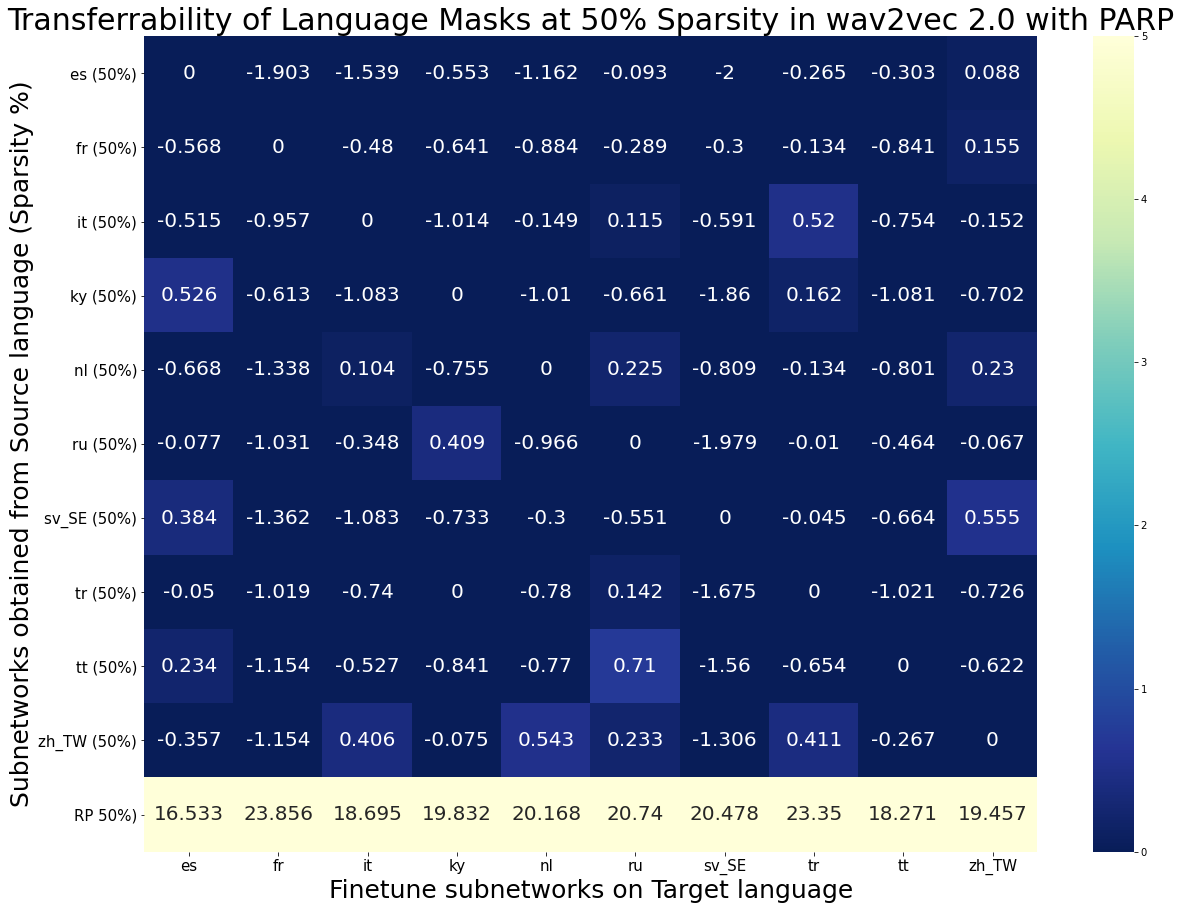}
\centering
\caption{
(\textbf{Left}) Cross-lingual {\tt OMP} mask transfer with regular subnetwork finetuning. 
(\textbf{Right}) Cross-lingual {\tt OMP} mask transfer with PARP.
Last rows are {\tt RP}.
Values are relative PER gains over same-language pair transfer (hence the darker the bettter).
Both are on H2L with pretrained {\tt wav2vec2}. 
The same observation is observed on CSR with pretrained {\tt xlsr} in Appendix~\ref{app:xlsr_mask_transfer}.}
\vspace{-6mm}
\label{fig:wav2vec2_mask_transfer}
\end{figure}

In this case, the initial subnetwork of {\tt PARP} is given by applying {\tt OMP} on another spoken language. 
According to Observation~\ref{observation:similarity}, {\tt PARP}'s Step 2 is effectively under-going cross-lingual subnetwork adaptation for the target language.
Figure~\ref{fig:wav2vec2_mask_transfer} shows the transferability results on H2L with pre-trained {\tt wav2vec2-base}.
On the left is a subnetwork at 50\% sparsity transfer with regular finetuning that contains subtle language clusters -- for example, when finetuning on \textit{ru}, source masks from \textit{es, fr, it, ky, nl} induces a much higher PER compare to that from \textit{sv-SE, tr, tt, zh-TW}.
On the right of Figure~\ref{fig:wav2vec2_mask_transfer}, we show that there is no cross-lingual PER degradation with {\tt PARP}, supporting our claim above. 

\vspace{-3mm}
\subsection{Discovering a Single Subnetwork for 10 Spoken Languages}
\label{subsec:joint}
\vspace{-2mm}
A major downside of pruning pre-trained SSL models for many downstream tasks is the exponential computational and memory costs. 
In H2L and CSR, the same pruning method needs to be repeatedly re-run for each downstream spoken language at each given sparsity. 
Therefore, we investigate the possibility of obtaining a single shared subnetwork for all downstream languages.
Instead of finetuning separately for each language, we construct a joint phoneme dictionary and finetune {\tt wav2vec2} and {\tt xlsr} on all 10 languages jointly in H2L and CSR. 
Note that {\tt PARP} with joint-finetuning can retrieve a shared subnetwork in a single run. 
The shared subnetwork can then be decoded for each language separately. 
The right side of Figure~\ref{fig:multilin_per_nl} illustrates the results. 

\begin{figure} [h]
\includegraphics[width=1.\linewidth]{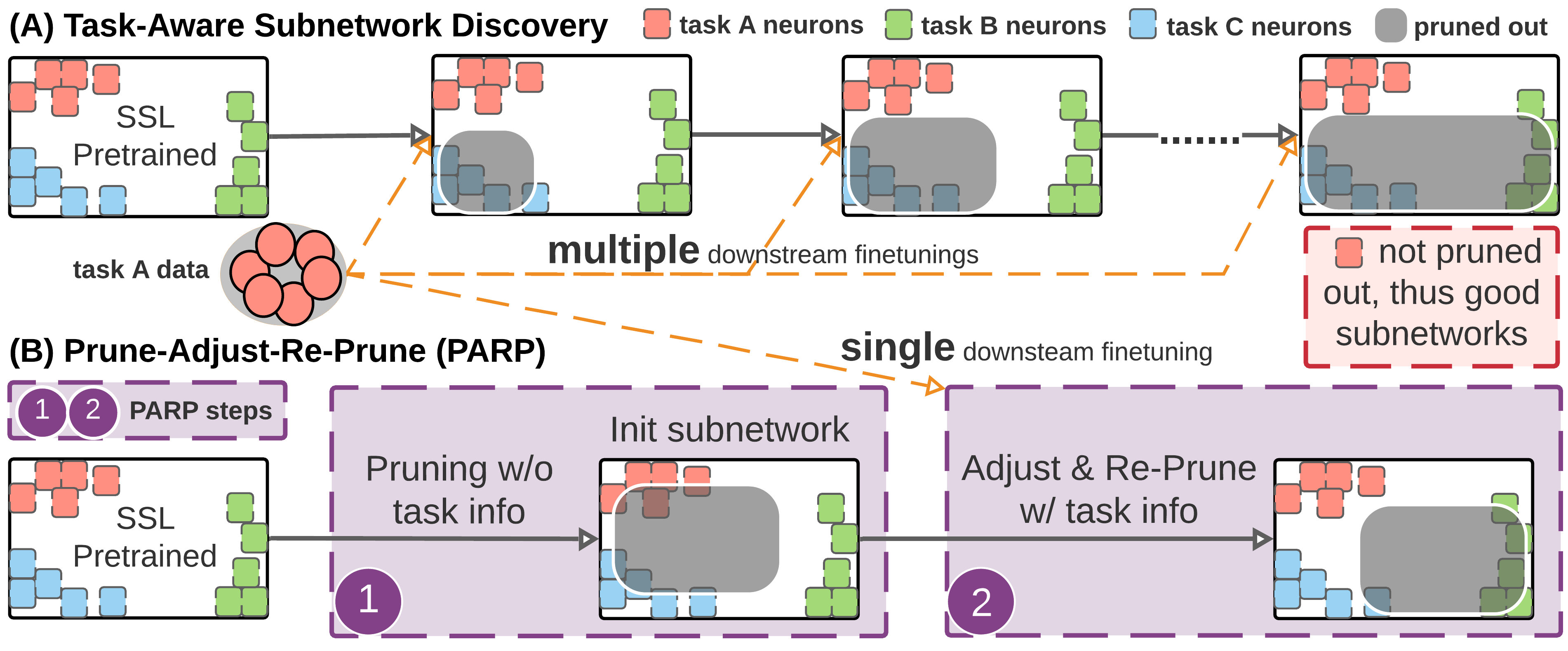}
\centering
\vspace{-5mm}
\caption{
Conceptual sketch of pruning the few task-specific important weights in pretrained SSL.
\textbf{(A)} Task-aware subnetwork discovery({\tt OMP}/{\tt IMP}) is more effective than task-agnostic pruning ({\tt MPI}) since it foresees the important weights in advance, via multiple downstream finetunings. 
\textbf{(B)} {\tt PARP} starts with an initial subnetwork given by {\tt MPI}. 
Observation~\ref{observation:similarity} suggests that the subnetwork is only off by the few important weights, and thus Step 2 revives them by adjusting the initial subnetwork.}
\label{fig:overview}
\end{figure}

Comparing joint-finetuning and individual-finetuning, in H2L, we found that the shared subnetwork obtained via {\tt OMP} has lower PERs between 60$\sim$80\% but slightly higher PERs in other sparsity regions; in CSR, the shared subnetwork from {\tt OMP} has slightly worse PERs at all sparsities. 
Comparing {\tt PARP} to {\tt OMP} in joint-finetuning, we found that while {\tt PARP} is effective in the individual-finetuning setting (left of Figure~\ref{fig:multilin_per_nl}), its shared subnetworks are only slightly better than {\tt OMP} in both H2L and CSR (right of Figure~\ref{fig:multilin_per_nl}). 
The smaller performance gain of {\tt PARP} over {\tt OMP} in pruning jointly-finetuned models is expected, since the important weights for each language are disjoint and  joint-finetuning may send mixed signal to the adjustment step in {\tt PARP} (see Figure~\ref{fig:overview} for better illustration). 

\vspace{-3mm}
\subsection{Does {\tt PARP} work on Pre-trained BERT/XLNet?}
\label{subsec:exp_bert}
\vspace{-2mm}
We also analyzed whether Observation~\ref{observation:similarity} holds for pre-trained BERT/XLNet on 9 GLUE tasks. 
Surprisingly, we found that there are also high (>98\%) overlaps between the 9 tasks' {\tt IMP} pruning masks.
Given this observation, we replicated the cross-task subnetwork transfer experiment (take subnetwork found by {\tt IMP} at task A and finetune it for task B) in BERT-Ticket~\cite{chen2020lottery} on pre-trained BERT/XLNet with {\tt PARP}. 
Table~\ref{tab:parp_on_bert} compares {\tt PARP} (averaged for each target task) to regular finetuning, hinting the applicability of {\tt PARP} to more pre-trained NLP models and downstream natural language tasks. 
Detailed scores and figures are in Appendix~\ref{app:bert_task_transfer}. 

\vspace{-2mm}
\begin{table*}[!ht]
    \small
    \caption{Comparison of cross-task transfer on GLUE (subnetwork from source task A is finetuned for target task B).
    Numbers are averaged acc. across source tasks for each target task.}
    \label{tab:parp_on_bert}
    \begin{center}
    \vspace{-3mm}
    \scalebox{0.9}{
    \begin{tabular}{lccccccccc}
        \toprule
        \multirow{2}{*}{Method} &
        \multicolumn{9}{c}{Averaged transferred subnetworks performance finetuned for} \\
        & CoLA & MRPC & QNLI & QQP & RTE & SST-2 & STS-B & WNLI & MNLI \\
        \midrule 
        \midrule 
        & \multicolumn{9}{c}{70\% sparse subnetworks from pre-trained BERT} \\
        Same-task Transfer (top line) & 38.89 & 75.57 & 88.89 & 89.95 & 58.37 & 89.99 & 87.34 & 53.87 & 82.56 \\
        \cdashlinelr{1-10} 
        Cross-task Transfer with {\tt PARP} & \bf 28.48 & \bf 75.98 & \bf 87.12	& \bf 90.40 & \bf 59.69 & \bf 89.59 & \bf 86.25 & \bf 54.62 & \bf 81.61 \\
        Regular Cross-task Transfer~\cite{chen2020lottery} & 10.12	& 71.94	& 86.54	& 88.50	& 57.59	& 88.80	& 80.27	& 54.03 & 80.48 \\        
        \midrule 
        \midrule 
        & \multicolumn{9}{c}{70\% sparse subnetworks from pre-trained XLNet} \\
        Same-task Transfer (top line) & 29.92 & 76.47 & 89.62 & 90.74 & 59.21 & 92.2 & 80.78 & 42.25 & 85.16 \\
        \cdashlinelr{1-10} 
        Cross-task Transfer with {\tt PARP} & \bf 30.09 & \bf 77.56 & \bf 87.10	& \bf 90.66 & \bf 58.88 & \bf 91.73 & \bf 83.80 & \bf 52.11 & \bf 83.87  \\
        Regular Cross-task Transfer~\cite{chen2020lottery} & 11.47 & 74.16 & 85.21 & 89.11	& 55.80	& 90.19	& 75.61	& 42.25 & 82.65 \\        
        \bottomrule
    \end{tabular}}
    \end{center}
\end{table*}




\vspace{-6.5mm}

\vspace{-5mm}
\subsection{Implications}
\label{subsec:implications}
\vspace{-2mm}
Observation~\ref{observation:similarity} is consistent with the findings of probing large pre-trained NLP models, that pre-trained SSL models are over-parametrized and there exist task-oriented weights/neurons. 
Figure~\ref{fig:lan_mask_overlap_matrix} implies that these important weights only account for a small part of the pre-trained speech SSL. 
In fact, a large body of NLP work is dedicated to studying task-oriented weights in pre-trained models. 
To name a few,~\cite{durrani2020analyzing,dalvi2019one,bau2018identifying,xin2019part} measured,~\cite{bau2018identifying,dai2021knowledge,kovaleva2019revealing} leveraged,~\cite{mu2020compositional,goh2021multimodal} visualized, and~\cite{voita2019analyzing,dalvi2020analyzing,cao2021low} pruned out these important weights/neurons via probing and quantifying contextualized representations.  
Based on Observation~\ref{observation:similarity}, we can project that these NLP results should in general transfer to speech, see pioneering studies~\cite{belinkov2017analyzing,belinkov2019analyzing,chung2021similarity,chowdhury2021end}. 
However, different from them, {\tt PARP} leverages important weights for {\tt UMP} on the whole network structure instead of just the contextualized representations.  

We could further hypothesize that a good pruning algorithm avoids pruning out task-specific neurons in pre-trained SSL~\cite{lee2018snip,guo2016dynamic,molchanov2019importance}, see Figure~\ref{fig:overview}.
This hypothesis not only offers an explanation on why {\tt PARP} is effective in high sparsity regions and cross-lingual mask transfer, it also suggests that an iterative method such as {\tt IMP} is superior to {\tt OMP} because {\tt IMP} gradually avoids pruning out important weights in several iterations, at the cost of more compute\footnote{From Section 6 of~\cite{frankle2018lottery}: "iterative pruning is computationally intensive, requiring training a network 15 or more times consecutively for multiple
trials." From Section 1 of~\cite{guo2016dynamic}: "several iterations of alternate pruning and retraining are necessary to get a fair compression rate on AlexNet, while each retraining process consists of millions of iterations, which can be very time consuming."}.
Finally, we make connections to prior work that showed {\tt RP} prevail~\cite{blalock2020state,chen2020lottery,liu2018rethinking,malach2020proving,ramanujan2020s} -- under a certain threshold and setting, task-specific neurons are less likely to get ``accidentally'' pruned and thus accuracy is preserved even with {\tt RP}. 

 

\vspace{-2mm}
\section{Related Work}
\vspace{-2mm}
\label{sec:related}
\textbf{Modern Speech Paradigm and ASR Pruning.}
As model scale~\cite{synnaeve2019end,baevski2020wav2vec,han2020contextnet,gulati2020conformer,yu2020universal,pratap2020scaling,pratap2020massively,yu2021dual,chen2021continuous,you2021speechmoe,li2021scaling} and model pre-training~\cite{baevski2020wav2vec,zhang2020pushing,conneau2020unsupervised,kong2020panns,jiang2020speech,lai2021semi,hsu2021hubert,xu2021self,chan2021speechstew,kanda2021large,sanabria2021talk,saeed2021contrastive,ng2021pushing,polyak2021speech,wang2021contrastive} have become the two essential ingredients for obtaining SOTA performance in ASR and other speech tasks, applying and developing various forms of memory-efficient algorithms, such as network pruning, to these large-scale pre-trained models will predictably soon become an indispensable research endeavor. 
Early work on ASR pruning can be dated back to pruning decoding search spaces~\cite{abdou2004beam,pylkkonen2005new,siivola2007growing,he2014reshaping,xu2018pruned,zhang2021tiny} and HMM state space~\cite{van1996adaptive}. 
Since the seminal work of Yu et al.~\cite{yu2012exploiting}, ASR pruning has focused primarily on end-to-end network architecture:~\cite{shangguan2019optimizing,wu2021dynamic} applied pruning and quantization to LSTM-based RNN-Transducers,~\cite{panchapagesan2021efficient} applied knowledge distillation to Conformer-based RNN-Transducers,~\cite{venkatesh2021memory,shi2021emformer,li2021efficient} designed efficient architecture/mechanisms for LSTM, Transformer, Conformer-based ASR models,~\cite{narang2017exploring} applied pruning to Deep Speech,~\cite{braun2019parameter} introduced SNR-based probabilistic pruning on LSTM-based CTC model,~\cite{gao2020rethinking} proposed entropy-regularizer for LSTM-based ASR model, ~\cite{xue2013restructuring,povey2018semi} applied SVD on ASR models' weight matrices. 
We emphasize that our work is the first on pruning large self-supervised pre-trained models for low-resource and multi-lingual ASR. 
In addition, to our knowledge, none of the prior speech pruning work demonstrated the pruned models attain superior performance than its original counterpart.

\vspace{-3mm}
\section{Conclusions}
\vspace{-2mm}
\label{sec:conclusion}
\vspace{-1mm}
We introduce {\tt PARP}, a simple and intuitive pruning method for self-supervised speech recognition. 
We conduct extensive experiments on pruning pre-trained wav2vec 2.0 and XLSR-53 under three low-resource settings, demonstrating (1) {\tt PARP} discovers better subnetworks than baseline pruning methods while requiring a fraction of their computational cost, (2) the discovered subnetworks yields over 10\% WER reduction over the full model, (3) {\tt PARP} induces minimal cross-lingual subnetwork adaptation errors, (4) {\tt PARP} can discover a shared subnetwork for multiple spoken languages in one pass, and (5) {\tt PARP} significantly reduces cross-task adaptation errors of pre-trained BERT/XLNet. 
Beyond the scope of our study, we aspire {\tt PARP} as the beginning of many future endeavours on developing more efficient speech SSL models. 

\textbf{Broader Impact.} The broader impact of this research work is making speech technologies more accessible in two orthogonal dimensions: (i) extending modern-day speech technology to many under-explored low-resource spoken languages, and (ii) introducing a new and flexible pruning technique to current and future speech SSL frameworks that reduces the computational costs required for adapting (finetuning) them to custom settings. 
We do not see its potential societal harm.

\vspace{-3mm}
\section*{Limitations and Future Work}
\vspace{-2mm}
\label{sec:limitations}
We make clear of the major limitations of our work, and the full list is in Appendix~\ref{app:limitations}.
The basis of all the pruning methods in the study is unstructured magnitude weight pruning. 
Although sparsity is explicitly enforced in the models, we do not suggest that the sparse models are more memory or energy efficient than the original dense models.
We do believe that our methodology and results should provide meaningful insights and be easily extended upon to more advanced unstructured or structured pruning methods.
We are also curious of the possibility of finetuning or storing modern speech SSL models on local hardware devices. 

Results on cross-lingual mask transfer on pre-trained wav2vec 2.0 in Section~\ref{subsec:mask_transfer} is limited to ASR.
We do not claim pruning masks to be transferrable across speech tasks (e.g. prune {\tt wav2vec2} for speaker ID and transfer for ASR).  
We provide a pilot cross-task mask transfer study on 3 speech tasks (phone recognition, speaker recognition, slot-filling) in SUPERB~\cite{yang2021superb}, and results is in Appendix~\ref{app:cross_task_superb}. 

We claim {\tt PARP} \textit{could} improve the downstream ASR performance over the full wav2vec 2.0, yet we do not claim it as a plug-and-play method into any SOTA ASR pipeline, such as~\cite{zhang2021bigssl}, to get a performance boost.
We provide a preliminary experiment on combining {\tt PARP} and transformer-LM decoding in Appendix~\ref{app:lm_decode_parp}.
Nonetheless, due to resource limitations and to isolate the effect of pruning, it remains upon investigations on the complete effects of speech pruning in different setups.


\vspace{-3mm}
\section*{Acknowledgments}
\vspace{-2mm}
\label{sec:acknowledgments}
\vspace{-1mm}
We thank IBM for the donation to MIT of the Satori GPU cluster, and John Cohn for maintaining the cluster. 
We also thank Lucy Chai, Wei-Ning Hsu, Desh Raj, Shu-wen Leo Yang, Abdelrahman Mohamedm, Erica Cooper, and anonymous reviewers for helpful suggestions and paper editing. 
This work is part of the low-resource language learning project funded by the MIT-IBM Waston AI Lab.


\newpage
{\small
\bibliographystyle{plain}
\bibliography{ref}}


\clearpage
\title{Supplementary Material}
\maketitleextra
\vspace{-7ex}
\setcounter{tocdepth}{2}
\tableofcontents
\newpage

\section{NeurIPS Paper Checklist}


\begin{enumerate}

\item For all authors...
\begin{enumerate}
  \item Do the main claims made in the abstract and introduction accurately reflect the paper's contributions and scope?
    \answerYes{Through experiments results are in Section~\ref{sec:exp}. For example, our claim that {\tt PARP} outperforms LTH is visible in Figure~\ref{fig:main_ler_results}.}
  \item Did you describe the limitations of your work?
    \answerYes{Refer to Section~\ref{sec:limitations} and Appendix~\ref{app:limitations}.}
  \item Did you discuss any potential negative societal impacts of your work?
    \answerNo{We mention in Section~\ref{sec:conclusion} on the broader impact of this research work. 
    Since this work is on pruning existing speech SSL models for low-resource spoken languages, we do not see its potential negative societal impacts. 
    However, we welcome reviewers and AC to raise such concerns, and we will include corresponding statements.}
  \item Have you read the ethics review guidelines and ensured that your paper conforms to them?
    \answerYes{}
\end{enumerate}

\item If you are including theoretical results...
\begin{enumerate}
  \item Did you state the full set of assumptions of all theoretical results?
    \answerNA{}
	\item Did you include complete proofs of all theoretical results?
    \answerNA{}
\end{enumerate}

\item If you ran experiments...
\begin{enumerate}
  \item Did you include the code, data, and instructions needed to reproduce the main experimental results (either in the supplemental material or as a URL)?
    \answerYes{Due to its simplicity, PARP only adds a few lines of code to. 
    Data and pre-trained models are all publicly available. These details are in the Appendix and in our project webpage: \url{https://people.csail.mit.edu/clai24/parp/}.}
  \item Did you specify all the training details (e.g., data splits, hyperparameters, how they were chosen)?
\answerYes{We follow~\cite{baevski2020wav2vec,conneau2020unsupervised} for the model configurations and fine-tuning hyper-parameters. These details are in Appendix~\ref{app:model_details}.}
	\item Did you report error bars (e.g., with respect to the random seed after running experiments multiple times)?
    \answerNo{Due to the computational expense and scale of our experiments, we were not able to extensively re-run. We do note that our re-created baselines match the numbers reported in prior work~\cite{baevski2020wav2vec,conneau2020unsupervised}.}
	\item Did you include the total amount of compute and the type of resources used (e.g., type of GPUs, internal cluster, or cloud provider)?
    \answerYes{We briefly mention the compute needed in the footnote in Page 2, and more details are in the Appendix~\ref{app:implementation}}.
\end{enumerate}

\item If you are using existing assets (e.g., code, data, models) or curating/releasing new assets...
\begin{enumerate}
  \item If your work uses existing assets, did you cite the creators?
    \answerYes{Our work (code and pre-trained models) are based on~\cite{baevski2020wav2vec,conneau2020unsupervised}.}
  \item Did you mention the license of the assets?
    \answerNA{}
  \item Did you include any new assets either in the supplemental material or as a URL?
    \answerNA{}
  \item Did you discuss whether and how consent was obtained from people whose data you're using/curating?
    \answerNo{No, we used published datasets and to the best of our knowledge, none of them have consent-related issues.}
  \item Did you discuss whether the data you are using/curating contains personally identifiable information or offensive content?
    \answerNo{We used published datasets and, to the best of our knowledge, all of them have been reviewed carefully by the authors/community.}
\end{enumerate}

\item If you used crowdsourcing or conducted research with human subjects...
\begin{enumerate}
  \item Did you include the full text of instructions given to participants and screenshots, if applicable?
    \answerNA{}
  \item Did you describe any potential participant risks, with links to Institutional Review Board (IRB) approvals, if applicable?
    \answerNA{}
  \item Did you include the estimated hourly wage paid to participants and the total amount spent on participant compensation?
    \answerNA{}
\end{enumerate}

\end{enumerate}

\section{Model Details} 
\label{app:model_details}
    Model and pruning configurations for {\tt wav2vec2-base}, {\tt wav2vec2-large}, and {\tt xlsr} can be found in Section~\ref{app:model_config}. 
    Fintuning hyper-parameters are generally the same as in~\cite{baevski2020wav2vec}, and we detailed them in Section~\ref{app:finetune_config}.
    {\tt PARP}'s hyper-parameter is detailed in Section~\ref{app:parp_config}.
    More details on system implementations is in Section~\ref{app:implementation}.
    \vspace{-1mm}
    \subsection{Model and Pruning Configurations}
    \label{app:model_config}
    wav2vec 2.0 consists of three modules: a 7-layer CNN feature encoder for pre-processing raw speech waveforms, a quantization layer for discretizating, and a BERT for learning contextualized representations. 
    Given that the feature encoder is fixed and the quantization layer is discarded during finetuning, we focus on pruning the BERT module in wav2vec 2.0 and XLSR-53. 
    We also do not prune the positional embedding layer nor the layer normalization layers within BERT. 
    This setup is consistent with BERT-Ticket~\cite{chen2020lottery}. 
    wav2vec 2.0 BASE ({\tt wav2vec2-base}) is based on BERT-BASE, which has 12 transformer blocks, hidden dimension 768, 12 self-attention heads, and 95M parameters. 
    wav2vec 2.0 LARGE (denote as {\tt wav2vec2-large}) is based on BERT-LARGE, which has 24 transformer blocks, hidden dimension 768, 16 self-attention heads, and 315M parameters. 
    XLSR-53 (denoted as {\tt xlsr}) shares the same architecture as {\tt wav2vec2-large}.
    We took {\tt wav2vec2-base} and {\tt wav2vec2-large} that were pre-trained on Librispeech 960h. 
    {\tt wav2vec2-base}, {\tt wav2vec2-large}, and {\tt xlsr} are pre-trained with the contrastive predictive coding objective.  
    
    \textbf{More on Pruning Configuration.}
    There are 3 components in {\tt wav2vec2}/{\tt xlsr} that we did not prune out: (1) CNN feature extractor, (2) layer norm running statistics, and (3) positional embedding/task-specific linear layer. 
    For (1), it is due to the CNN feature extractor being fixed during finetuning by default, and the majority of the model parameters lie in the BERT module in {\tt wav2vec2}/{\tt xlsr}. 
    For (2)(3), we simply follow the setup described in BERT-Ticket~\cite{chen2020lottery}. 
    These 3 decisions is why in left of Figure~\ref{fig:multilin_per_nl}, {\tt PARP} (black line) attains $\sim$50\% PER at 100\% sparsity. 
    In fact, while re-producing BERT-Ticket~\cite{chen2020lottery}, we were surprised that BERT’s layer norm statistics plus its final linear layer achieve non trivial loss/accuracy (e.g. BERT’s MLM at 0\% sparsity is $\sim$60\% accuracy while at 100\% sparsity is $\sim$15\% accuracy.). 
    
    \vspace{-1mm}
    \subsection{Finetuning Hyper-Parameters}
    \label{app:finetune_config}
    {\tt wav2vec2} is finetuned for 20k steps on the 10h split, 15k steps on the 1h split, and 12k steps on the 10min split. 
    {\tt xlsr} is finetuned for 12k steps for each spoken languages. 
    In the default setup in~\cite{baevski2020wav2vec}, {\tt wav2vec2} except the final linear layer is freezed for 10k steps, however, we observe doing so on the pruned models may lead to training instability. 
    Therefore, we do not include this trick in our fine-tuning setups.
    The learning rate ramps up linearly for first 10\% of the steps, remains the same for 40\% of the steps, and decay exponentially for 50\% of the steps. 
    The waveform encoder output is randomly masked according to~\cite{baevski2020wav2vec}.
    For LSR, the validation set is the dev-other subset from Librispeech. 
    
    \vspace{-1mm}
    \subsection{{\tt PARP} Hyper-Parameters}
    \label{app:parp_config}
    {\tt PARP} introduces an additional pruning frequency hyper-parameter, $n$ in Algorithm Table~\ref{alg:parp}.
    As long as $n$ is a sensible small number (e.g. 5-50 out of 10k+ steps), the final pruned models should have similar performance.
    We heuristically set $n=5$ for pruning {\tt XLSR} on all spoken language splits; we set $n=50$ for {\tt wav2vec2-base} on 10min/1h, $n=5$ for {\tt wav2vec2-base} on 10h, $n=5$ for {\tt wav2vec2-large} on 10min, $n=2$ for {\tt wav2vec2-large} on 1h, and $n=1$ for {\tt wav2vec2-large} on 10h.

    \vspace{-1mm}
    \subsection{Implementation}
    \label{app:implementation}
    All experiments are based on the Fairseq repository\footnote{\url{https://github.com/pytorch/fairseq}} and Wav2letter++ decoding\footnote{\url{https://github.com/flashlight/wav2letter}}.
    We took publicly available pre-trained {\tt wav2vec2-base}, {\tt wav2vec2-large}, and {\tt xlsr}\footnote{Pre-trained models available at \url{https://github.com/pytorch/fairseq/blob/master/examples/wav2vec/README.md}}.
    The pruning code is based on PyTorch's pruning module~\footnote{\url{https://pytorch.org/tutorials/intermediate/pruning_tutorial.html}}. 
    For each experiment, we fine-tune the model on either 2 or 4 GPUs in parallel, and unlike the standard wav2vec 2.0 fine-tuning setup, we do not include a LM for validation during fine-tuning. 
    Given that not all of our GPUs support FP16, our fine-tuning setup is on FP32. 
    For fair comparison, we imposed a reasonable computational budget for all pruning methods used in this study\footnote{Each finetuning run is capped at a total of 100 V100 hours. For example, {\tt OMP} requires 2 finetunings, so we will run it for at most a total of 50 hours on across 4 V100s.}.

\section{Experimental Setup for LSR, H2L, and CSR}
\label{app:exp_setup}
    For LSR, we finetune pre-trained {\tt wav2vec2-base} and {\tt wav2vec2-large} on the 10h/1h/10min splits from Librispeech and Libri-light, as this is the \textit{de facto} setup for studying speech representation learning~\cite{baevski2020wav2vec}.
    For H2L, we replicate the setting described in~\cite{riviere2020unsupervised,conneau2020unsupervised}, where pre-trained {\tt wav2vec2-base} is finetuned on 10 spoken languages (1 hour each) from CommonVoice: \textit{Spanish (es), French (fr), Italian (it), Kyrgyz (ky), Dutch (nl), Russian (ru), Swedish (sv-SE), Turkish(tr), Tatar (tt), and Mandarin (zh-TW)}.
    For CSR, we replicate the setting in~\cite{conneau2020unsupervised}, where pre-trained {\tt xlsr} is finetuned on the same 10 languages as in H2L.
    Studying LSR can inform us the effect of amount of finetuning supervision (10min$\sim$10h) and pre-trained model scales ({\tt base} v.s. {\tt large}) on pruning;
    on the other hand, comparing CSR and H2L could yield insights on the effect of mono-lingual versus cross-lingual pre-training on pruning.
    
    \textbf{Evaluation Criteria.} Word Error Rate (WER) is reported for LSR; Phone Error Rate (PER) is reported for H2L and CSR\footnote{WER/PER (lower the better) is standard criteria for ASR. This is opposite to previous work on pruning CV or NLP models, where accuracy or BLEU scores (higher the better) was reported.}. 
    Earlier work on pruning sequence to sequence tasks, such as ASR~\cite{braun2019parameter} or Machine Translation~\cite{yu2019playing,gale2019state}, showed that pruned models do not match or outperform the full model, albeit with ``minimal degradation''.
    Moreover, to isolate the effects of different pruning methods, we~\textbf{do not} include any external LM nor any means of self-training~\cite{xu2021self} during training or decoding.  
    To provide an unbiased grounding and accurate reflection of the pruned models, we thus report relative gains of our proposed method over {\tt OMP}/{\tt IMP}/{\tt MPI}, in addition to their raw WER/PERs. 

\section{How important is the {\tt IMP} rewinding starting point?}
\label{app:imp_rewinding}
\begin{figure*} [!hbtp]
    \includegraphics[width=\linewidth]{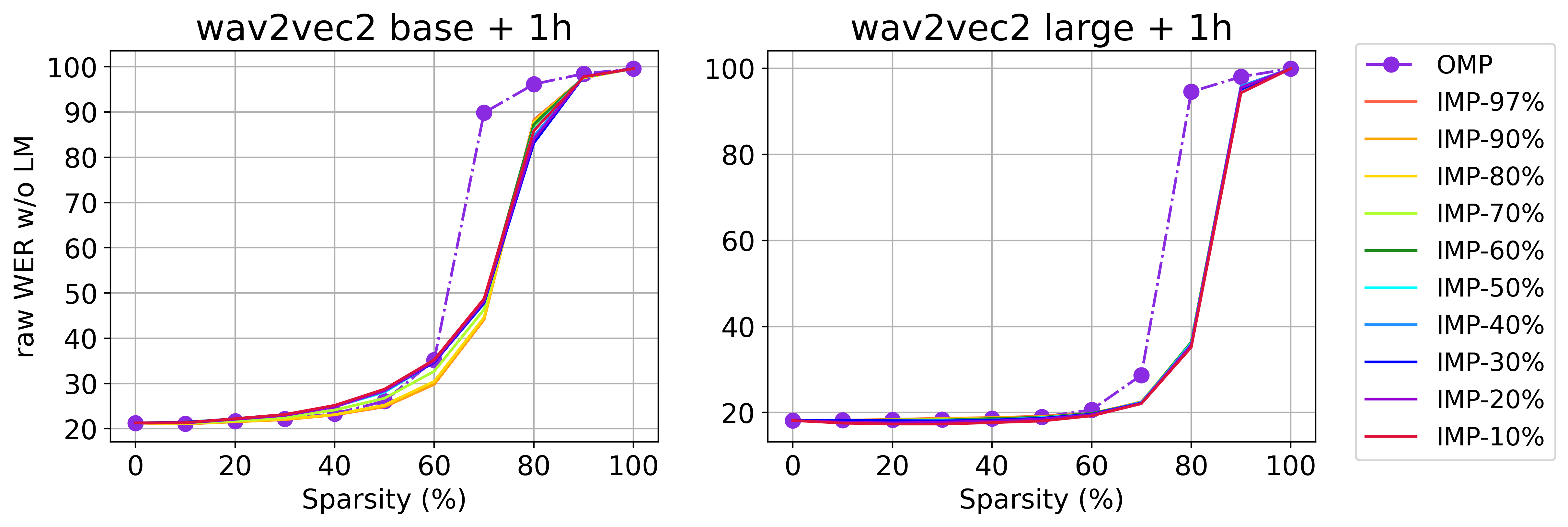}
    \centering
    \caption{{\tt IMP} on {\tt wav2vec2-base} and {\tt wav2vec2-large} with different rewinding starting point within the downstream ASR finetuning.
    Its effect diminishes when pruning {\tt wav2vec2-large}.}
    \label{fig:rewind}
    \end{figure*}

    We also examined the effectiveness of {\tt IMP} rewinding~\cite{frankle2020linear,renda2020comparing} for pruning speech SSL, where instead of re-starting each {\tt IMP} pruning iteration all the way back from pre-trained SSL initializations, the iteration starts at some points during the downstream ASR finetuning. 
    For example, in figure~\ref{fig:rewind}, {\tt IMP} with $10\%$ rewinding (dark red line) means that each pruning iteration starts at $10\%$ into the ASR downstream finetuning;  
    We find that rewinding has minimal effect for pruning speech SSL, which aligns with the results in NLP~\cite{chen2020lottery}. 
    Curiously, we observe the effect diminishes when the pre-training model size is scaled up from {\tt base} to {\tt large}. 
    
\newpage
\section{{\tt OMP} Masks Overlap in H2L and CSR}
\label{app:mask_overlap_matrices}

    We provide the rest of Figure~\ref{fig:lan_mask_overlap_matrix} at other sparsities to support Observation~\ref{observation:similarity}. 
    For readability, we re-state it again: 
    \begin{tcolorbox}
    \vspace{-2mm}
    \textit{For any sparsity, any amount of finetuning supervision, any pre-training model scale, and any downstream spoken languages, the non-zero ASR pruning masks obtained from task-agnostic subnetwork discovery has high {\tt IOU}s with those obtained from task-aware subnetwork discovery.}
    \vspace{-2mm}
    \end{tcolorbox}
    
    In addition to {\tt IOU}, we also provide the overlap percentage between masks\footnote{Instead of taking the Union in the denominator as in {\tt IOU}, simply take the full number of parameters.}.
    We divide this section into {\tt OMP} masks overlap over spoken language pairs on finetuned {\tt wav2vec2-base} in H2L (Section~\ref{app:mask_overlap_matrices-wav2vec2}) and overlaps on finetuned {\tt xlsr} in CSR (Section~\ref{app:mask_overlap_matrices-xlsr}). 
    
    \subsection{{\tt OMP} Masks Overlap in H2L}
    \label{app:mask_overlap_matrices-wav2vec2}
    \textbf{H2L {\tt OMP} masks overlap procedure.} Each set of experiments require 10$\times$10 rounds of {\tt xlsr} finetunings because there are 10 downstream spoken languages ASR. 
    The experimental procedure is: 
    \begin{enumerate}
        \item Finetune {\tt wav2vec2-base} for a source spoken language ASR. 
        \item Prune the finetuned model and obtain an {\tt OMP} mask for each spoken language ASR.
        \item Calculate {\tt IOU}/mask overlap over all pairs of spoken language masks at each sparsity.
    \end{enumerate}

        \begin{figure*} [!hbtp]
        \includegraphics[width=0.47\linewidth]{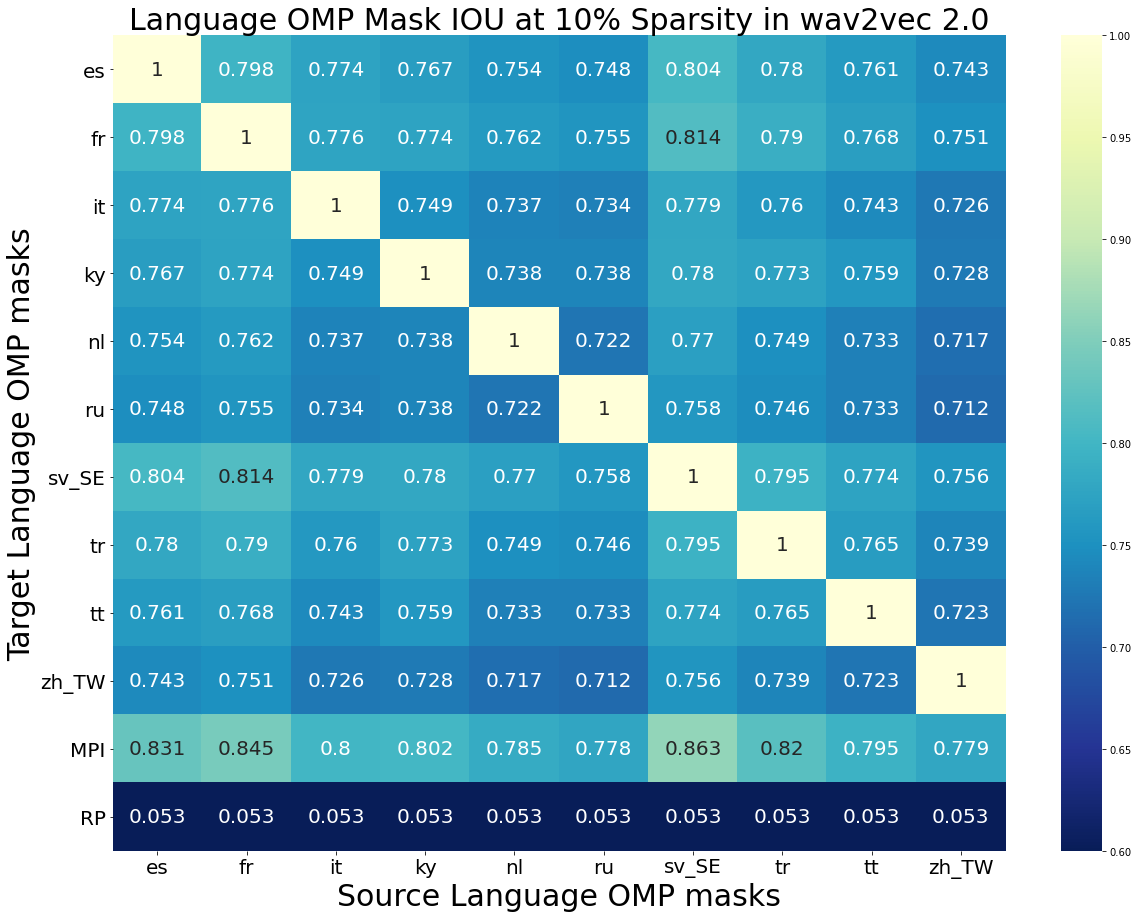}
        \hspace{.5cm}
        \includegraphics[width=0.47\linewidth]{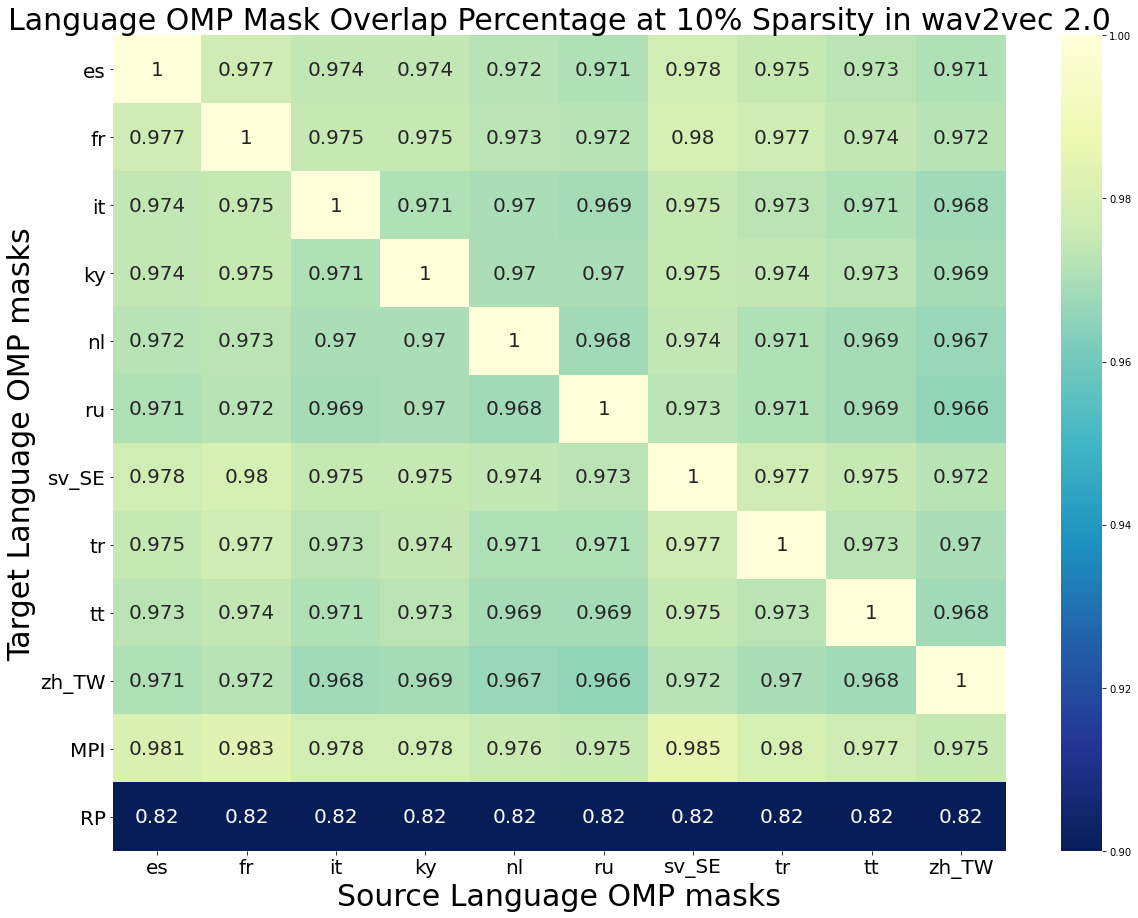}
        \centering
        \caption{
        {\tt OMP} pruning masks {\tt IOU}s and overlap percentages on finetuned {\tt wav2vec2} at 10\% sparsity.}
        \end{figure*}
        
        \begin{figure*} [!h]
        \includegraphics[width=0.47\linewidth]{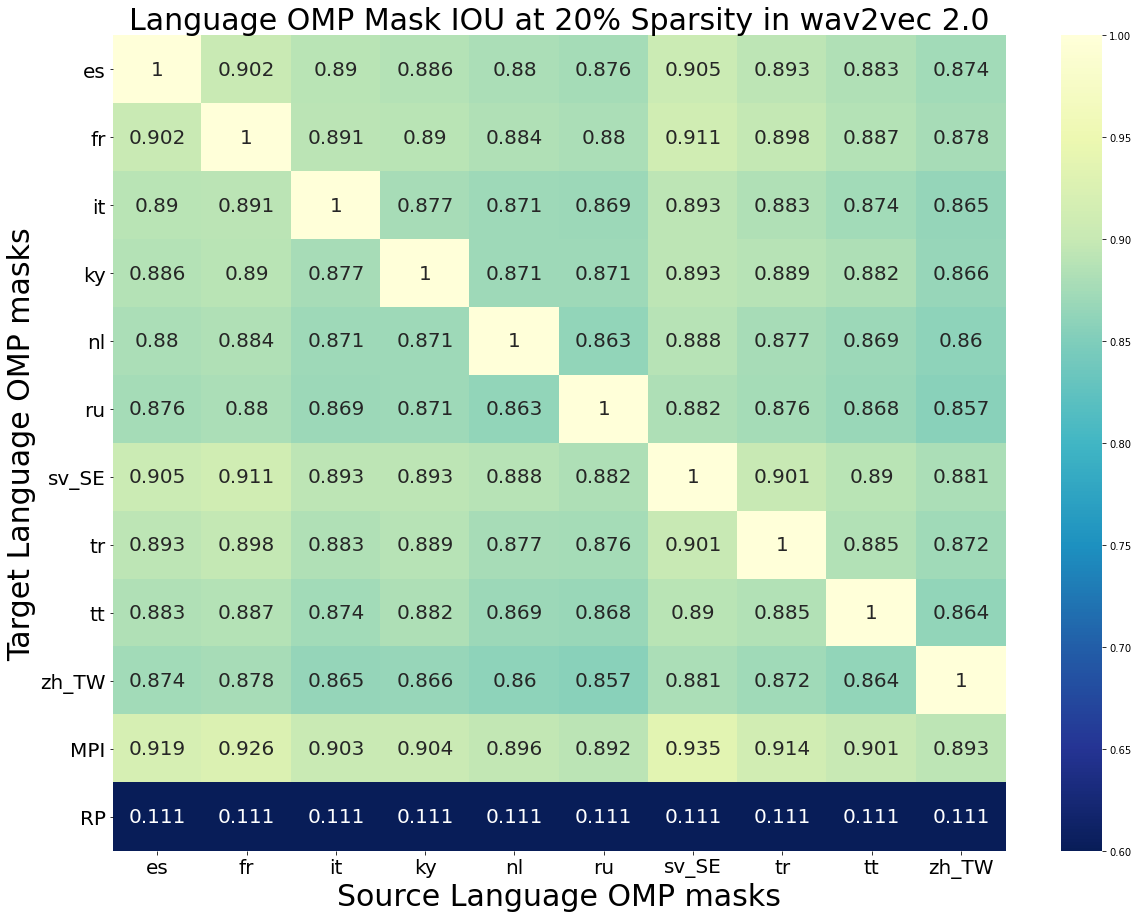}
        \hspace{.5cm}
        \includegraphics[width=0.47\linewidth]{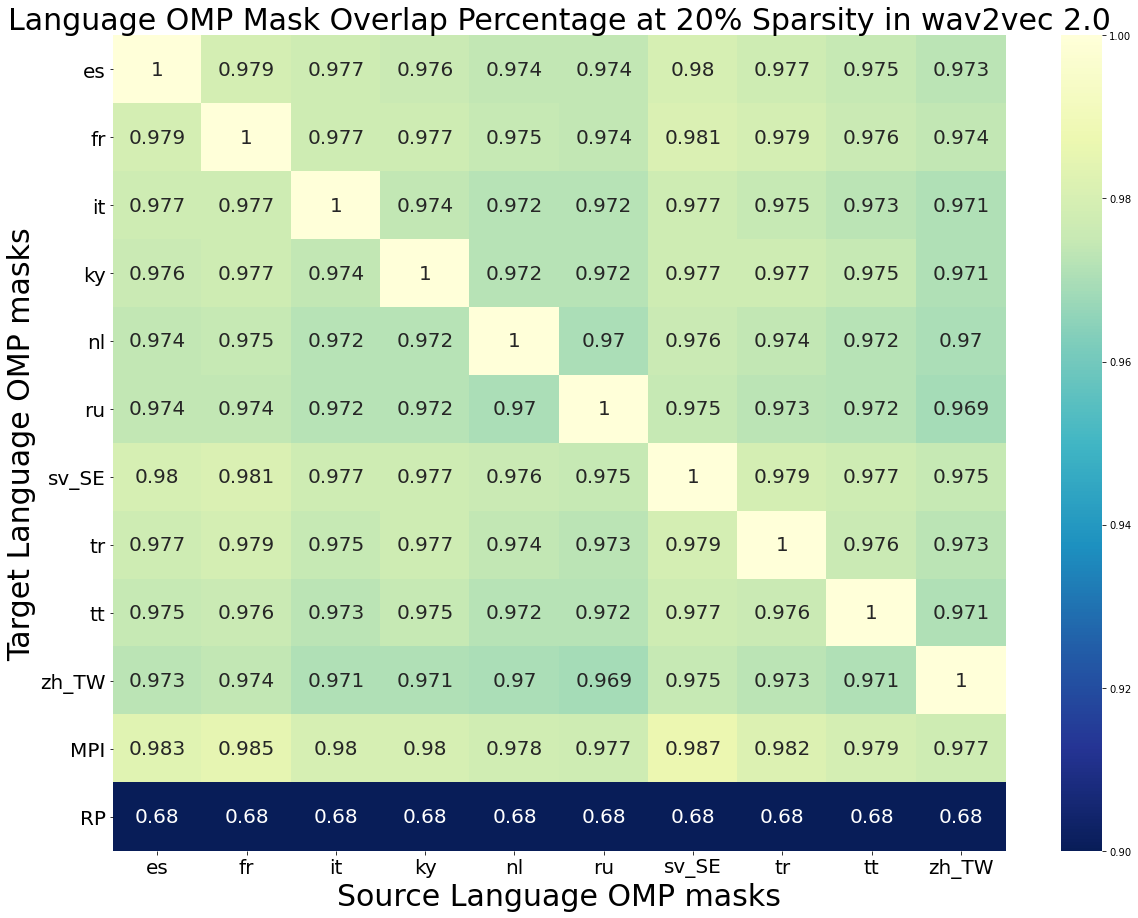}
        \centering
        \caption{
        {\tt OMP} pruning masks {\tt IOU}s and overlap percentages on finetuned {\tt wav2vec2} at 20\% sparsity.}
        \end{figure*}
        
        \begin{figure*} [!h]
        \includegraphics[width=0.47\linewidth]{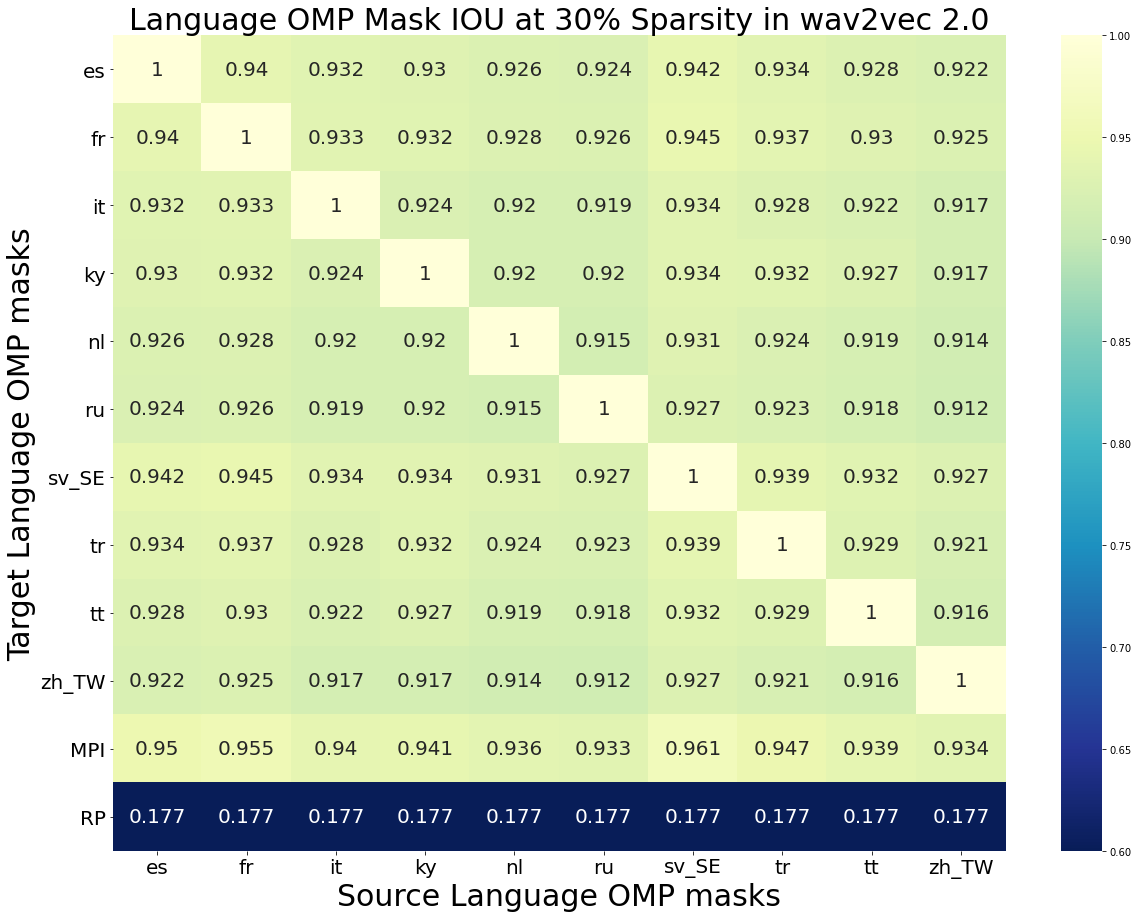}
        \hspace{.5cm}
        \includegraphics[width=0.47\linewidth]{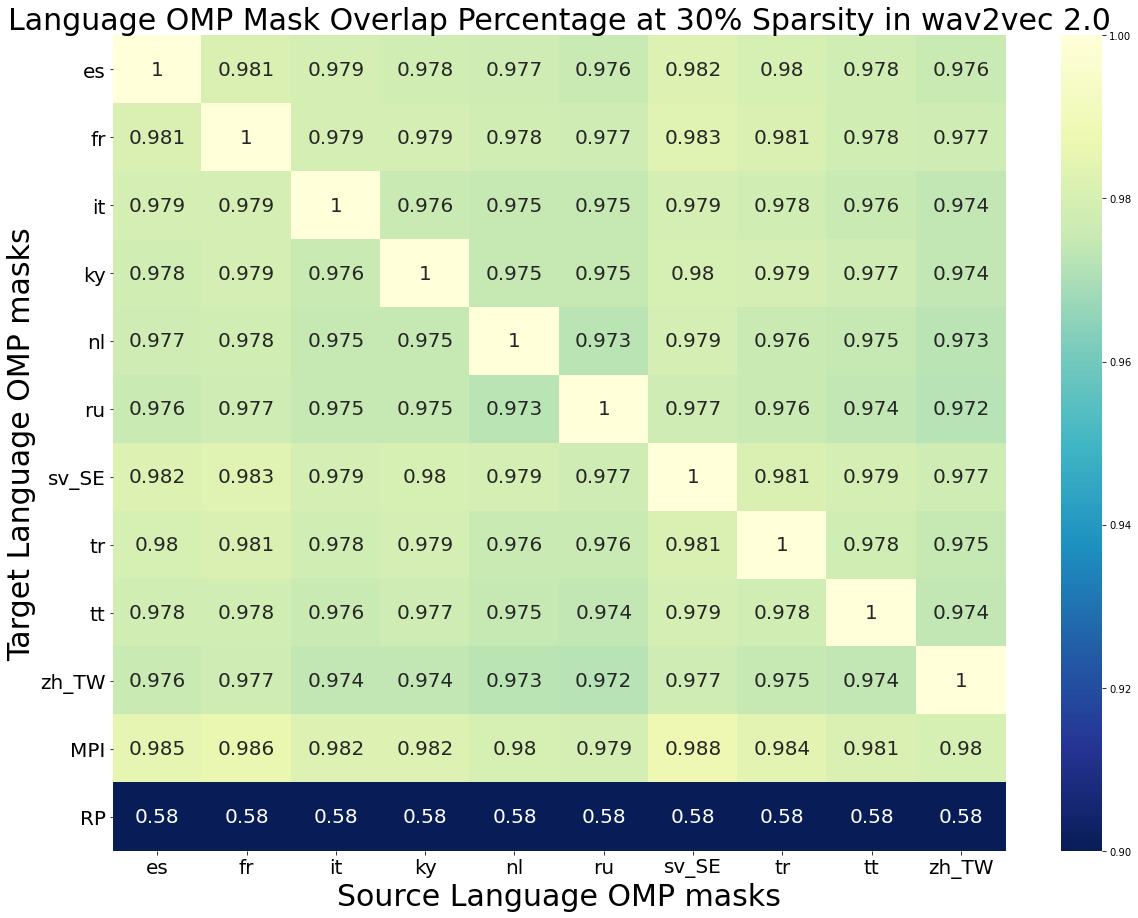}
        \centering
        \caption{
        {\tt OMP} pruning masks {\tt IOU}s and overlap percentages on finetuned {\tt wav2vec2} at 30\% sparsity.}
        \end{figure*}
        
        \begin{figure*} [!h]
        \includegraphics[width=0.47\linewidth]{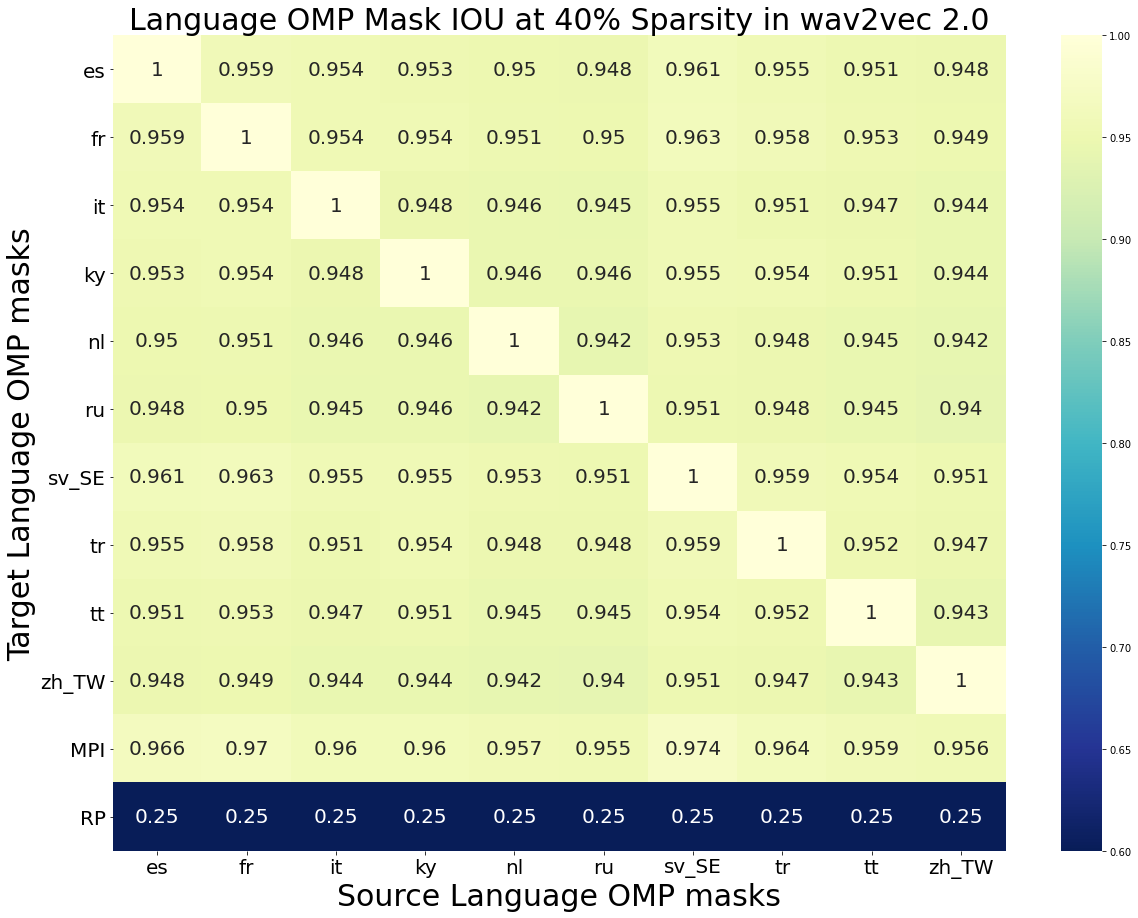}
        \hspace{.5cm}
        \includegraphics[width=0.47\linewidth]{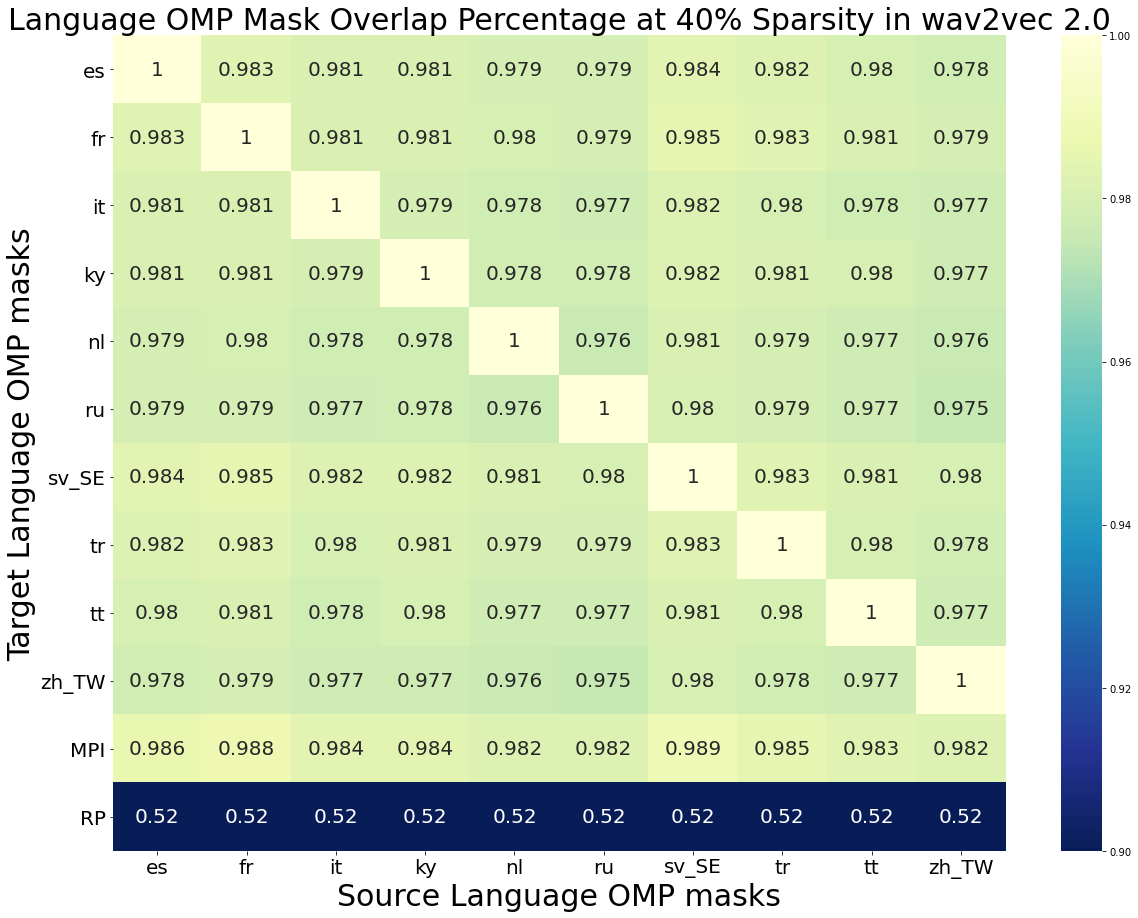}
        \centering
        \caption{
        {\tt OMP} pruning masks {\tt IOU}s and overlap percentages on finetuned {\tt wav2vec2} at 40\% sparsity.}
        \end{figure*}
        
        \begin{figure*} [!h]
        \includegraphics[width=0.47\linewidth]{figs/mask-overlap/wav2vec2_mask_IOU_50.png}
        \hspace{.5cm}
        \includegraphics[width=0.47\linewidth]{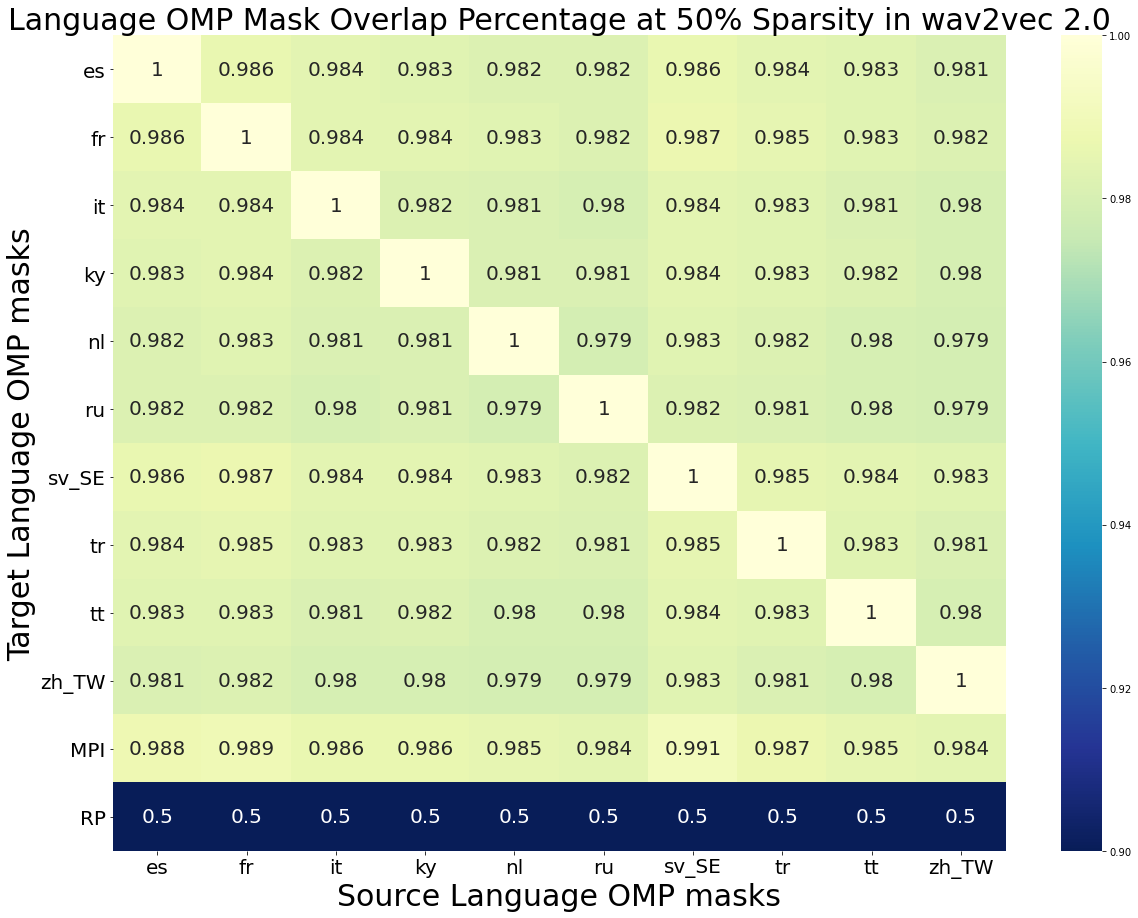}
        \centering
        \caption{
        {\tt OMP} pruning masks {\tt IOU}s and overlap percentages on finetuned {\tt wav2vec2} at 50\% sparsity.}
        \end{figure*}
        
        \begin{figure*} [!h]
        \includegraphics[width=0.47\linewidth]{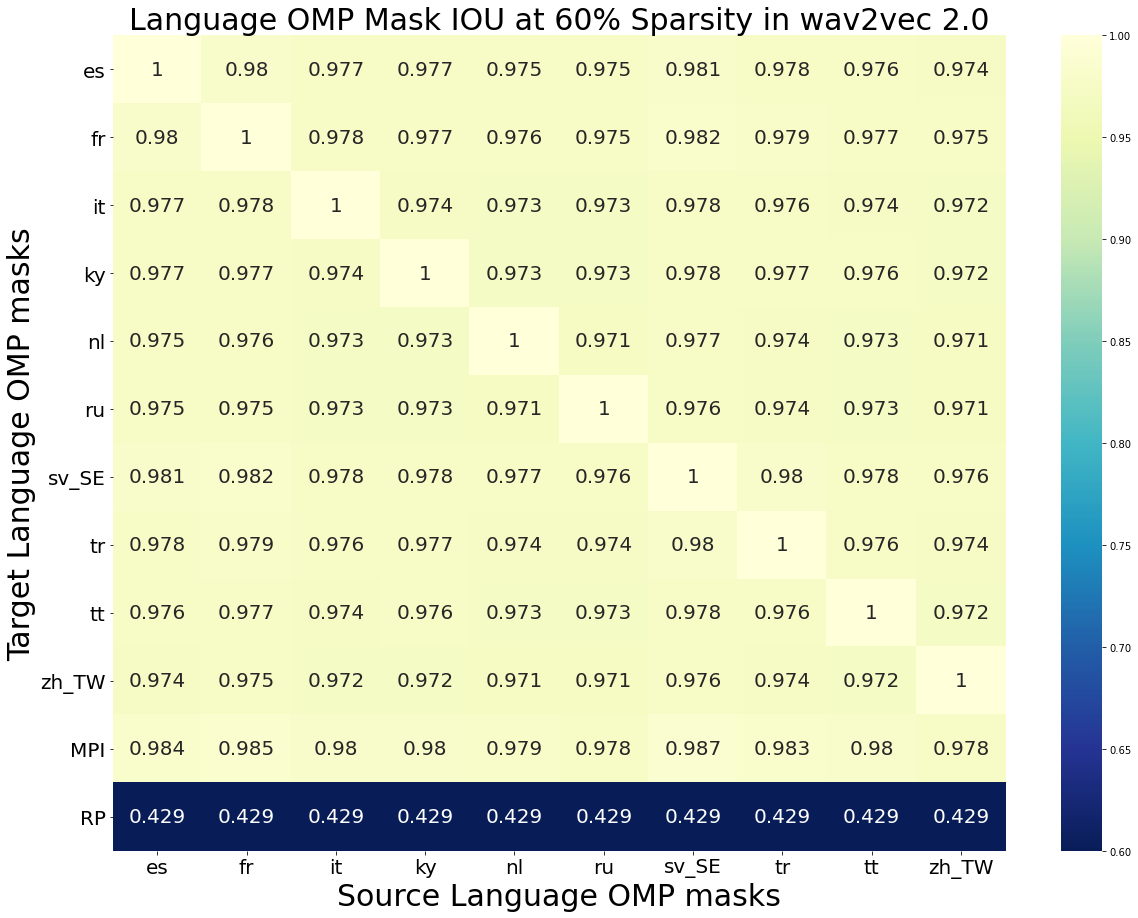}
        \hspace{.5cm}
        \includegraphics[width=0.47\linewidth]{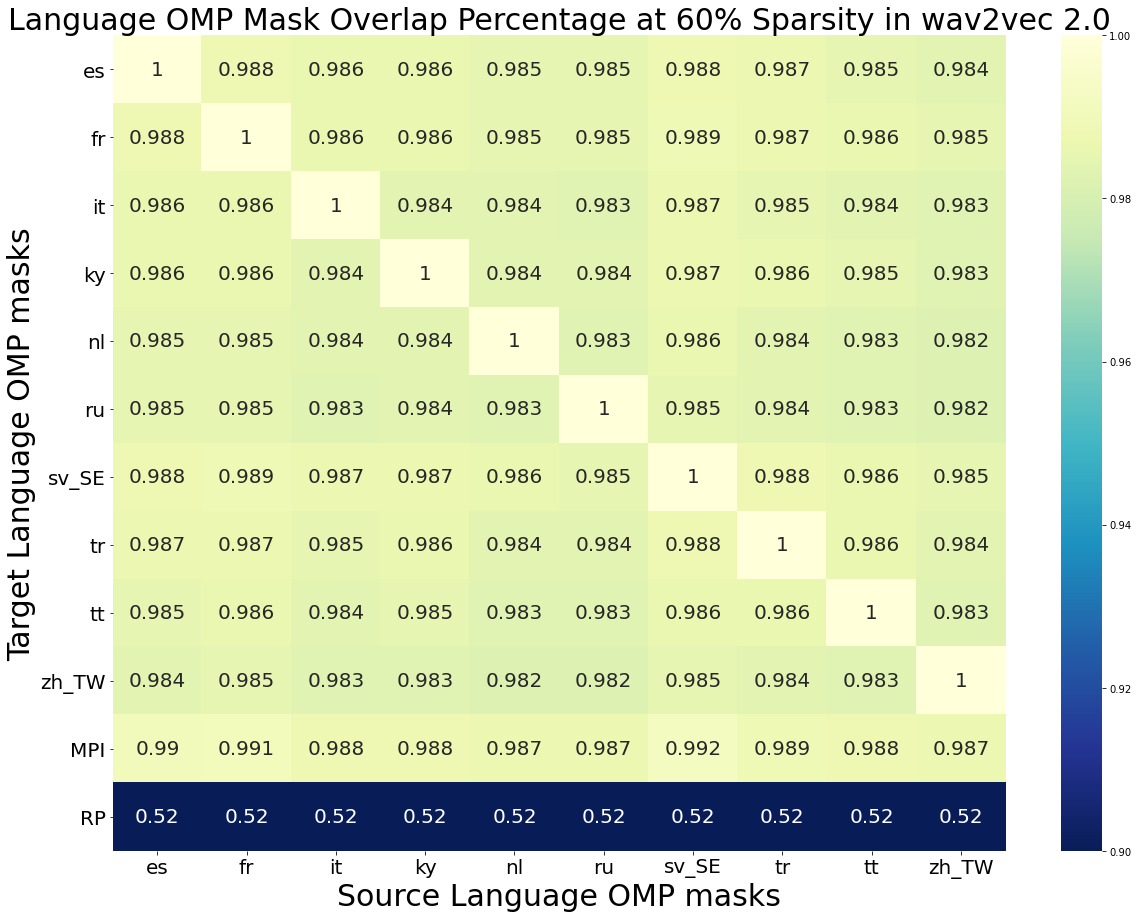}
        \centering
        \caption{
        {\tt OMP} pruning masks {\tt IOU}s and overlap percentages on finetuned {\tt wav2vec2} at 60\% sparsity.}
        \end{figure*}
        
        \begin{figure*} [!h]
        \includegraphics[width=0.47\linewidth]{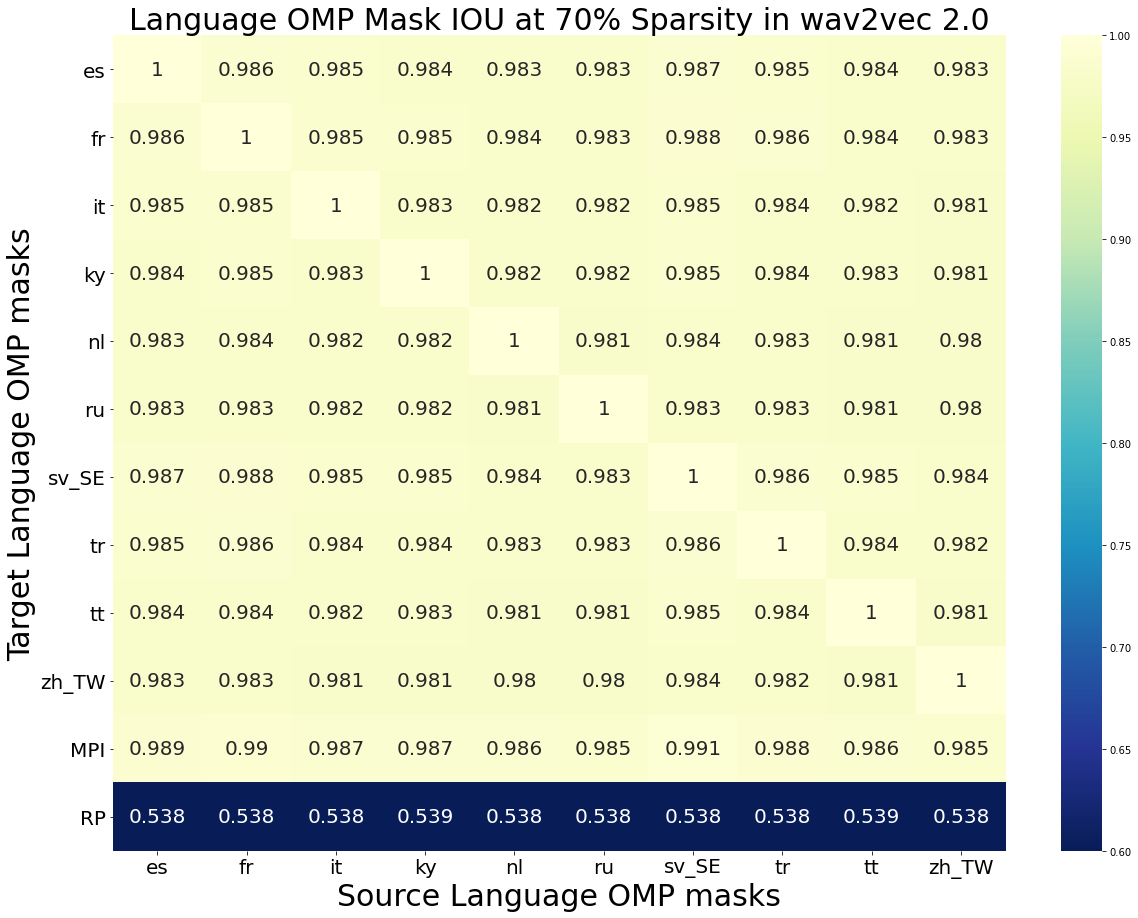}
        \hspace{.5cm}
        \includegraphics[width=0.47\linewidth]{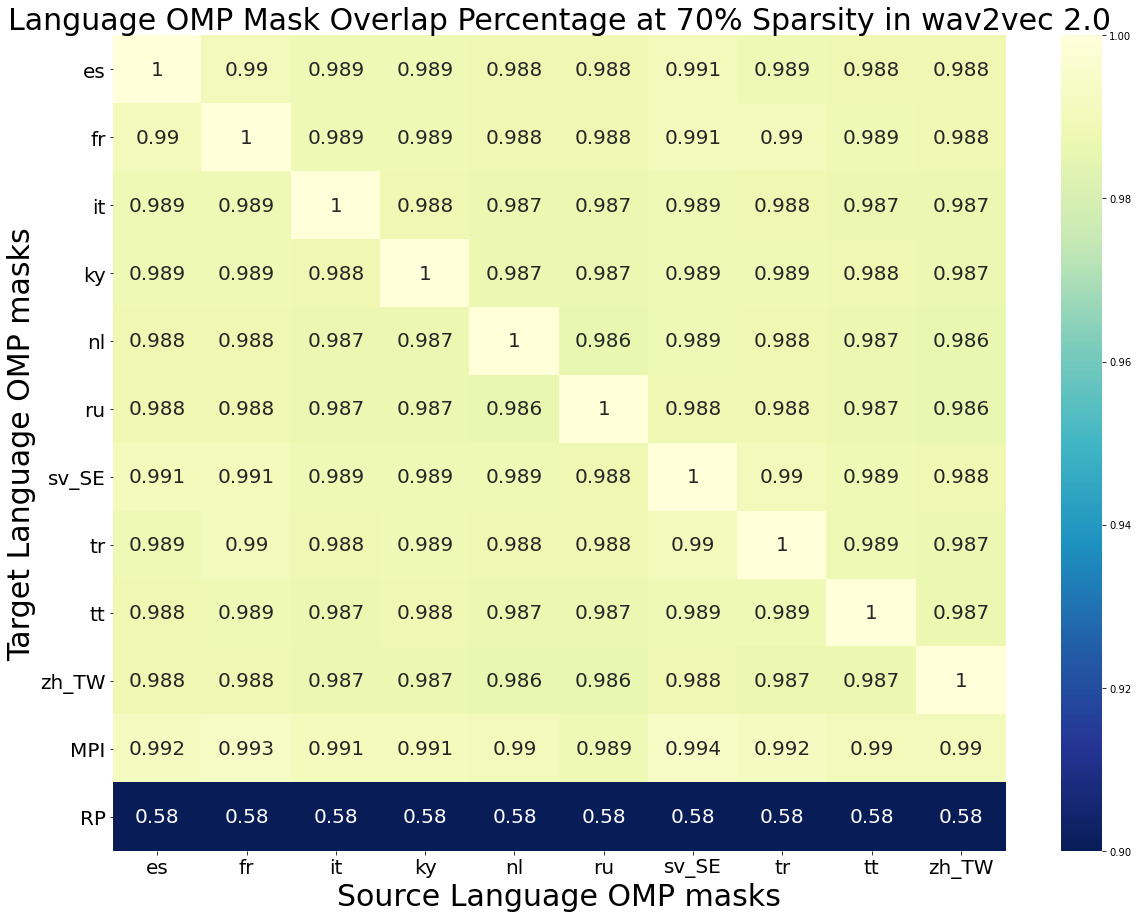}
        \centering
        \caption{
        {\tt OMP} pruning masks {\tt IOU}s and overlap percentages on finetuned {\tt wav2vec2} at 70\% sparsity.}
        \end{figure*}
        
        \begin{figure*} [!h]
        \includegraphics[width=0.47\linewidth]{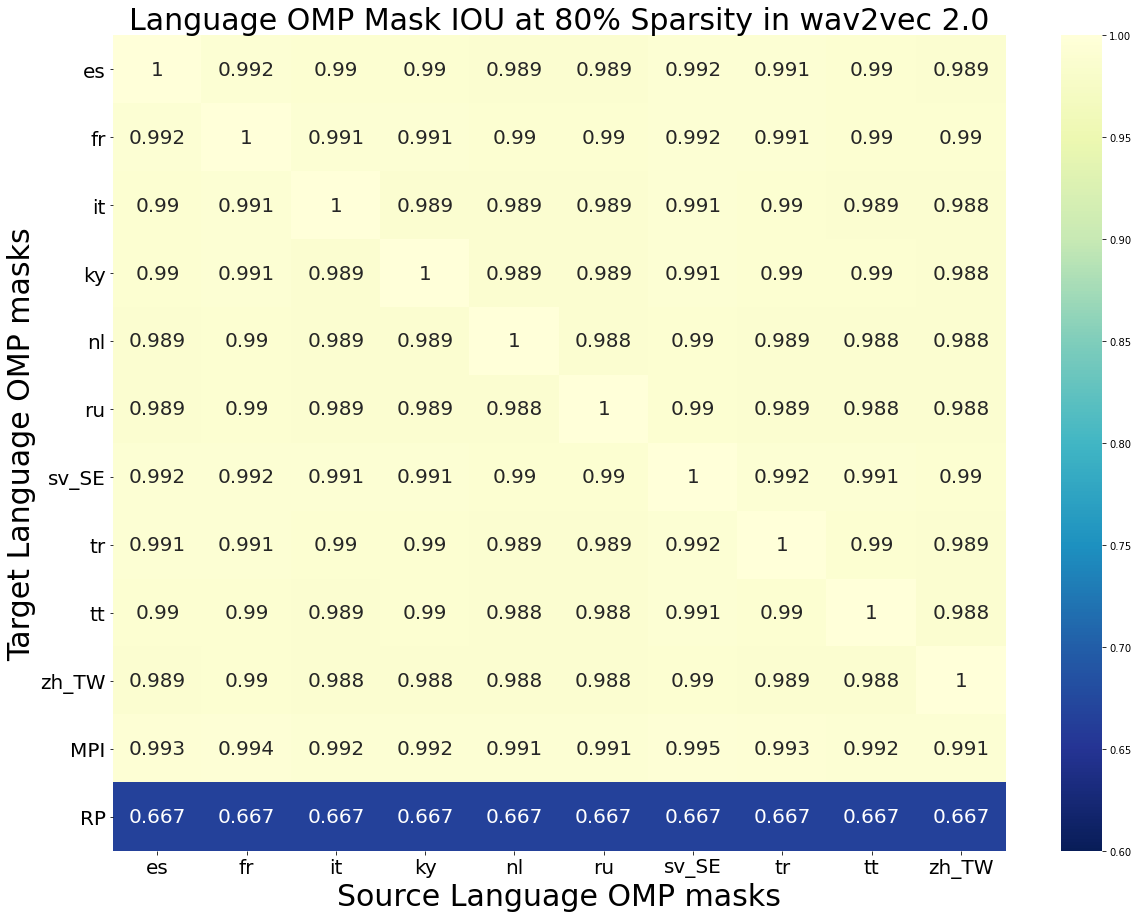}
        \hspace{.5cm}
        \includegraphics[width=0.47\linewidth]{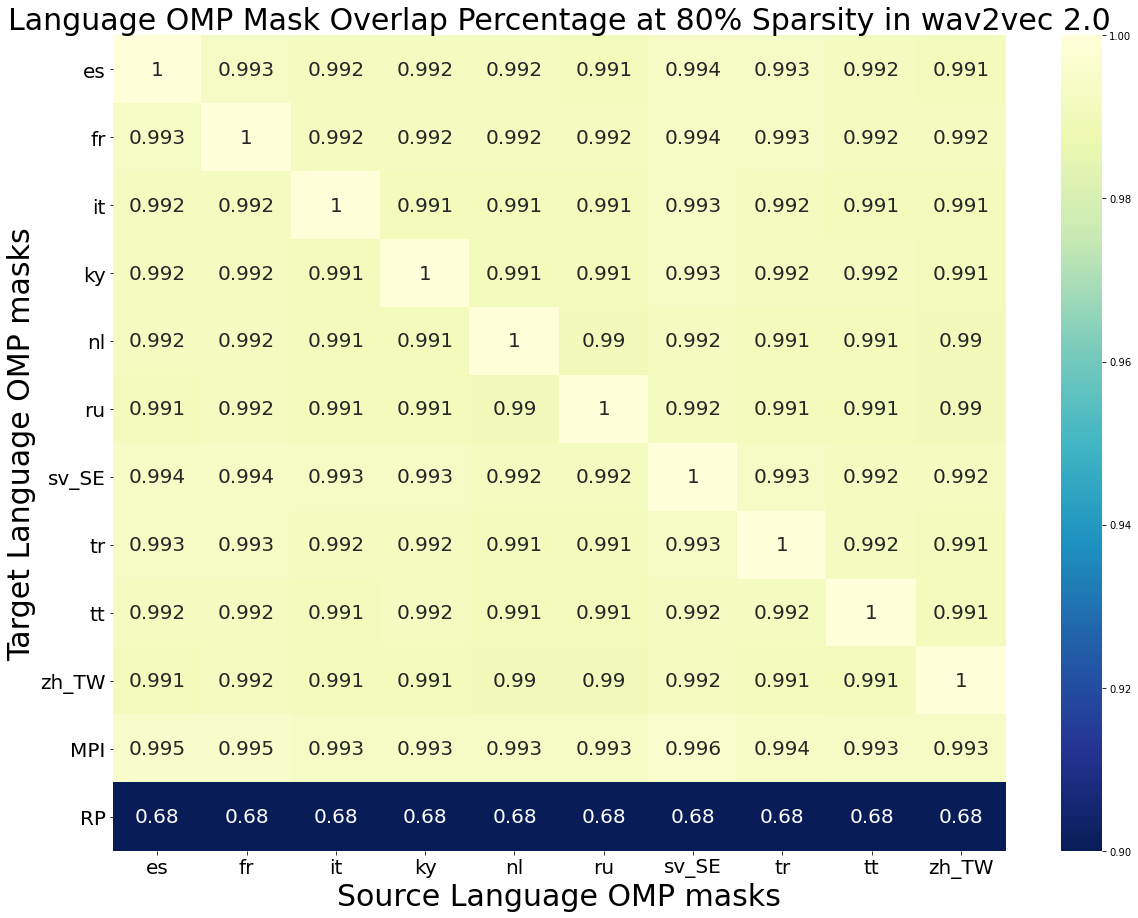}
        \centering
        \caption{
        {\tt OMP} pruning masks {\tt IOU}s and overlap percentages on finetuned {\tt wav2vec2} at 80\% sparsity.}
        \end{figure*}
        
        \begin{figure*} [!h]
        \includegraphics[width=0.47\linewidth]{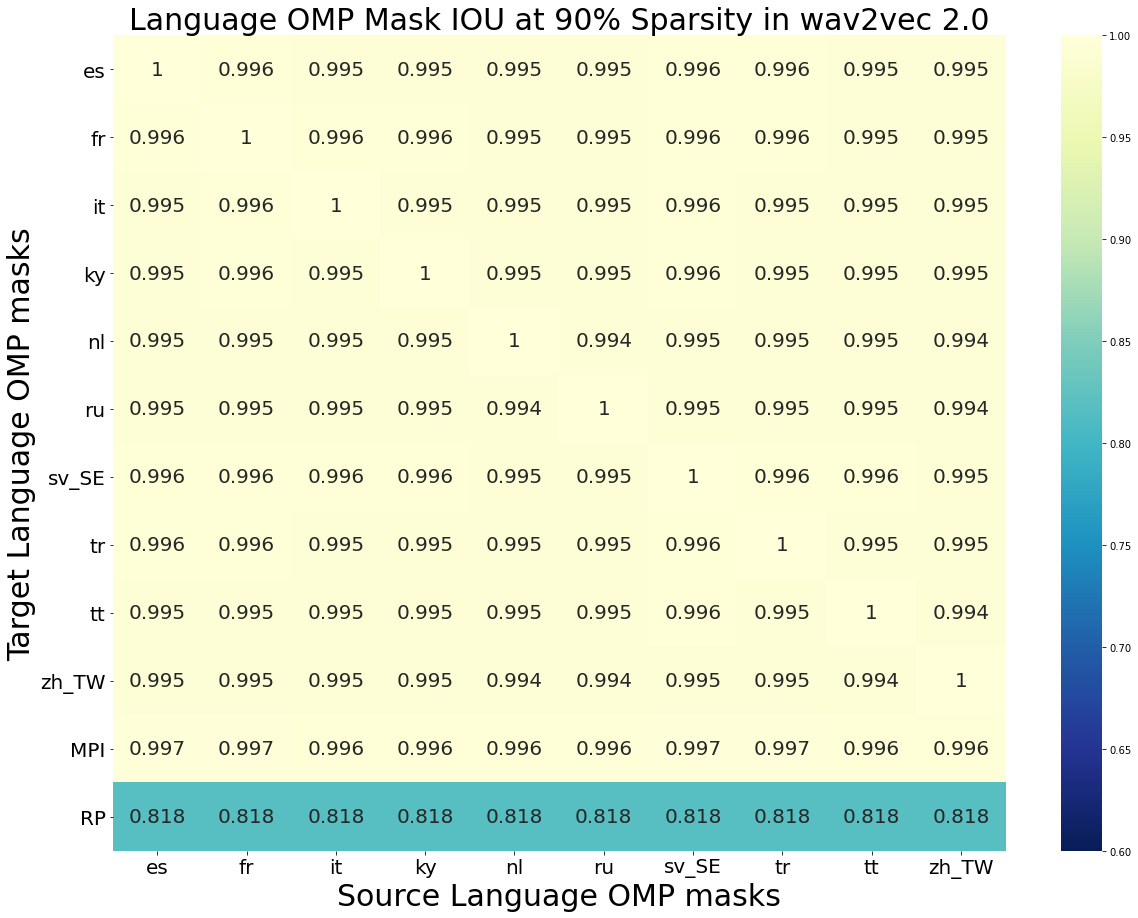}
        \hspace{.5cm}
        \includegraphics[width=0.47\linewidth]{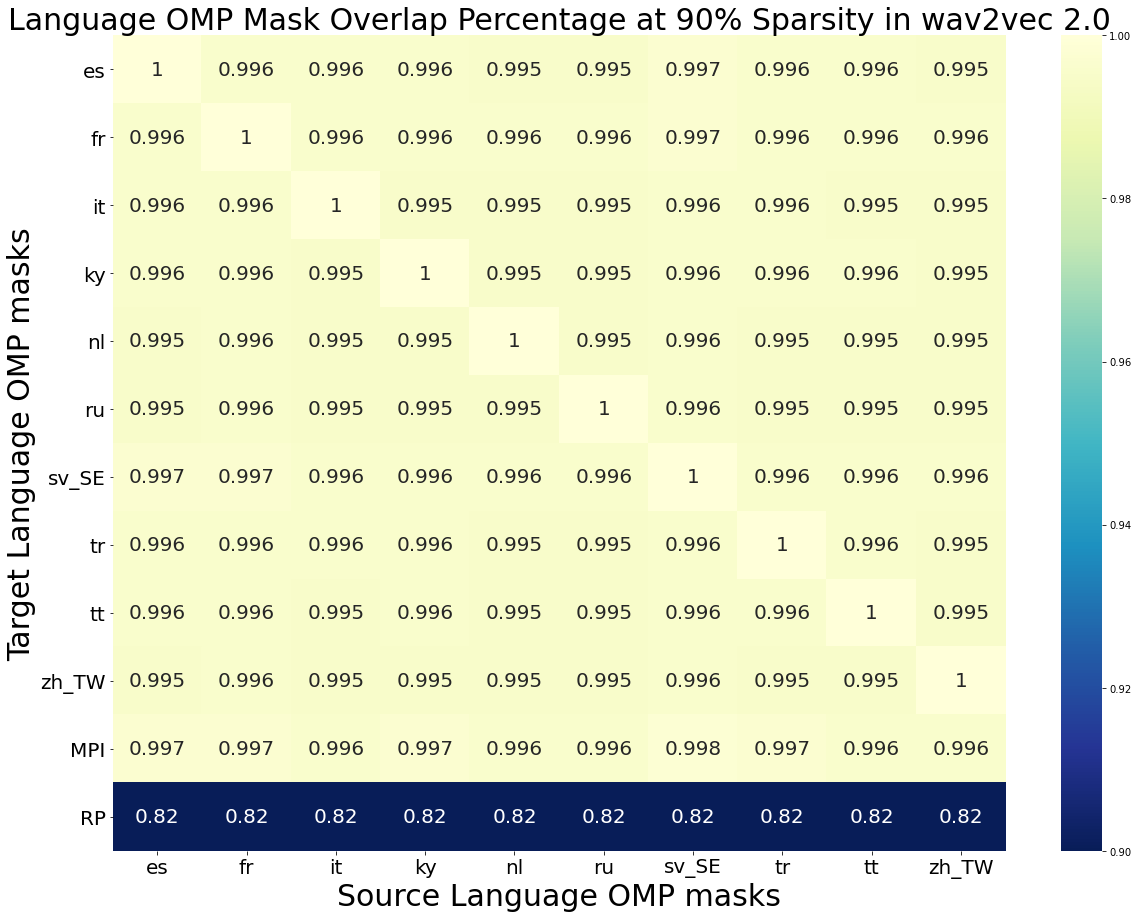}
        \centering
        \caption{
        {\tt OMP} pruning masks {\tt IOU}s and overlap percentages on finetuned {\tt wav2vec2} at 90\% sparsity.}
        \end{figure*}    

    \newpage~\newpage~\newpage~\newpage
    
    \vspace{-3mm}
    \subsection{{\tt OMP} Masks Overlap in CSR}
    \label{app:mask_overlap_matrices-xlsr}
    \textbf{CSR {\tt OMP} masks overlap procedure.} Each set of experiments require 10$\times$10 rounds of {\tt xlsr} finetunings because there are 10 downstream spoken languages ASR. 
    The experimental procedure is: 
    \begin{enumerate}
        \item Finetune {\tt xlsr} for a source spoken language ASR. 
        \item Prune the finetuned model and obtain an {\tt OMP} mask for each spoken language ASR.
        \item Calculate {\tt IOU}/mask overlap over all pairs of spoken language masks at each sparsity.
    \end{enumerate}
    
        \begin{figure*} [!h]
        \includegraphics[width=0.47\linewidth]{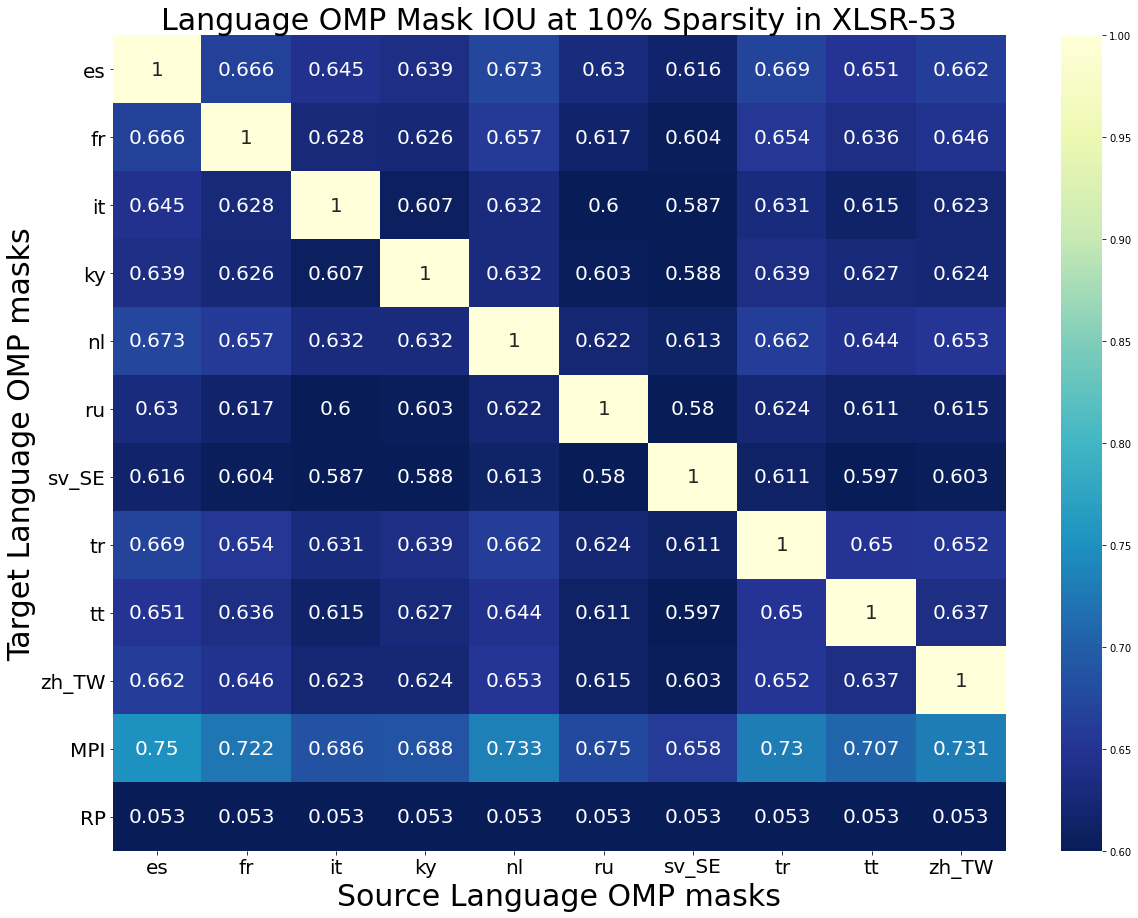}
        \hspace{.5cm}
        \includegraphics[width=0.47\linewidth]{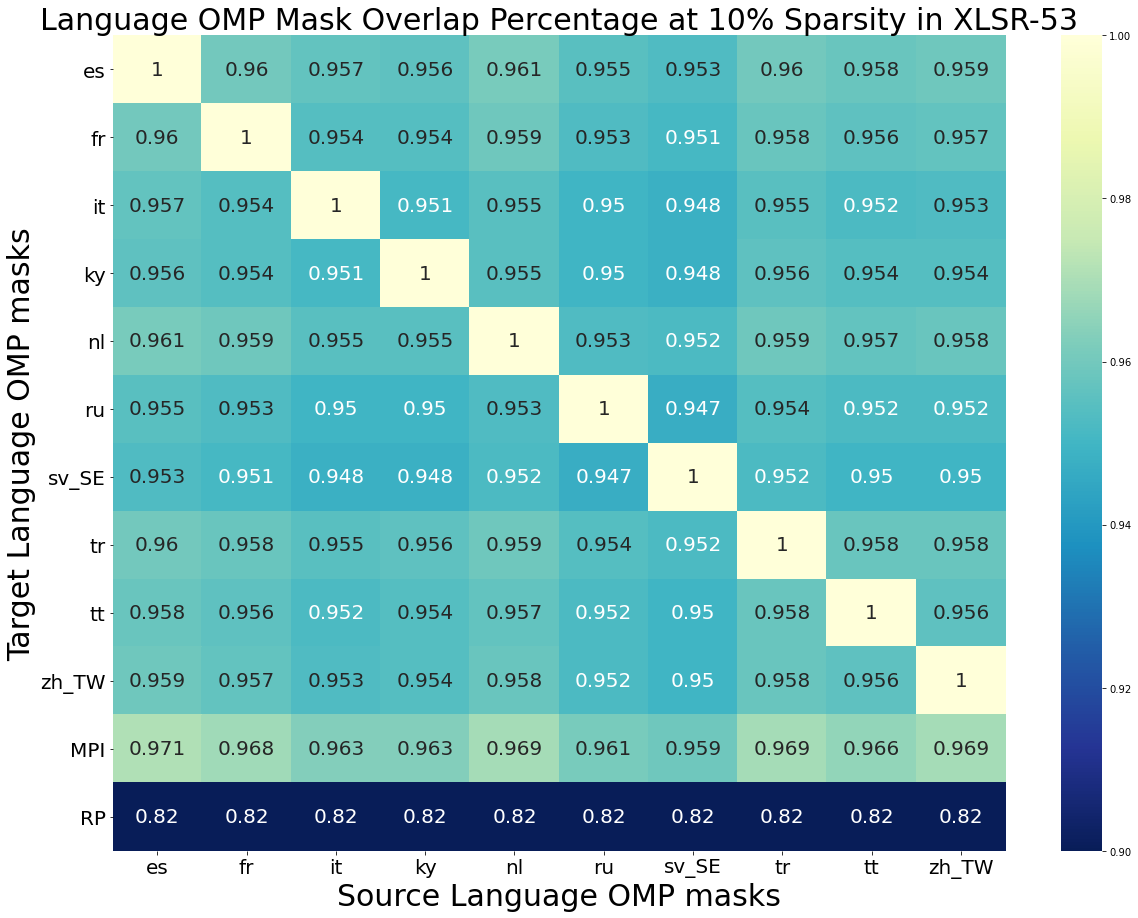}
        \centering
        \caption{
        {\tt OMP} pruning masks {\tt IOU}s and overlap percentages on finetuned {\tt xlsr} at 10\% sparsity.}
        \end{figure*}
        
        \begin{figure*} [h]
        \includegraphics[width=0.47\linewidth]{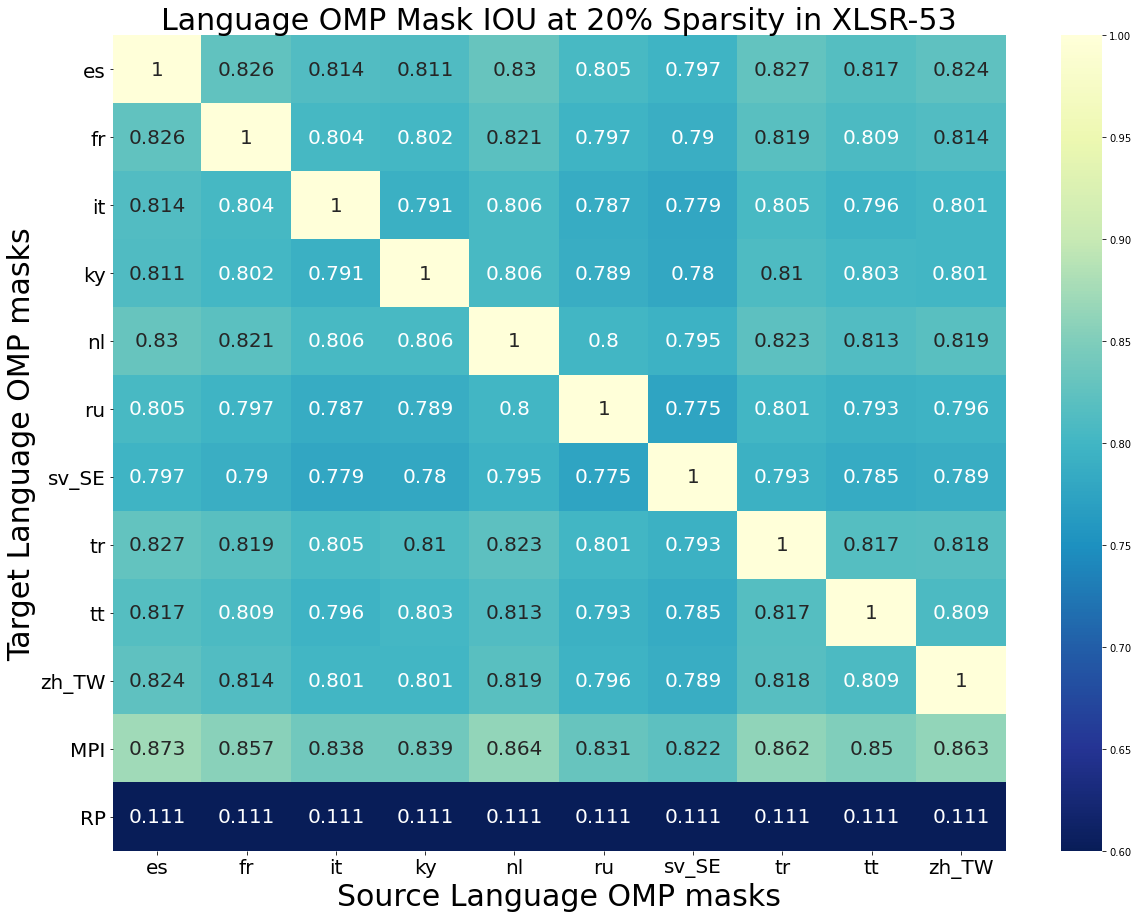}
        \hspace{.5cm}
        \includegraphics[width=0.47\linewidth]{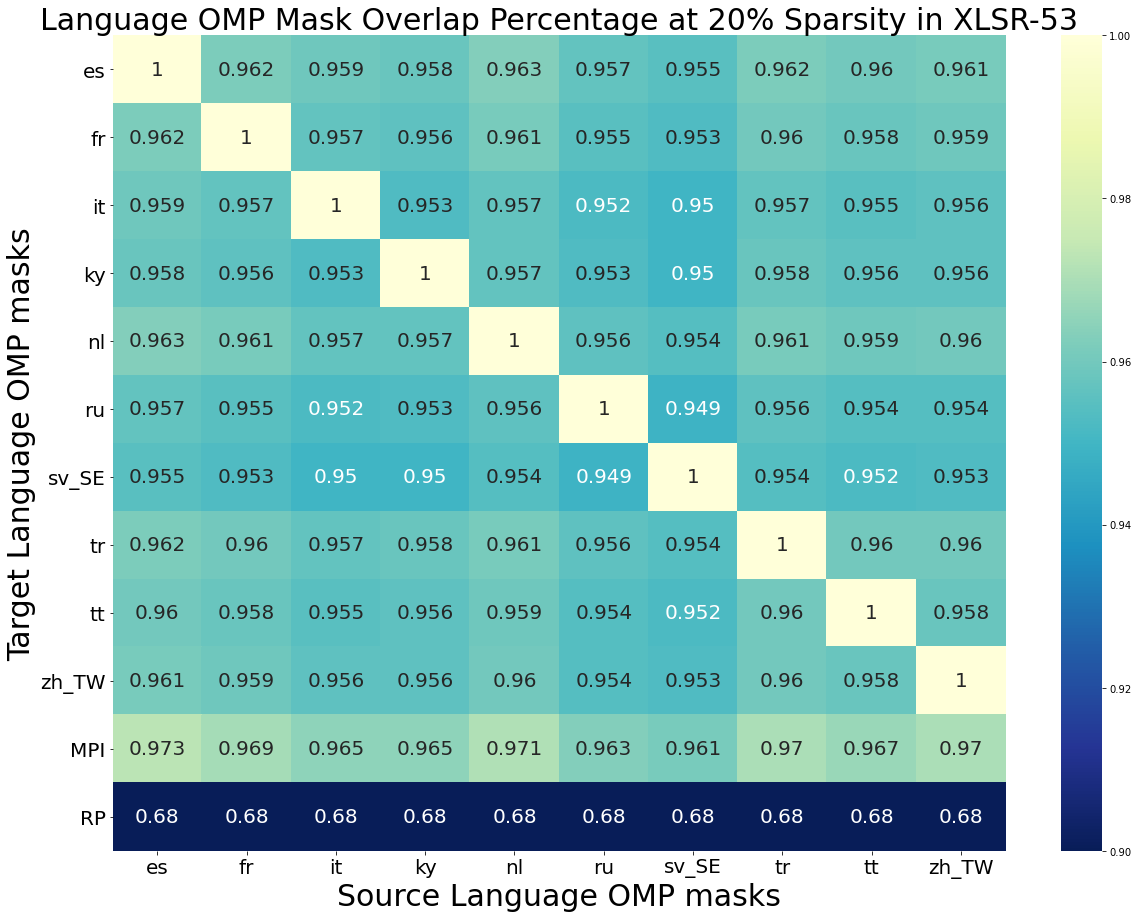}
        \centering
        \caption{
        {\tt OMP} pruning masks {\tt IOU}s and overlap percentages on finetuned {\tt xlsr} at 20\% sparsity.}
        \end{figure*}
        
        \begin{figure*} [!h]
        \includegraphics[width=0.47\linewidth]{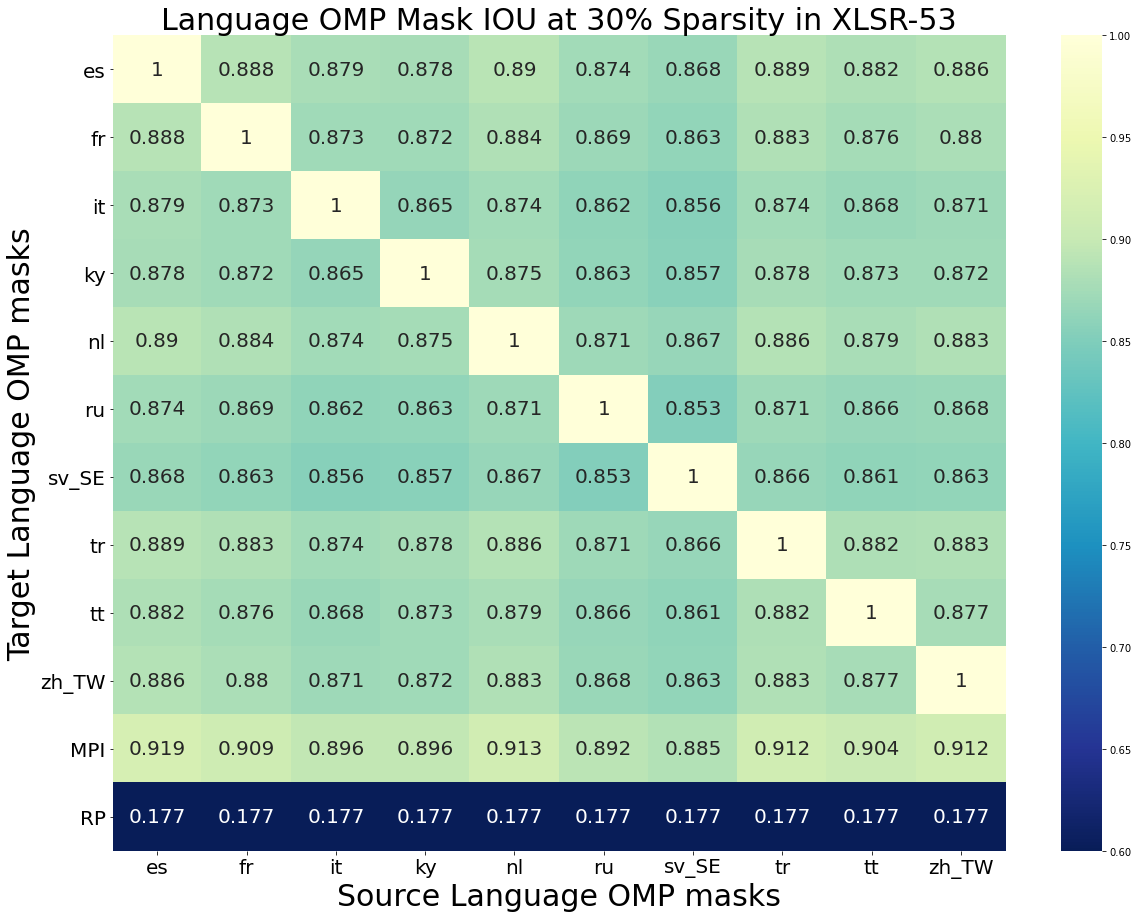}
        \hspace{.5cm}
        \includegraphics[width=0.47\linewidth]{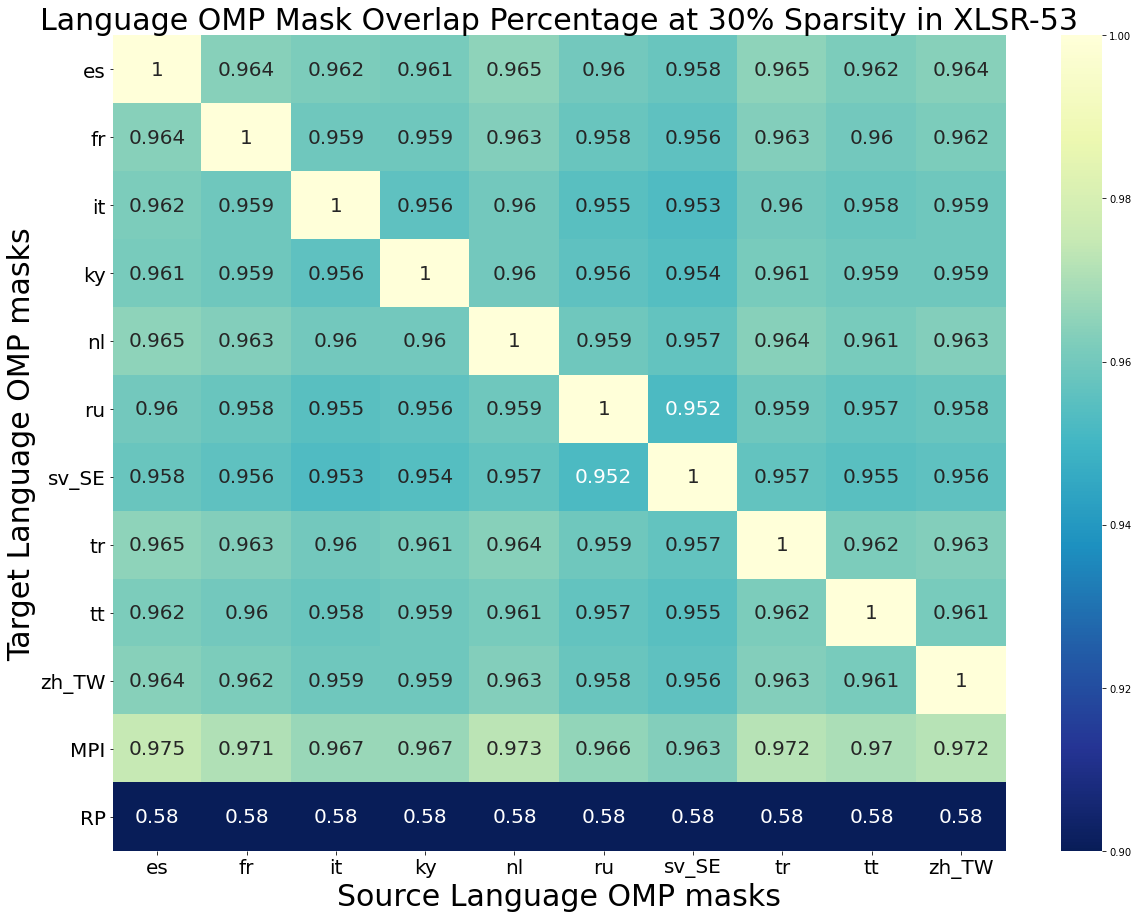}
        \centering
        \caption{
        {\tt OMP} pruning masks {\tt IOU}s and overlap percentages on finetuned {\tt xlsr} at 30\% sparsity.}
        \end{figure*}
        
        \begin{figure*} [!h]
        \includegraphics[width=0.47\linewidth]{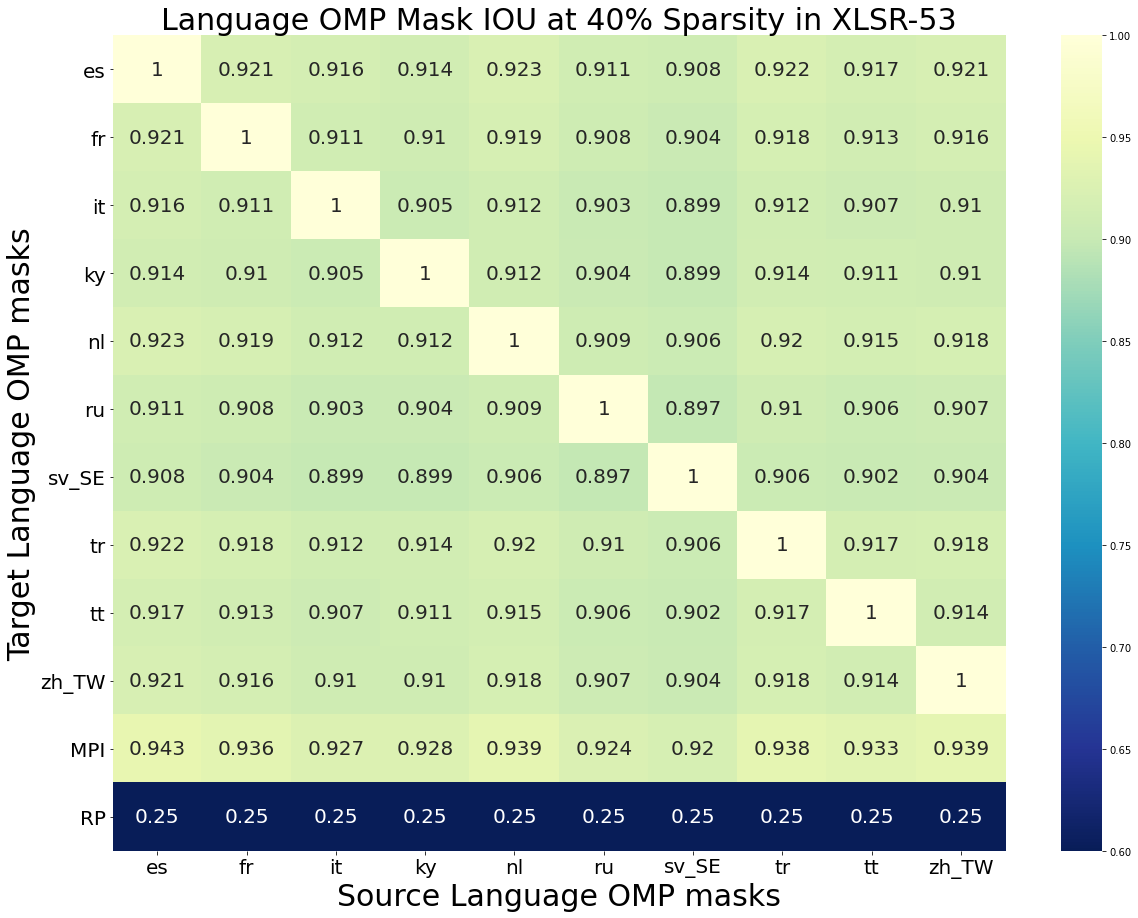}
        \hspace{.5cm}
        \includegraphics[width=0.47\linewidth]{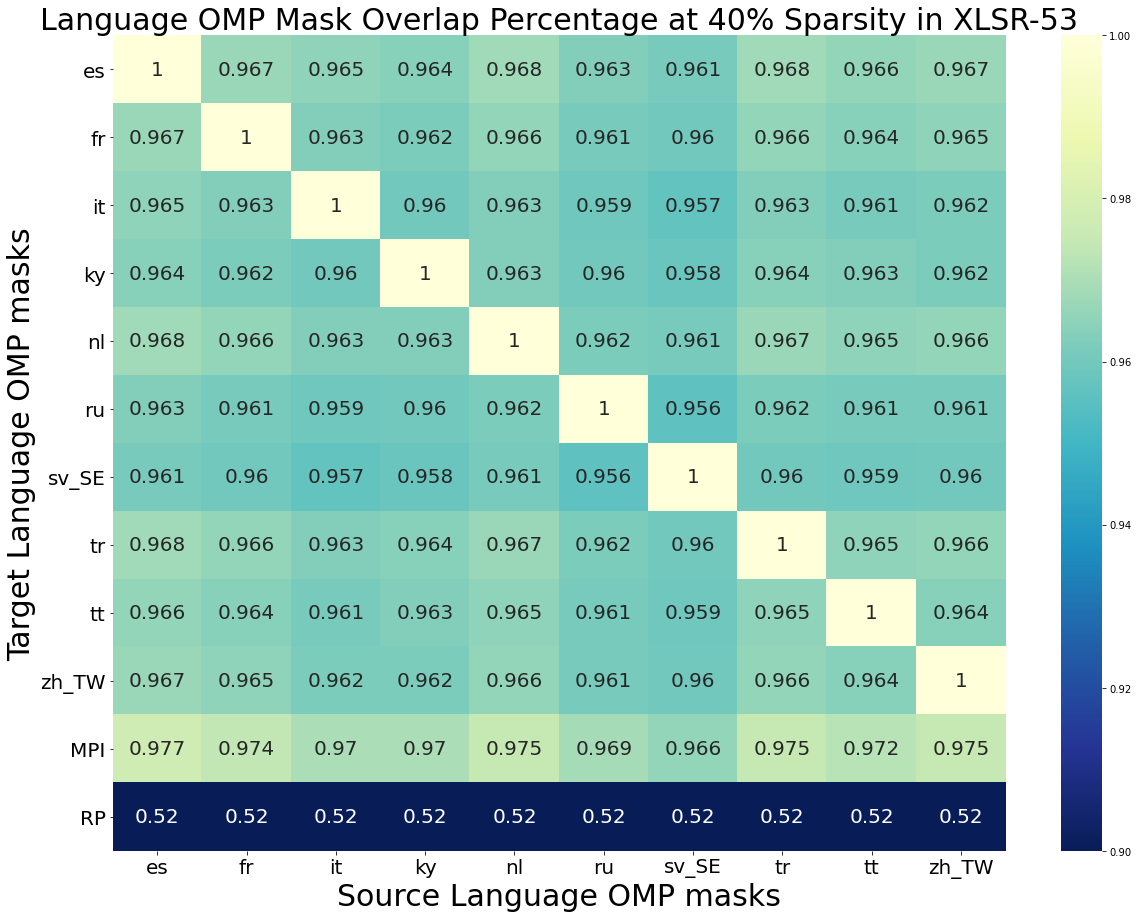}
        \centering
        \caption{
        {\tt OMP} pruning masks {\tt IOU}s and overlap percentages on finetuned {\tt xlsr} at 40\% sparsity.}
        \end{figure*}
        
        \begin{figure*} [!h]
        \includegraphics[width=0.47\linewidth]{figs/mask-overlap/xlsr_mask_IOU_50.png}
        \hspace{.5cm}
        \includegraphics[width=0.47\linewidth]{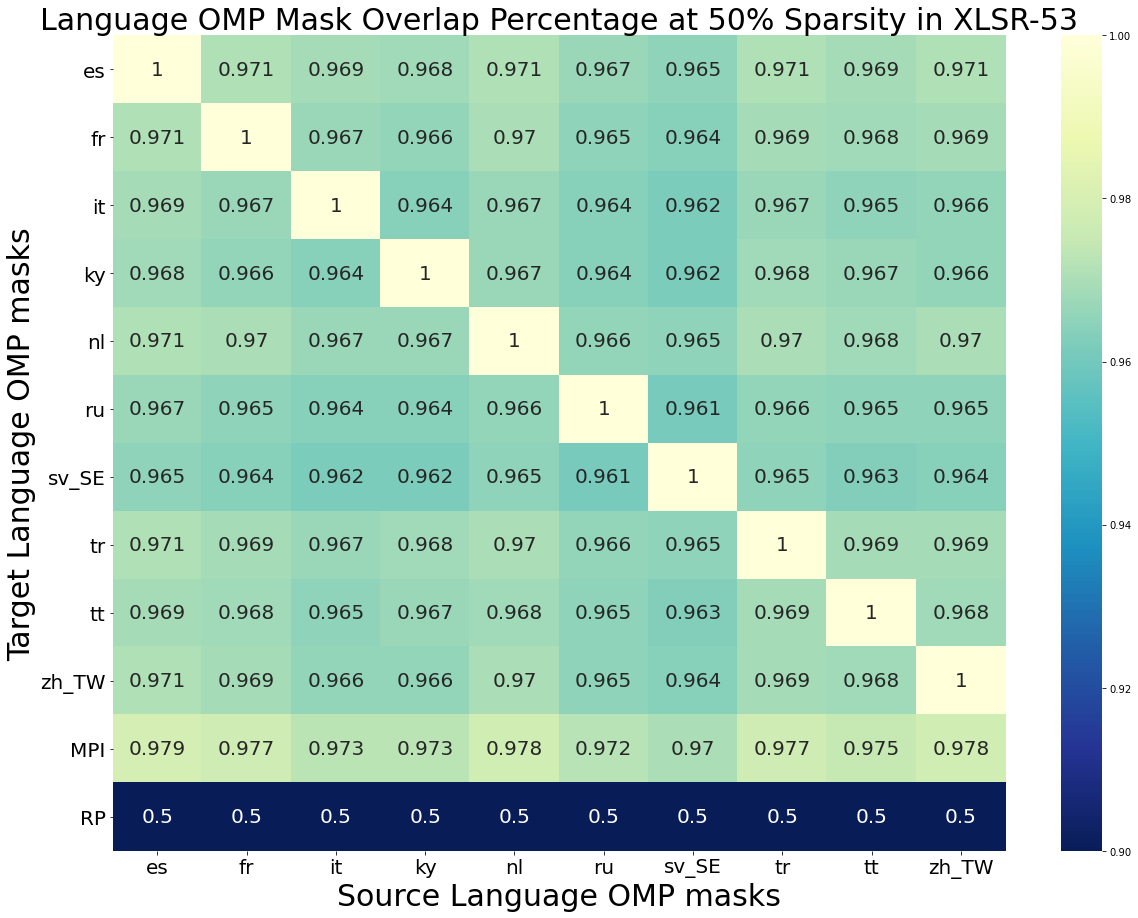}
        \centering
        \caption{
        {\tt OMP} pruning masks {\tt IOU}s and overlap percentages on finetuned {\tt xlsr} at 50\% sparsity.}
        \end{figure*}
        
        \begin{figure*} [!h]
        \includegraphics[width=0.47\linewidth]{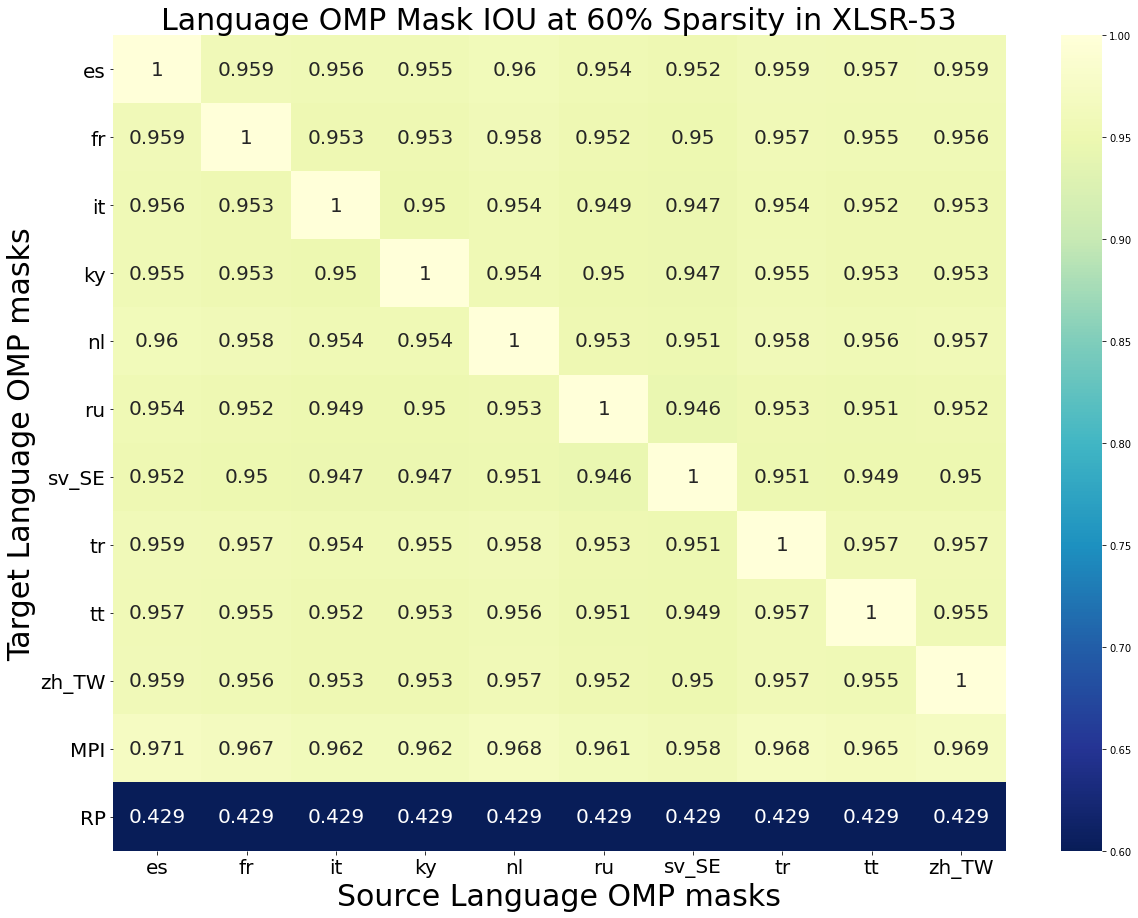}
        \hspace{.5cm}
        \includegraphics[width=0.47\linewidth]{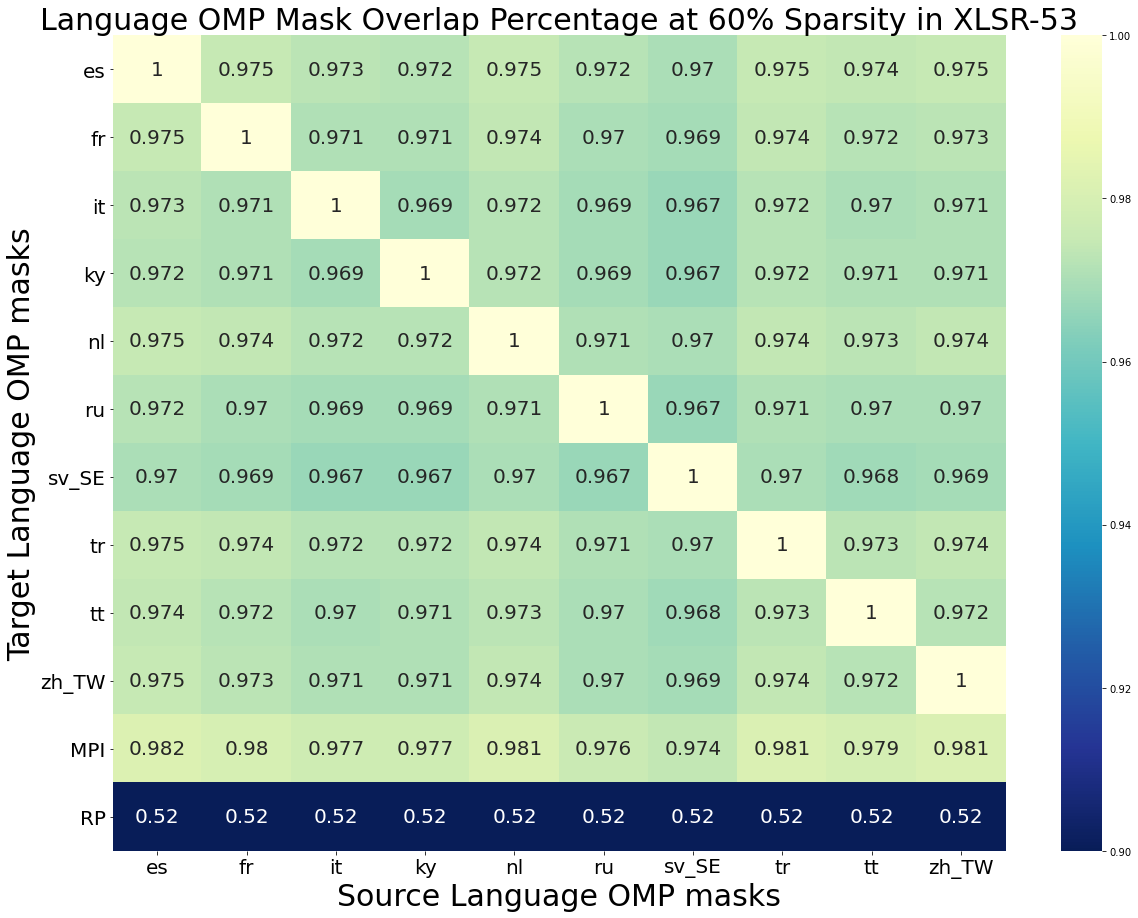}
        \centering
        \caption{
        {\tt OMP} pruning masks {\tt IOU}s and overlap percentages on finetuned {\tt xlsr} at 60\% sparsity.}
        \end{figure*}
        
        \begin{figure*} [!h]
        \includegraphics[width=0.47\linewidth]{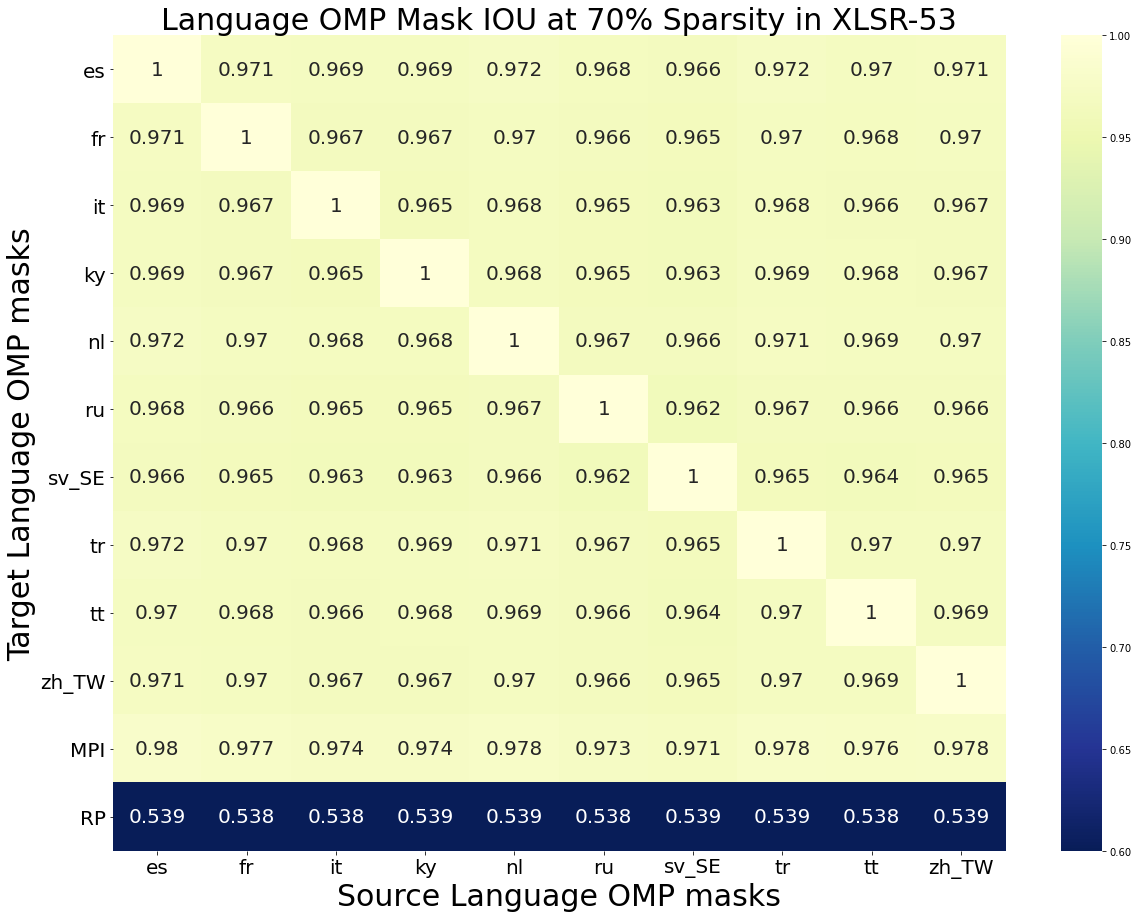}
        \hspace{.5cm}
        \includegraphics[width=0.47\linewidth]{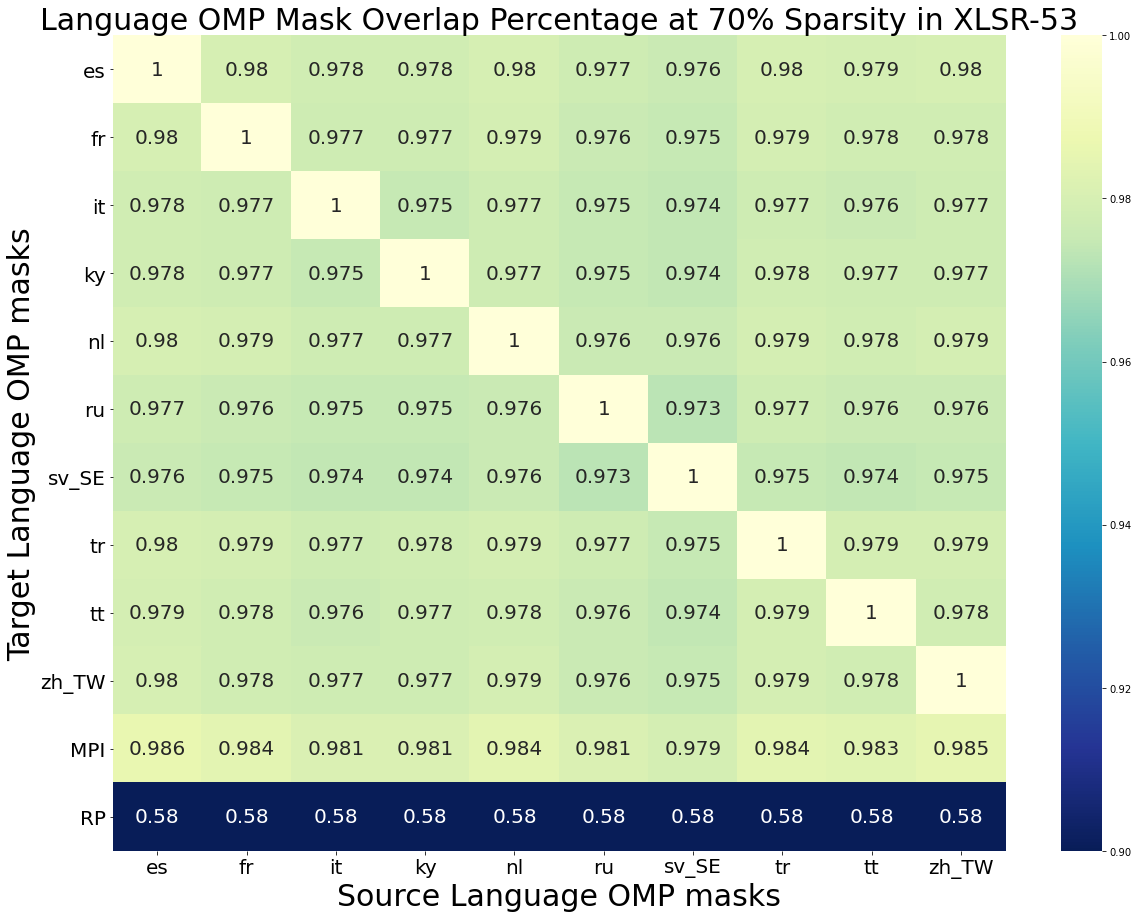}
        \centering
        \caption{
        {\tt OMP} pruning masks {\tt IOU}s and overlap percentages on finetuned {\tt xlsr} at 70\% sparsity.}
        \end{figure*}
        
        \begin{figure*} [!h]
        \includegraphics[width=0.47\linewidth]{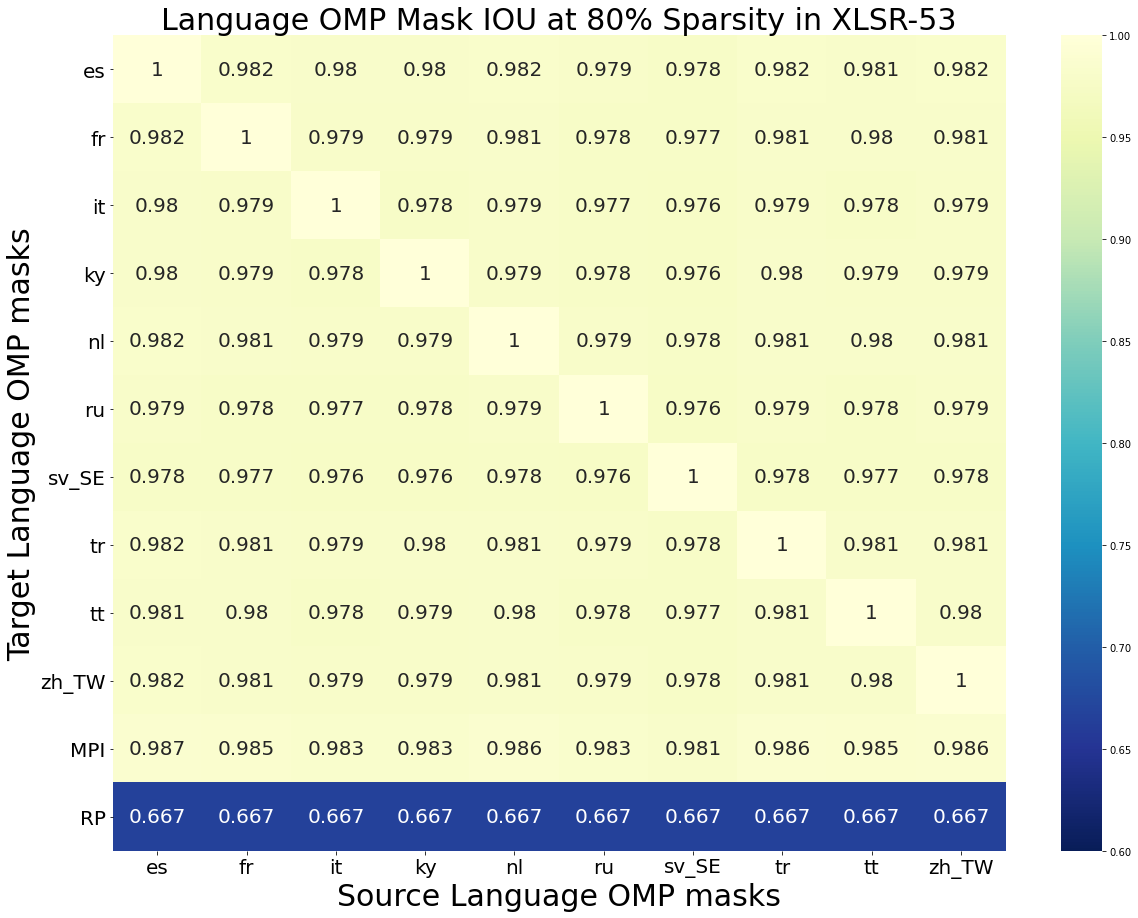}
        \hspace{.5cm}
        \includegraphics[width=0.47\linewidth]{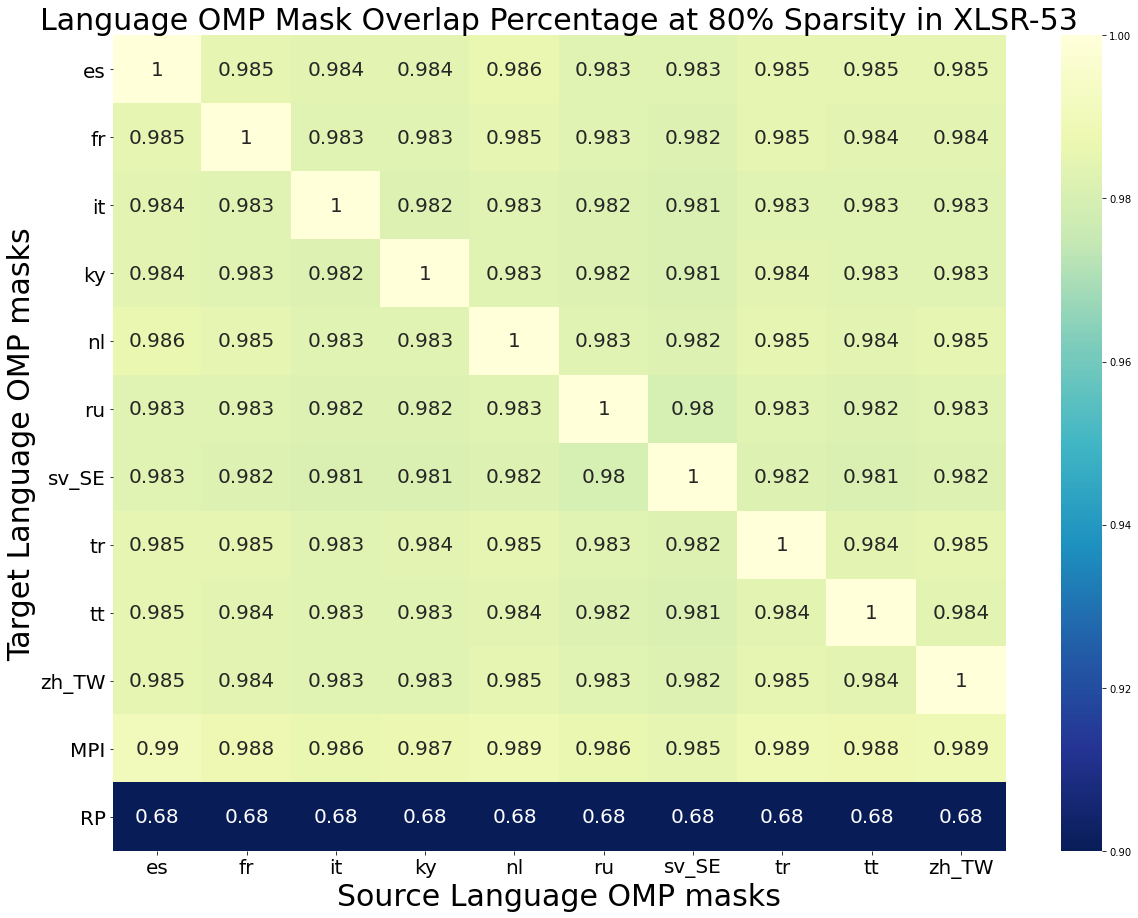}
        \centering
        \caption{
        {\tt OMP} pruning masks {\tt IOU}s and overlap percentages on finetuned {\tt xlsr} at 80\% sparsity.}
        \end{figure*}
        
        \begin{figure*} [!h]
        \includegraphics[width=0.47\linewidth]{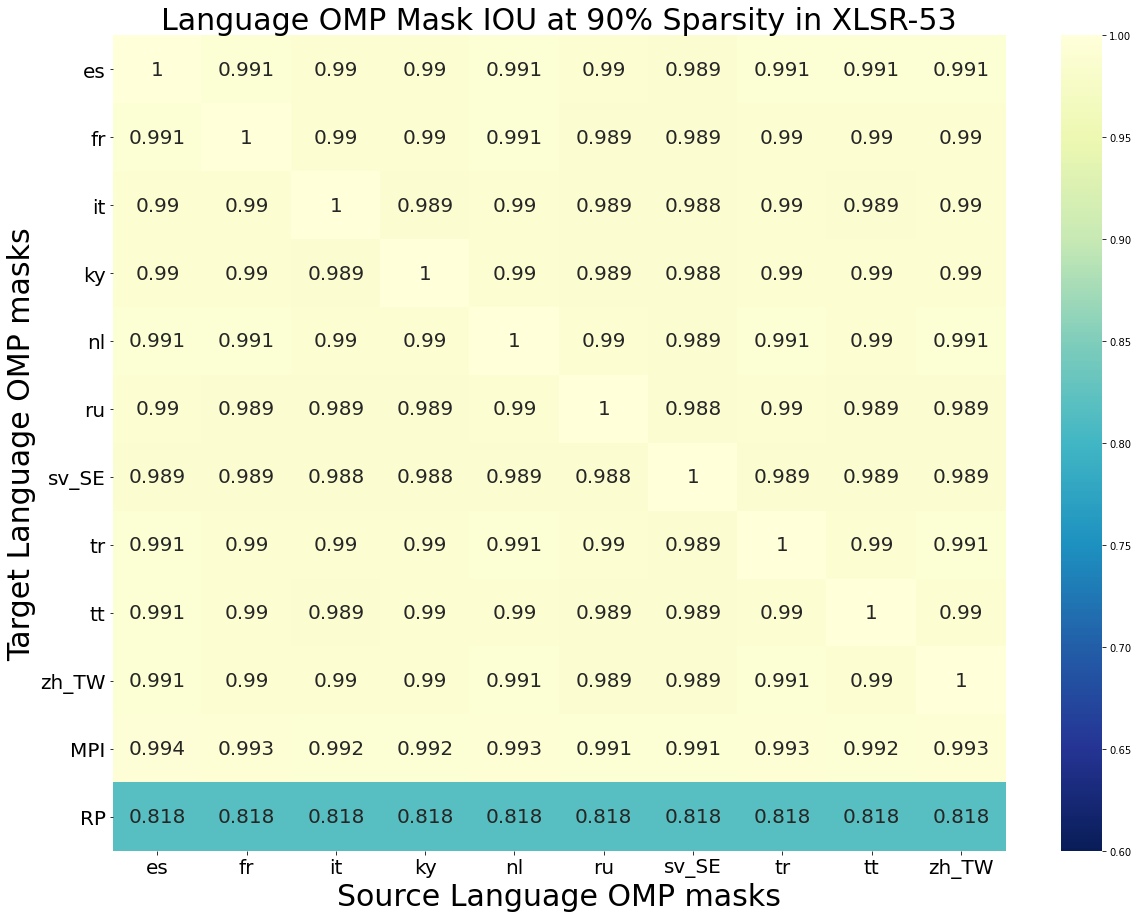}
        \hspace{.5cm}
        \includegraphics[width=0.47\linewidth]{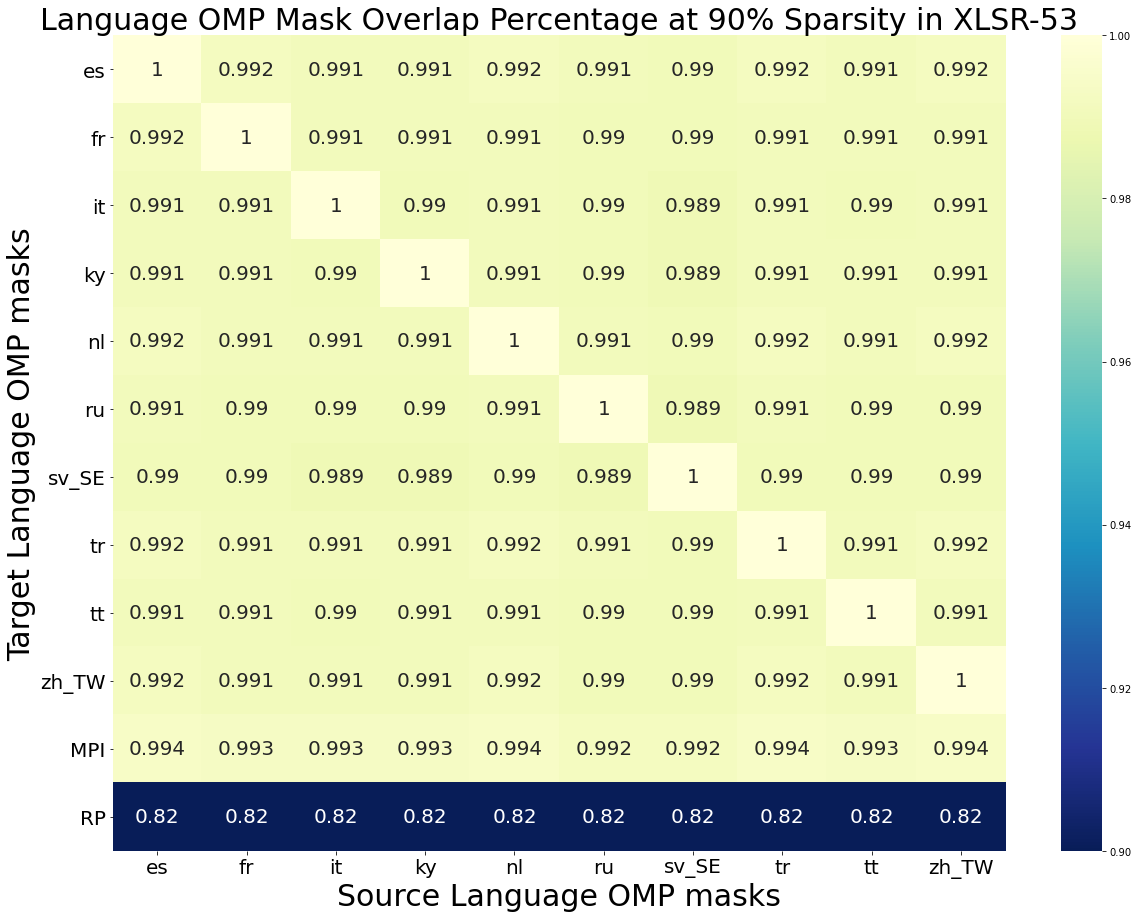}
        \centering
        \caption{
        {\tt OMP} pruning masks {\tt IOU}s and overlap percentages on finetuned {\tt xlsr} at 90\% sparsity.}
        \end{figure*}  

\newpage~\newpage~\newpage~\newpage


\section{{\tt xlsr} Cross-Lingual Mask Transfer}
\label{app:xlsr_mask_transfer}
\textbf{Cross-lingual mask transfer procedure.} Each set of experiments require 10$\times$10$\times$2 rounds of {\tt xlsr} finetunings because there are 10 downstream spoken languages ASR, and we finetune for each spoken language ASR twice (the first one for retrieving mask, and second one for mask transfer). 
The experimental procedure is: 
\begin{enumerate}
    \item Finetune {\tt xlsr}/{\tt wav2vec2} for a source spoken language ASR. 
    \item Prune the finetuned model and obtain an {\tt OMP} mask for each spoken language ASR.
    \item Apply the {\tt OMP} mask at {\tt xlsr} pre-trained initializations and finetune for a target spoken language ASR with {\tt PARP}. 
\end{enumerate}

\begin{figure*} [!hbtp]
    \includegraphics[width=\linewidth]{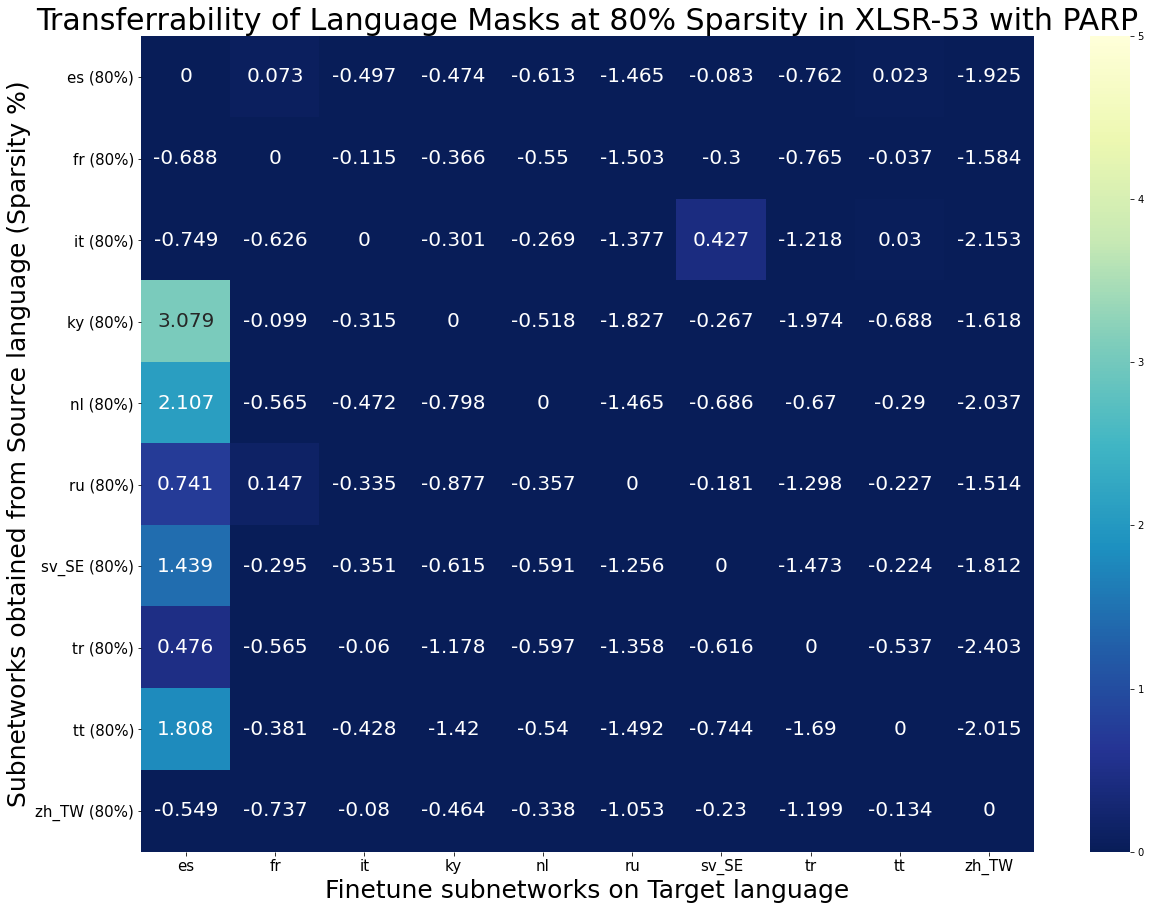}
    \centering
    \caption{Cross-lingual mask transfer for {\tt xlsr}. Cross-lingual mask transfer with {\tt PARP} has minimal PER degradation (darker the better).}
    \label{fig:xlsr_mask_transfer}
    \end{figure*}

Figure~\ref{fig:xlsr_mask_transfer} is the result, and it has the same cross-lingual mask transfer setup as that in Section~\ref{subsec:mask_transfer} and Figure~\ref{fig:wav2vec2_mask_transfer}, except the pre-trained model is {\tt xlsr} instead of {\tt wav2vec2}.

\newpage

\section{Details of Task Transfer Results on Pre-trained BERT}
\label{app:bert_task_transfer}
\textbf{Cross-task mask transfer procedure.} Each set of experiments require 9$\times$9$\times$2 rounds of finetunings because there are 9 subtasks in GLUE, and we finetune for each subtask twice (the first one for retrieving mask, and second one for mask transfer).
We first note that our cross-task transfer experimental designs are closely knitted to NLP probing work’s experimental setup~\cite{wu2020similarity,dalvi2020analyzing}, i.e. pretrained BERT/XLNet on 9 subtasks in GLUE. 
The experimental procedure is: 
\begin{enumerate}
    \item Finetune BERT/XLNet for a source task in GLUE. 
    \item Prune the finetuned model and obtain an {\tt IMP} mask for each task.
    \item Apply the {\tt IMP} mask at BERT/XLNet pre-trained initializations and finetune for a target task in GLUE with {\tt PARP}. 
\end{enumerate}

\begin{figure*} [!h]
    \includegraphics[width=\linewidth]{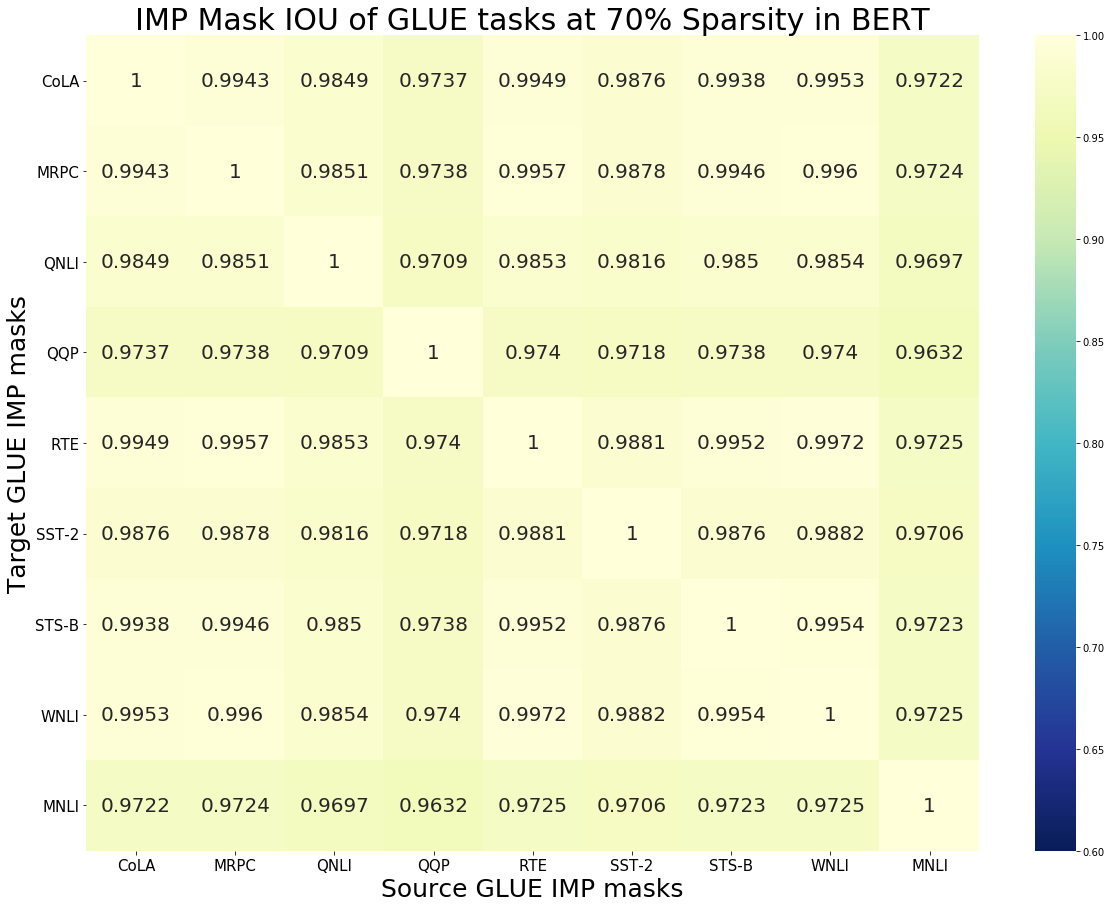}
    \centering
    \caption{
    {\tt IOU}s over all GLUE tasks' {\tt IMP} pruning masks on finetuned BERT at 70\% sparsity. 
    Notice the high overlap rates, which aligns with Observation~\ref{observation:similarity}.
    }
    \label{fig:glue_iou}
    \end{figure*}

\begin{figure*} [!hbtp]
    \includegraphics[width=0.8\linewidth]{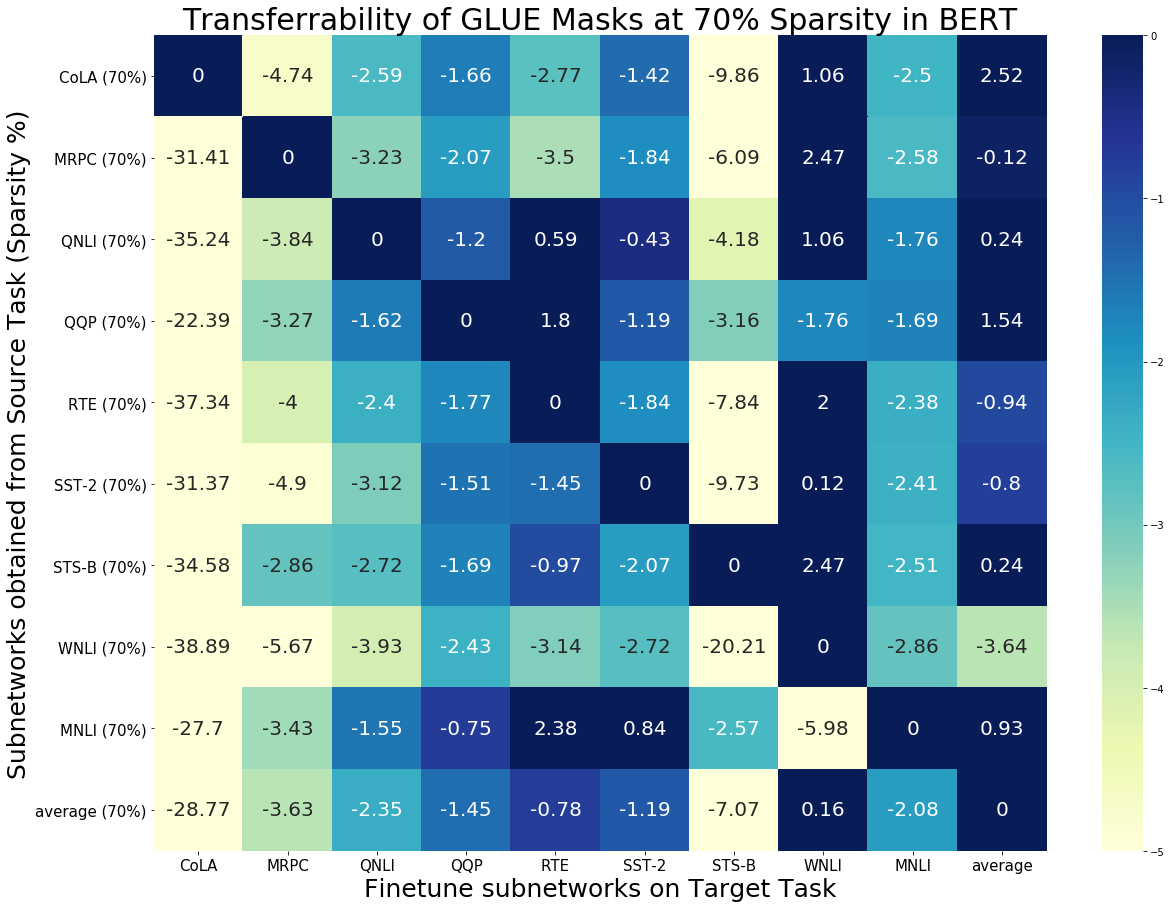}
    \hspace{5cm}
    \includegraphics[width=0.8\linewidth]{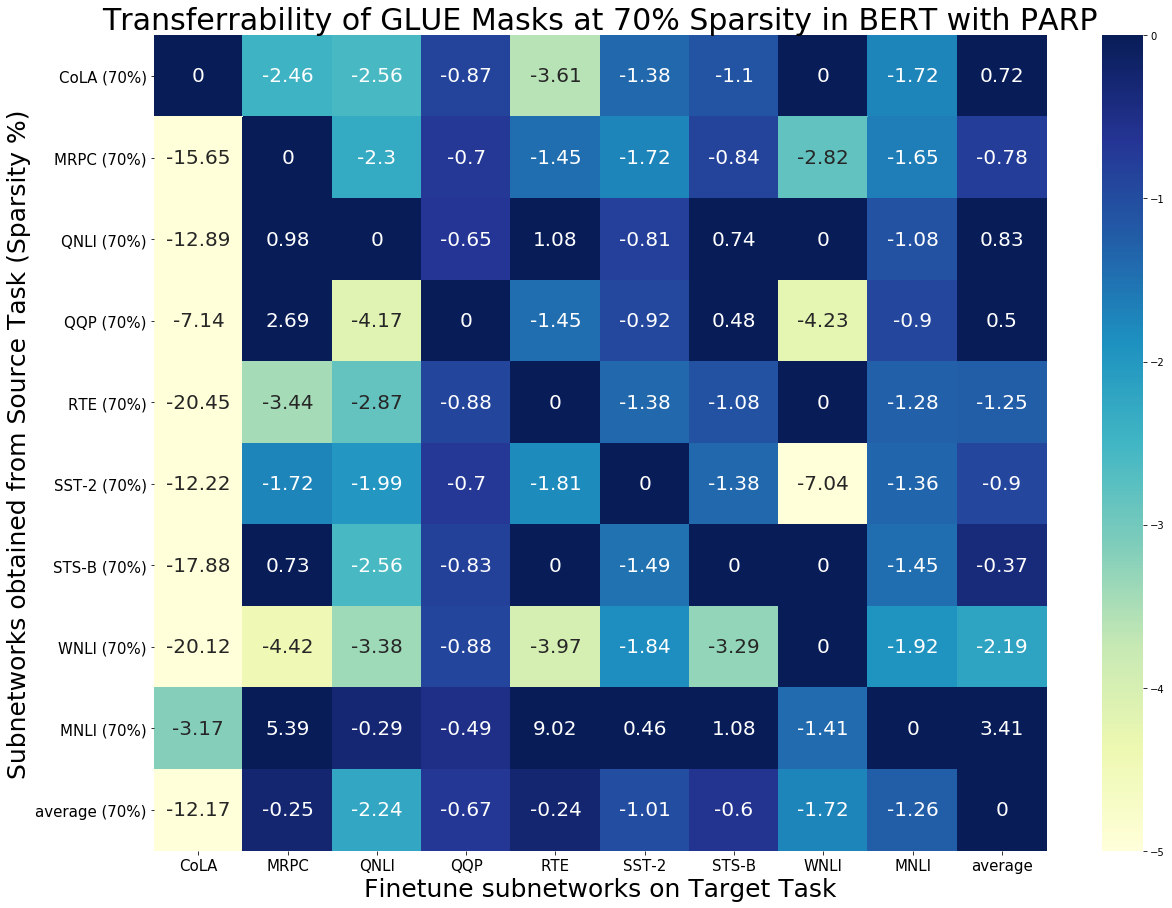}
    \centering
    \caption{
    Results for subnetwork transfer experiment (take subnetwork found by {\tt IMP} at task A and finetune it for task B). 
    \textbf{Top:} the transfer results in BERT-Ticket~\cite{chen2020lottery}. 
    \textbf{Bottom:} transfer with {\tt PARP} finetuning instead. Each row is a source task A, and each column is a target task B.
    All numbers are subtracted by the scores of same-task transfer (task A = task B, and the darker the better). 
    }
    \label{fig:glue}
    \end{figure*}

Figure~\ref{fig:glue_iou} is the {\tt IMP} mask overlap for pre-trained BERT on the 9 natural language tasks in GLUE. 
Figure~\ref{fig:glue} is the cross-task transfer result.
For all the GLUE tasks, {\tt PARP} can achieve better results compared to BERT-Ticket (cross-task subnetwork regular finetuning)~\cite{chen2020lottery}. 
For the tasks with poor transferability in BERT-Ticket~\cite{chen2020lottery}, like CoLA and STS-B, {\tt PARP} still achieves good transfer scores.

~\newpage

\section{Full H2L and CSR Pruning Results}
    We provide the full set of H2L and CSR pruning (refer to Section~\ref{subsec:exp_main_result} and Section~\ref{subsec:joint} for experimental description). 
    Below are the rest of Figure~\ref{fig:multilin_per_nl} to other spoken languages from CommonVoice: \textit{Spanish (es), French (fr), Italian (it), Kyrgyz (ky), Dutch (nl), Russian (ru), Swedish (sv-SE), Turkish(tr), Tatar (tt), and Mandarin (zh-TW)}

    \begin{figure*} [!hbtp]
    \includegraphics[width=\linewidth]{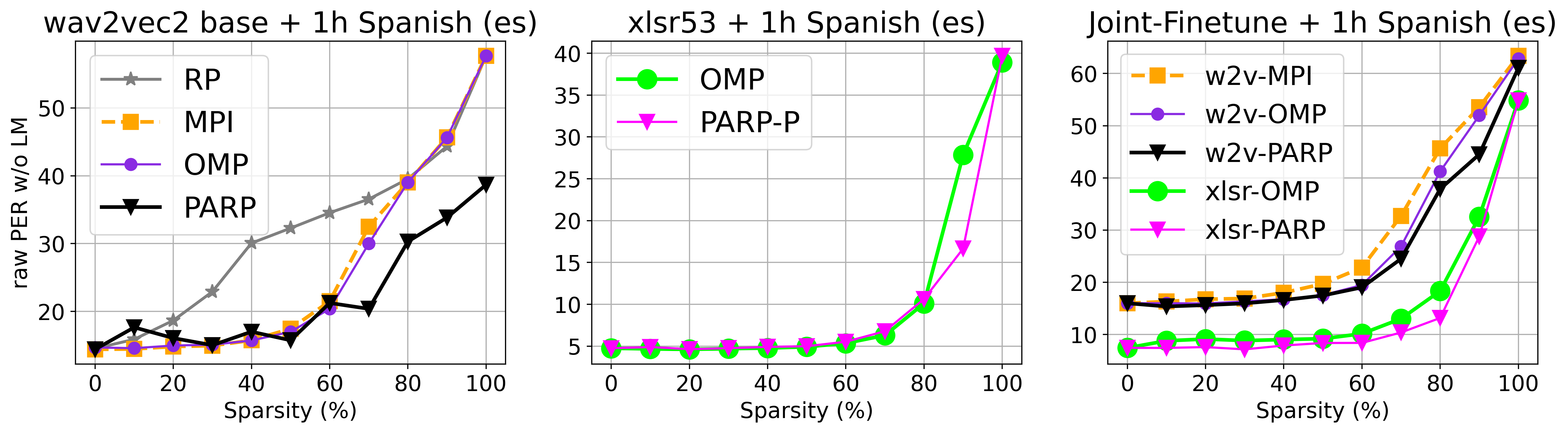}
    \centering
    \caption{Comparison of pruning techniques on H2L \& CSR with 1h of Spanish (\textit{es}) ASR finetuning. \textbf{(Left)} Pruning H2L ({\tt wav2vec2-base} + \textit{es}). \textbf{(Center)} Pruning CSR ({\tt xlsr} + \textit{es}). \textbf{(Right)} Pruning jointly-finetuned {\tt wav2vec2-base} and {\tt xlsr} on \textit{es}.}
    \label{fig:multilin_per_es}
    \end{figure*}
    \begin{figure*} [!h]
    \includegraphics[width=\linewidth]{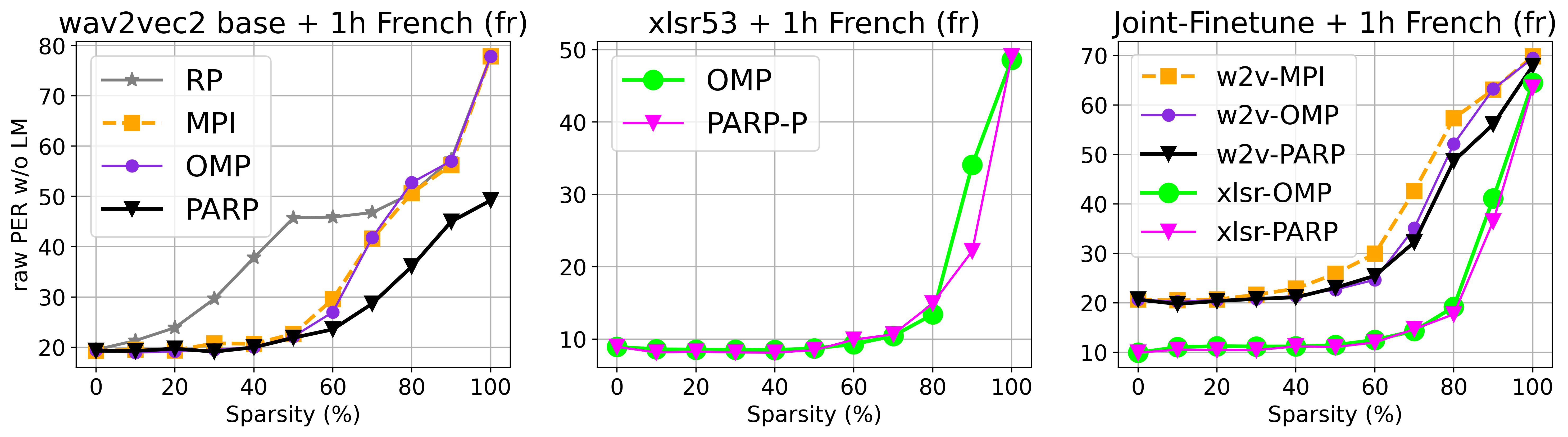}
    \centering
    \caption{Comparison of pruning techniques on H2L \& CSR with 1h of French (\textit{fr}) ASR finetuning. \textbf{(Left)} Pruning H2L ({\tt wav2vec2-base} + \textit{fr}). \textbf{(Center)} Pruning CSR ({\tt xlsr} + \textit{fr}). \textbf{(Right)} Pruning jointly-finetuned {\tt wav2vec2-base} and {\tt xlsr} on \textit{fr}.}
    \label{fig:multilin_per_fr}
    \end{figure*}
    \begin{figure*} [!h]
    \includegraphics[width=\linewidth]{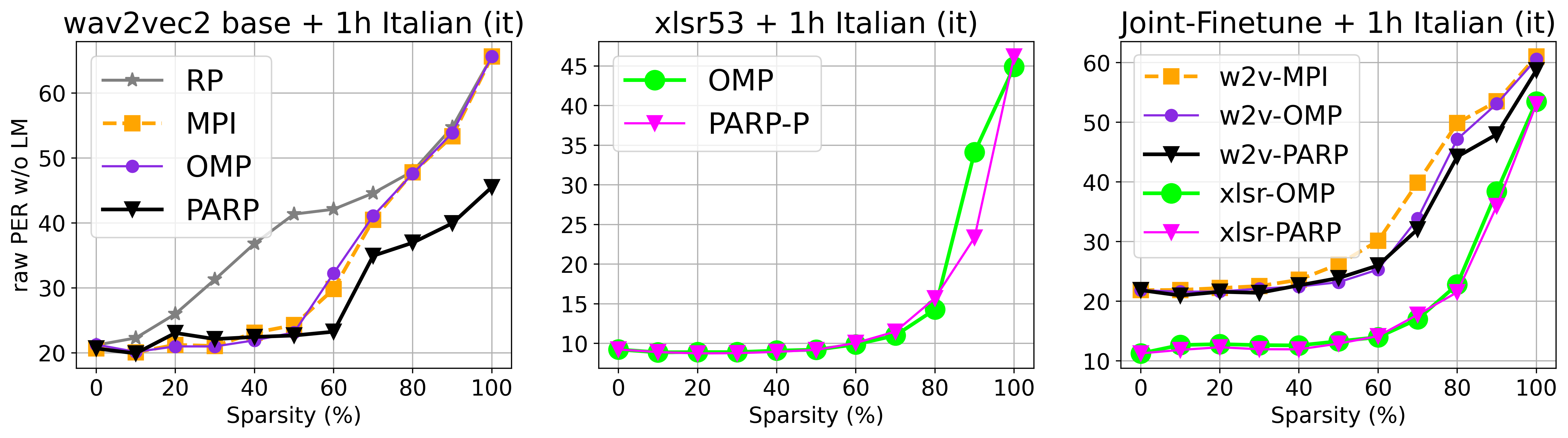}
    \centering
    \caption{Comparison of pruning techniques on H2L \& CSR with 1h of Italian (\textit{it}) ASR finetuning. \textbf{(Left)} Pruning H2L ({\tt wav2vec2-base} + \textit{it}). \textbf{(Center)} Pruning CSR ({\tt xlsr} + \textit{it}). \textbf{(Right)} Pruning jointly-finetuned {\tt wav2vec2-base} and {\tt xlsr} on \textit{it}.}
    \label{fig:multilin_per_it}
    \end{figure*}
    \begin{figure*} [!h]
    \includegraphics[width=\linewidth]{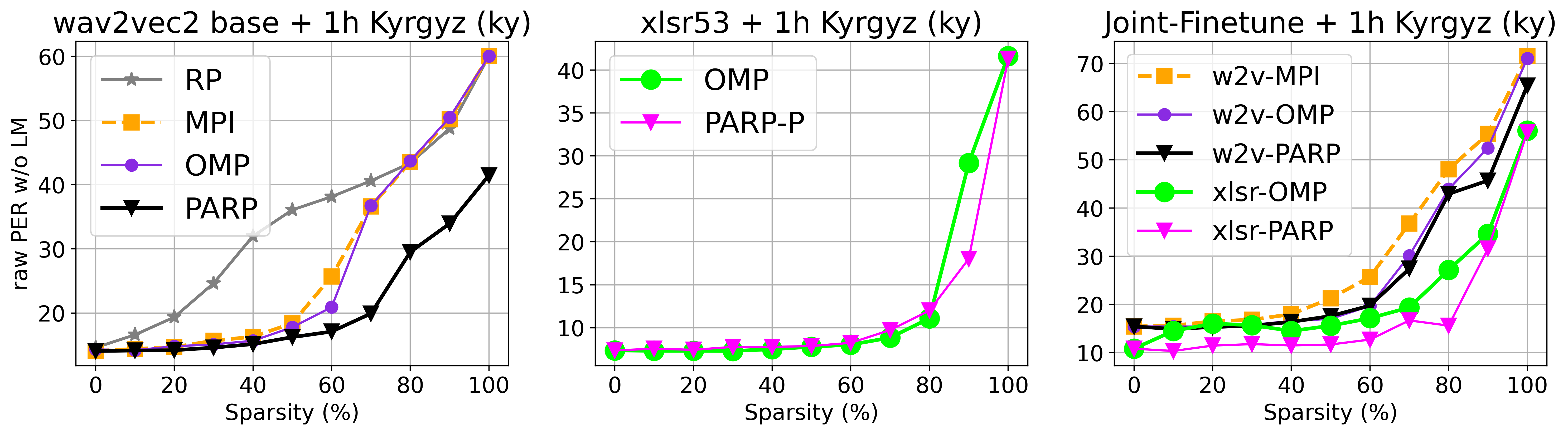}
    \centering
    \caption{Comparison of pruning techniques on H2L \& CSR with 1h of Kyrgyz (\textit{ky}) ASR finetuning. \textbf{(Left)} Pruning H2L ({\tt wav2vec2-base} + \textit{ky}). \textbf{(Center)} Pruning CSR ({\tt xlsr} + \textit{ky}). \textbf{(Right)} Pruning jointly-finetuned {\tt wav2vec2-base} and {\tt xlsr} on \textit{ky}.}
    \label{fig:multilin_per_ky}
    \end{figure*}
    \begin{figure*} [!h]
    \includegraphics[width=\linewidth]{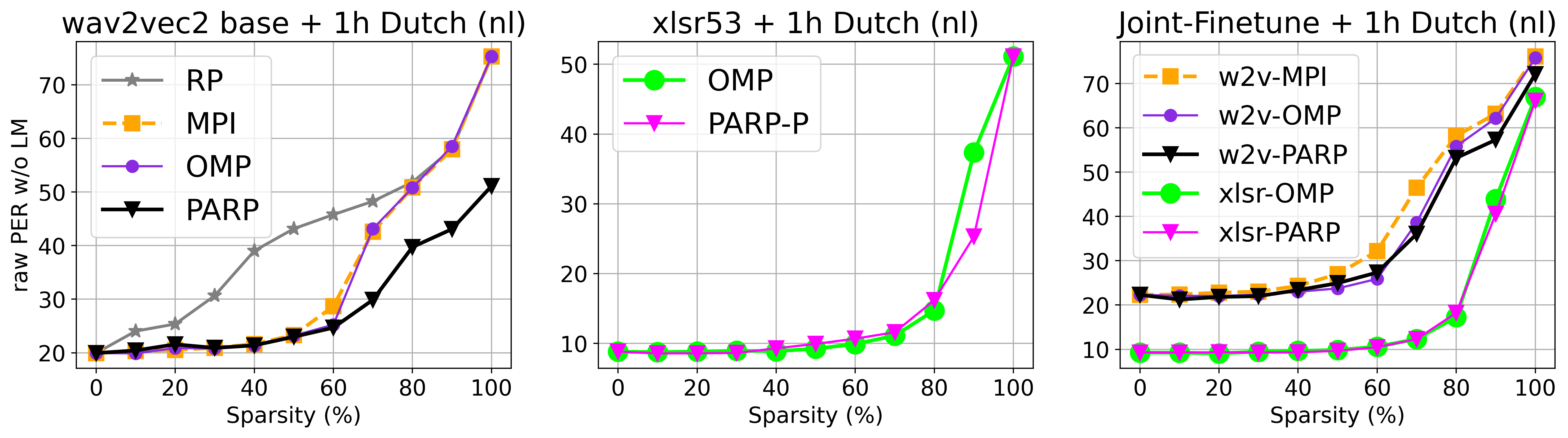}
    \centering
    \caption{Comparison of pruning techniques on H2L \& CSR with 1h of Dutch (\textit{nl}) ASR finetuning. \textbf{(Left)} Pruning H2L ({\tt wav2vec2-base} + \textit{nl}). \textbf{(Center)} Pruning CSR ({\tt xlsr} + \textit{nl}). \textbf{(Right)} Pruning jointly-finetuned {\tt wav2vec2-base} and {\tt xlsr} on \textit{nl}.}
    \label{fig:multilin_per_nl2}
    \end{figure*}
    \begin{figure*} [!h]
    \includegraphics[width=\linewidth]{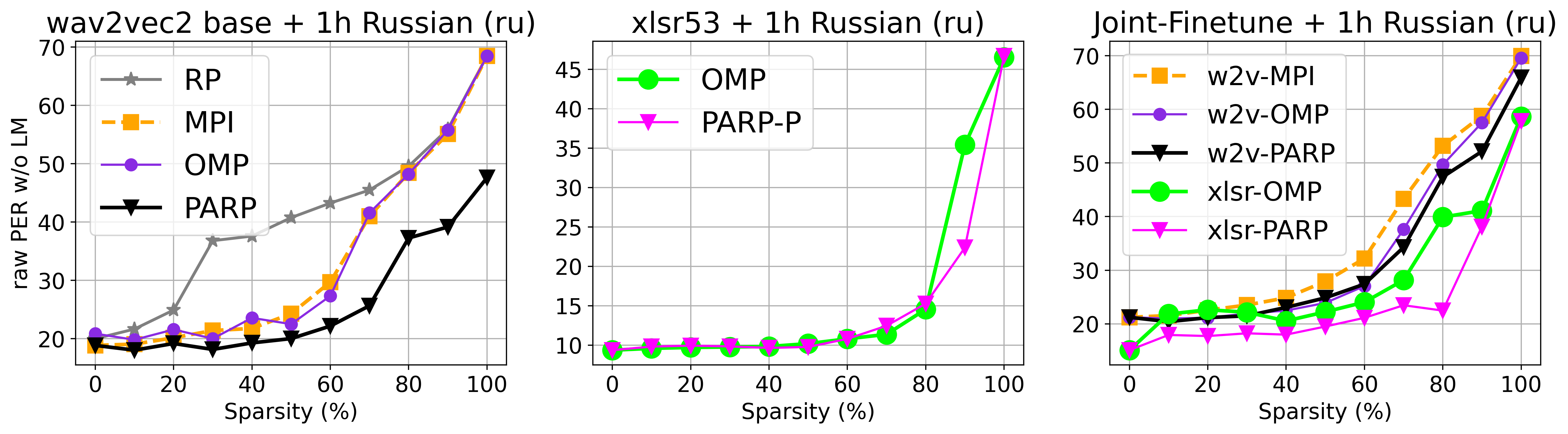}
    \centering
    \caption{Comparison of pruning techniques on H2L \& CSR with 1h of Russian (\textit{ru}) ASR finetuning. \textbf{(Left)} Pruning H2L ({\tt wav2vec2-base} + \textit{ru}). \textbf{(Center)} Pruning CSR ({\tt xlsr} + \textit{ru}). \textbf{(Right)} Pruning jointly-finetuned {\tt wav2vec2-base} and {\tt xlsr} on \textit{ru}.}
    \label{fig:multilin_per_ru}
    \end{figure*}
    \begin{figure*} [!h]
    \includegraphics[width=\linewidth]{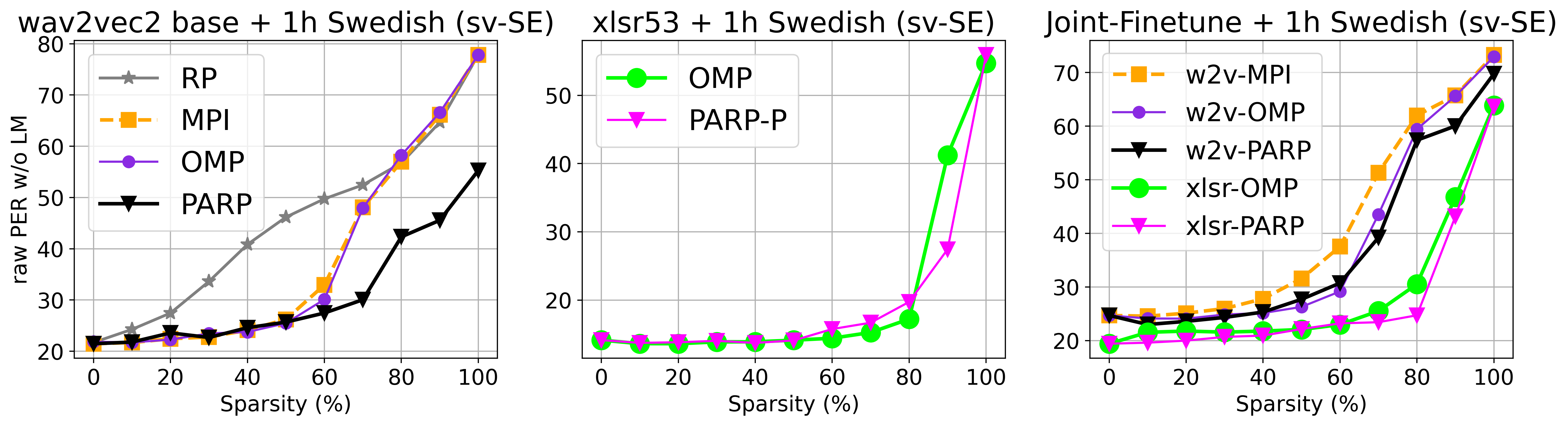}
    \centering
    \caption{Comparison of pruning techniques on H2L \& CSR with 1h of Swedish (\textit{sv-SE}) ASR finetuning. \textbf{(Left)} Pruning H2L ({\tt wav2vec2-base} + \textit{sv-SE}). \textbf{(Center)} Pruning CSR ({\tt xlsr} + \textit{sv-SE}). \textbf{(Right)} Pruning jointly-finetuned {\tt wav2vec2-base} and {\tt xlsr} on \textit{sv-SE}.}
    \label{fig:multilin_per_sv-SE}
    \end{figure*}
    \begin{figure*} [!h]
    \includegraphics[width=\linewidth]{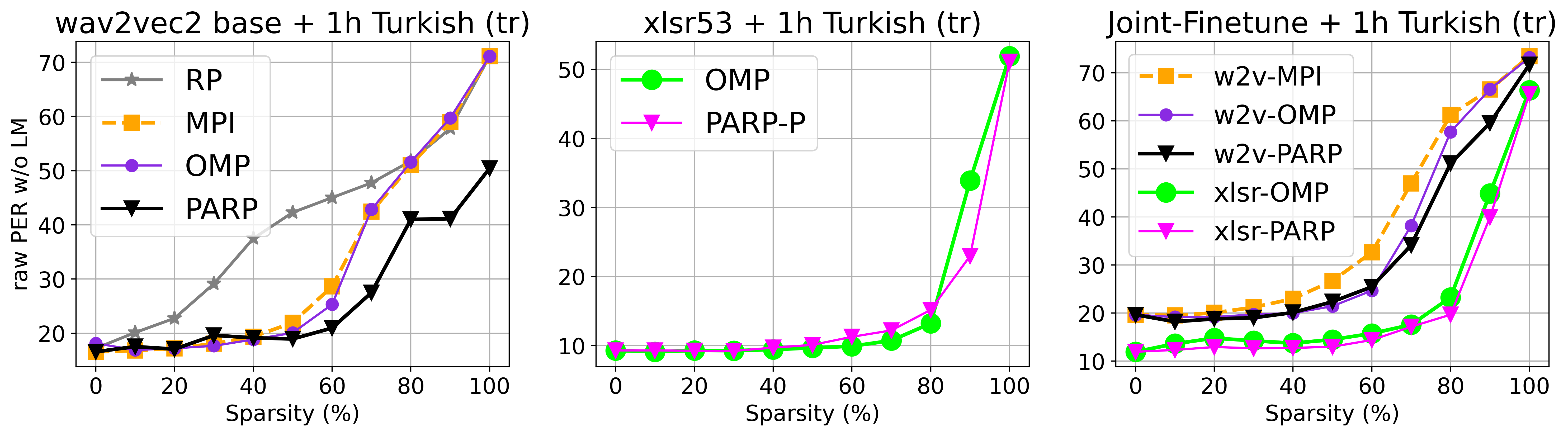}
    \centering
    \caption{Comparison of pruning techniques on H2L \& CSR with 1h of Turkish (\textit{tr}) ASR finetuning. \textbf{(Left)} Pruning H2L ({\tt wav2vec2-base} + \textit{tr}). \textbf{(Center)} Pruning CSR ({\tt xlsr} + \textit{tr}). \textbf{(Right)} Pruning jointly-finetuned {\tt wav2vec2-base} and {\tt xlsr} on \textit{tr}.}
    \label{fig:multilin_per_tr}
    \end{figure*}
    \begin{figure*} [!h]
    \includegraphics[width=\linewidth]{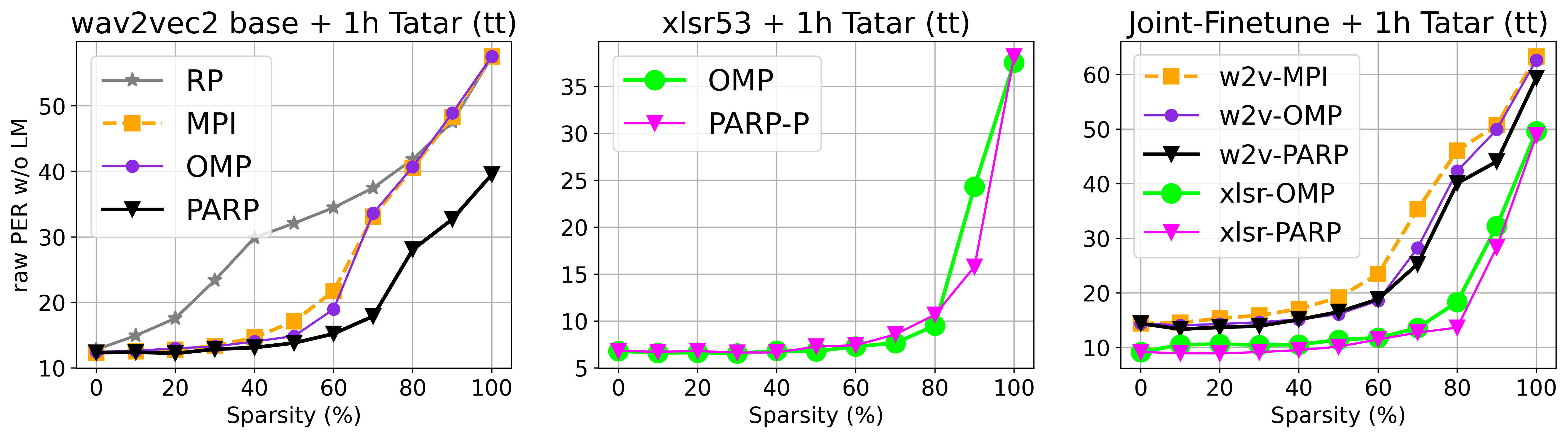}
    \centering
    \caption{Comparison of pruning techniques on H2L \& CSR with 1h of Tatar (\textit{tt}) ASR finetuning. \textbf{(Left)} Pruning H2L ({\tt wav2vec2-base} + \textit{tt}). \textbf{(Center)} Pruning CSR ({\tt xlsr} + \textit{tt}). \textbf{(Right)} Pruning jointly-finetuned {\tt wav2vec2-base} and {\tt xlsr} on \textit{tt}.}
    \label{fig:multilin_per_tt}
    \end{figure*}
    \begin{figure*} [!h]
    \includegraphics[width=\linewidth]{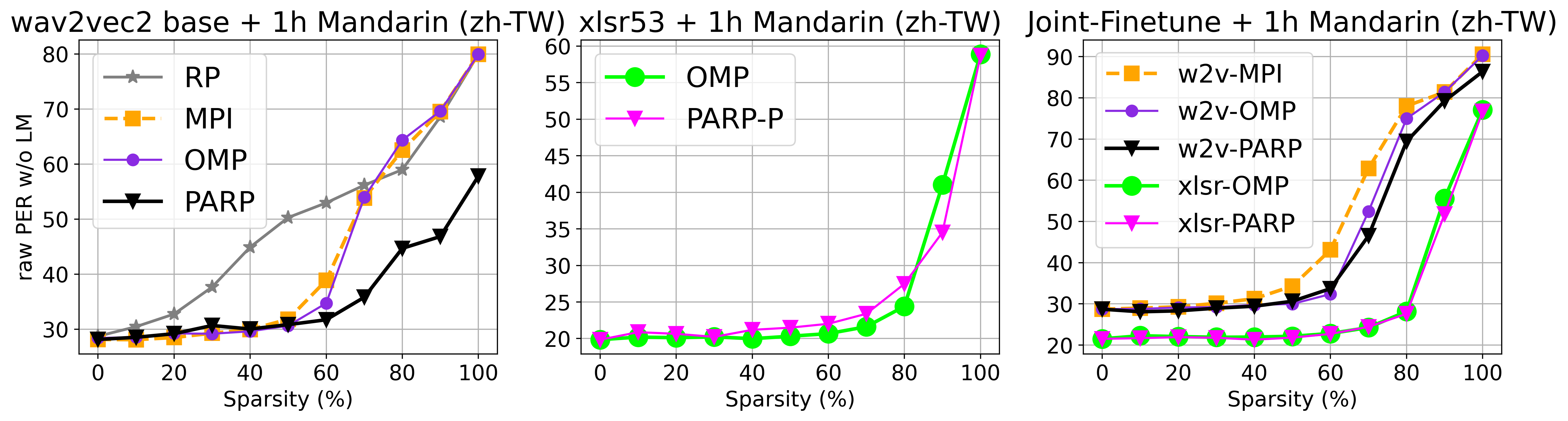}
    \centering
    \caption{Comparison of pruning techniques on H2L \& CSR with 1h of Mandarin (\textit{zh-TW}) ASR finetuning. \textbf{(Left)} Pruning H2L ({\tt wav2vec2-base} + \textit{zh-TW}). \textbf{(Center)} Pruning CSR ({\tt xlsr} + \textit{zh-TW}). \textbf{(Right)} Pruning jointly-finetuned {\tt wav2vec2-base} and {\tt xlsr} on \textit{zh-TW}.}
    \label{fig:multilin_per_zh-TW}
    \end{figure*}

\newpage~\newpage~\newpage~\newpage


\section{{\tt wav2vec2} + {\tt PARP} with Random Seeds and LM Decoding}
\label{app:lm_decode_parp}
We re-iterate the two reasons why we did not including LM decoding in our main results.
First, we isolate the effect of pruning on ASR. 
Note that the standard LM (either 4-gram/transformer) used in the wav2vec series are also trained on Librispeech (text-corpus)~\cite{baevski2020wav2vec,baevski2021unsupervised}. 
Therefore, the LMs can easily recover errors made by the acoustic model. 
Secondly, note that the 4-gram/transformer LM decoding hyper-parameters are carefully searched via Bayesian optimization\footnote{\url{https://github.com/facebook/Ax}} in the wav2vec series. 
Such optimization procedure would be quite expensive to run for \textbf{just one} model, let alone thousands of pruned models produced in this work. 

We provide two sets of results to validate our claim that applying {\tt PARP} on {\tt wav2vec2} reduces the downstream ASR:
\vspace{-2mm}
\begin{itemize}
    \item The first result is the impact of random seeds.
    We finetune {\tt wav2vec2-base} with 10min data at 10\% sparsity with {\tt PARP} at 8 additional seeds. 
    Table~\ref{tab:random_seed} is the result with Viterbi decoding without LM. 
    We can see that at different seed values, pruned {\tt wav2vec2-base} all converged to similar WERs, which is $\sim$10\% WER reductions compared to a full {\tt wav2vec2-base}. 
    
    \item The second result is pruned {\tt wav2vec2-base} with the official 4-gram/transformer LM decoding. 
    The pruned {\tt wav2vec-base} is finetuned on the 10min Librispeech split and pruned at 10\% sparsity with {\tt PARP}.
    Since we do not have the compute resource to replicate 1500 beam 4-gram decoding and 500 beam transformer-LM decoding used in the original paper~\cite{baevski2020wav2vec}, this experiment is based on a more moderate beam size. 
    Similar to~\cite{baevski2020wav2vec}, decoding hyper-parameters are searched via Ax on the dev-other Librispeech subset over 128 trials.  
    As shown in Table~\ref{tab:parp_lm}, the performance gain over the full {\tt wav2vec2-base} reduces with LM decoding, but we still observe a performance improvement at 10\% sparsity with {\tt PARP}.
\end{itemize}  

\begin{table*}[!h]
    \small
    \caption{Pruning {\tt wav2vec-base} with {\tt PARP} at different trainnig seeds. 
    Setting is on Librispeech 10min without LM decoding.}
    \label{tab:random_seed}
    \begin{center}
    \vspace{-3mm}
    \scalebox{1.}{
    \begin{tabular}{l|c|c}
        \toprule
        Method & seed & test-clean/test-other \\
        \midrule 
        Full {\tt wav2vec2-base} & 2447 & 49.3/53.2 \\ 
        \cdashlinelr{1-3} 
        \multirow{9}{*}{{\tt wav2vec2-base} + 10\% {\tt PARP}}  & 2447 & 38.04/44.33 \\ 
         & 0    & 37.01/43.02  \\
         & 1    & 37.82/43.66 \\
         & 2    & 37.59/43.55 \\
         & 3    & 37.57/43.29 \\
         & 5    & 37.48/44.10 \\
         & 6    & 37.87/43.55 \\
         & 7    & 37.65/43.53 \\
         & 8    & 38.22/43.91  \\
    \bottomrule
    \end{tabular}}
    \end{center}
\end{table*}

\begin{table*}[!h]
    \small
    \caption{Decode pruned {\tt wav2vec-base} with official 4-gram/transformer LMs.
    Setting is on Librispeech 10min.}
    \label{tab:parp_lm}
    \begin{center}
    \vspace{-3mm}
    \scalebox{1.}{
    \begin{tabular}{l|c|c|c}
        \toprule
        Method & decoding algorithm & beam size & test-clean/test-other \\
        \midrule 
        \multirow{3}{*}{Full {\tt wav2vec2-base}} & viterbi (no LM) & & 49.3/53.2 \\ 
        & 4-gram LM & 5 & 27.82/32.02 \\ 
        & transformer LM & 5 & 27.16/32.68 \\ 
        \cdashlinelr{1-4} 
        \multirow{3}{*}{{\tt wav2vec2-base} + 10\% {\tt PARP} averaged}  & viterbi (no LM) & & 37.69/43.66 \\ 
        & 4-gram LM  & 5 & 25.17/32.13 \\
        & transformer LM  & 5 & 25.45/32.46 \\
    \bottomrule
    \end{tabular}}
    \end{center}
\end{table*}

\newpage
\section{{\tt wav2vec2} Cross-Task Mask Transfer on SUPERB}
\label{app:cross_task_superb}

We extend experiments in Section~\ref{subsec:mask_transfer} to downstream tasks other than ASR, i.e. extend the transferability of pruning masks across speech tasks. 
We selected three drastically different target tasks from SUPERB~\cite{yang2021superb}: Phone Recognition with 10h Librispeech data (in PER), Automatic Speaker Verification on VoxCeleb (in EER), and Slot Filling on audio SNIPS (in slot type $F_1$/slot value CER). 
PER/EER/CER are lower the better, and $F_1$ is higher the better.
The experiment procedure~\footnote{All experiments are run with SUPERB's toolkit \url{https://github.com/s3prl/s3prl}.} is as follows: 
\begin{enumerate}
    \item Finetune {\tt wav2vec2} for a source task in SUPERB. 
    \item Prune the finetuned model and obtain an {\tt OMP} mask for each task. 
    \item Apply the {\tt OMP} mask at {\tt wav2vec2} pre-trained initializations and finetune for a target task in SUPERB with {\tt PARP}.
\end{enumerate}

Table~\ref{tab:superb_task_transfer} is the {\tt wav2vec-base} cross-task transfer result in SUPERB.
We did learning rate grid search over $\{1.0\times10^{-3}, 1.0\times10^{-4}, 1.0\times10^{-5}, 2.0\times10^{-5}, 3.0\times10^{-5}, 1.0\times10^{-6}, 1.0\times10^{-7}\}$, and presented the best number. 
Note that different from SUPERB's default setup, we make the upstream {\tt wav2vec2} jointly finetunable for {\tt PARP}. 
Therefore, the hyper-parameters for each task finetuning are not optimized, and the results here have to be taken with a grain of salt. 
\begin{table*}[!h]
    \small
    \caption{Cross-task mask transfer for {\tt wav2vec-base} at 50\% sparsity.}
    \label{tab:superb_task_transfer}
    \begin{center}
    \vspace{-3mm}
    \scalebox{0.9}{
    \begin{tabular}{l|c|c|c}
        \toprule
        \multirow{2}{*}{Source task} & Target task 1: & Target task 2: & Target task 3: Slot Filling \\
        & Phone Recog (in PER) & Speaker Verification (in EER) & (in slot type $F_1$/slot value CER) \\
        \midrule 
        10h Librispeech ASR & 0.0567 & 0.1230 & 0.7635/0.4432 \\
        1h Librispeech ASR & 0.0567 & 0.1316 & 0.7563/0.4470 \\
        10min Librispeech ASR & 0.0576 & 0.1399 & 0.7452/0.4596 \\
        \cdashlinelr{1-4} 
        10h Phone Recog & 0.0471 & 0.1392 & 0.7575/0.4468 \\
        1h Phone Recog & 0.0483 & 0.1138 & 0.7508/0.4537 \\
        10min Phone Recog & 0.0535 & 0.1224 & 0.7519/0.4596 \\
        \cdashlinelr{1-4} 
        Intent Classification & 0.0617 & 0.1165 & 0.7490/0.4621 \\
        Slot Filling & 0.0601 & 0.1097 & 0.7708/0.4327 \\
        \cdashlinelr{1-4} 
        Keyword Spotting & 0.0656 & 0.1303 & 0.7490/0.4661 \\
        \cdashlinelr{1-4} 
        Speaker Verification & 0.0790 & 0.1131 & 0.7497/0.4654 \\
        Speaker ID & 0.0677 & 0.1271 & 0.7581/0.4559 \\
        Speaker Diarization & 0.0756 & 0.1104 & 0.7449/0.4623 \\
    \bottomrule
    \end{tabular}}
    \end{center}
\end{table*}

We first see that indeed the more similar source and target tasks are, the performance are better. 
For instance, source subnetwork obtained from speaker related task perform better than those obtained from ASR/keyword spotting on speaker verification. 
For another, source subnetwork obtained from ASR/phone recognition perform better than those obtained from speaker related task on phone recognition. 
We do note that the numbers are not off by too much, and the differences could be potentially reduced via hyper-parameter tuning. 
This pilot study also suggests that subnetworks transferability depends on task similarity. 
Lastly, this experiment does not contradict our main setting, as we were primarily interested in cross-lingual transferability of subnetworks in Section~\ref{subsec:mask_transfer}. 

\newpage
\section{Does Observation~\ref{observation:similarity} generalize across Pre-Training Objectives?}
\label{app:observation1_across_pretrain_obj}

Observation~\ref{observation:similarity} states that:
\begin{tcolorbox}
\vspace{-2mm}
\textit{For any sparsity, any amount of finetuning supervision, any pre-training model scale, and any downstream spoken languages, the non-zero ASR pruning masks obtained from task-agnostic subnetwork discovery has high {\tt IOU}s with those obtained from task-aware subnetwork discovery.}
\vspace{-2mm}
\end{tcolorbox}
We provide analysis on whether Observation~\ref{observation:similarity} holds \textit{across} pre-training objectives, i.e. does pruning masks from {\tt wav2vec2} have high similarity with those from {\tt hubert}~\cite{hsu2021hubert}? 
The setup follows that of Section~\ref{app:cross_task_superb} and is based on the downstream tasks in SUPERB\footnote{For this set of experiments, we used the same optimization method (Adam with constant $1.0\times 10^{-5}$ learning rate) for finetuning {\tt wav2vec-base} and {\tt hubert-base}.}: 
\begin{enumerate}
    \item Finetune {\tt wav2vec2} for all tasks in SUPERB. 
    \item Prune the finetuned models and obtain an {\tt OMP} mask for each task. 
    \item Finetune {\tt hubert} for all tasks in SUPERB
    \item Prune the finetuned models and obtain an {\tt OMP} mask for each task. 
    \item For each task in SUPERB and at a fixed sparsity, calculate the mask {\tt IOU} between {\tt wav2vec2} and {\tt hubert}.
\end{enumerate}

Table~\ref{tab:mask_overlap_between_wav2vec2_and_hubert} is the mask {\tt IOU}s at 50\% sparsity between {\tt wav2vec-base} and {\tt hubert-base} on tasks in SUPERB.
The table indicates that while Observation~\ref{observation:similarity} holds separately for {\tt wav2vec2} (contrastive pre-training) and {\tt hubert} (mask-predict pre-training), it does not generalize across pre-training method give the close to random mask {\tt IOU}s (c.f. last row of Table~\ref{tab:mask_overlap_between_wav2vec2_and_hubert}).
Therefore,
\begin{tcolorbox}
\vspace{-2mm}
\textit{Observation~\ref{observation:similarity} holds true conditioned on the same speech SSL pre-training objective.}
\vspace{-2mm}
\end{tcolorbox}

\begin{table*}[!h]
    \small
    \caption{Mask {\tt IOU} between {\tt wav2vec-base} and {\tt hubert-base} at 50\% sparsity.}
    \label{tab:mask_overlap_between_wav2vec2_and_hubert}
    \begin{center}
    \vspace{-3mm}
    \scalebox{1.}{
    \begin{tabular}{l|c}
        \toprule
        target task & mask {\tt IOU} between {\tt wav2vec-base} and {\tt hubert-base} \\
        \midrule 
        10h Librispeech ASR & 0.3472 \\
        1h Librispeech ASR & 0.3473 \\
        10min Librispeech ASR & 0.3473 \\
        \cdashlinelr{1-2} 
        10h Phone Recog & 0.3473 \\
        1h Phone Recog & 0.3473 \\
        10min Phone Recog & 0.3473 \\
        \cdashlinelr{1-2} 
        Intent Classification & 0.3473 \\
        Slot Filling & 0.3472 \\
        \cdashlinelr{1-2} 
        Keyword Spotting & 0.3473 \\
        \cdashlinelr{1-2} 
        Speaker Verification & 0.3473 \\
        Speaker ID & 0.3473 \\
        Speaker Diarization & 0.3472 \\
        \cdashlinelr{1-2} 
        Random Pruning & 0.3473 \\
    \bottomrule
    \end{tabular}}
    \end{center}
\end{table*}

This finding is perhaps not so surprising, see prior work on similarity analysis between contextualized speech~\cite{chung2021similarity} and word~\cite{wu2020similarity} representations. 
They suggest that different pre-trained models’ contextualized representations have low similarities, e.g. BERT v.s. XLNet. 
We stress that this does not invalidate {\tt PARP}. 
As long as Observation~\ref{observation:similarity} holds, {\tt PARP}’s step 2 should make learnable adjustments to the initial mask given the high overlaps between pruning masks.

\newpage
\section{Pruned Weights Localization Across Layers}
\label{app:weight_localization}

The wav2vec series~\cite{baevski2020wav2vec,baevski2021unsupervised,hsu2021hubert,hsu2021hubertlarge} is known to have more valuable contextualized representations towards the middle of the network for downstream ASR. 
We examine whether previous observations holds true for pruning, that weights in middle layers are pruned less. To understand such a phenomenon, we calculated the distributions of the pruned weights/neurons across each layer, and an example is shown in Table~\ref{tab:specific_sparsity_distribution_example}. 

\begin{table*}[!h]
    \small
    \caption{{\tt wav2vec-base} finetuned for Spanish (\textbf{H2L} setting) pruned at 50\% sparsity with {\tt OMP}.}
    \label{tab:specific_sparsity_distribution_example}
    \begin{center}
    \vspace{-3mm}
    \scalebox{0.9}{
    \begin{tabular}{l|c|c|c|c|c|c|c|c|c|c|c|c}
        \toprule
        layer & 1 & 2 & 3 & 4 & 5 & 6 & 7 & 8 & 9 & 10 & 11 & 12 \\
        \midrule 
        sparsity (\%) & 53.52 &	52.45 &	49.24 &	47.90 &	46.51 &	46.84 &	45.97 &	45.58 &	45.96 &	47.96 &	52.54 &	65.53 \\
    \bottomrule
    \end{tabular}}
    \end{center}
\end{table*}

Table~\ref{tab:specific_sparsity_distribution_example} shows that indeed bottom and higher layers of {\tt wav2vec2-base} are pruned more, while the middle layers are pruned less. 
We observe similar pruned weight distributions across spoken languages (10 languages) and sparsities (10\%, 20\%, 30\%, $\dots$, 90\%). 
See the rest of the sparsity distribution in the Figures below. 
This analysis suggests that regardless of spoken languages, intermediate layers’ neurons are more valuable than lower and higher-level layers, manifested by the layer's sparsity ratio. 

    \begin{figure*} [!h]
    \includegraphics[width=\linewidth]{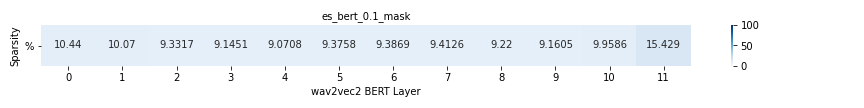}
    \centering
    \caption{Sparsity over layers for {\tt wav2vec-base} finetuned for Spanish \textit{es} at 10\% sparsity.}
    \label{fig:weight_dist_es_0.1}
    \end{figure*}
    \begin{figure*} [!h]
    \includegraphics[width=\linewidth]{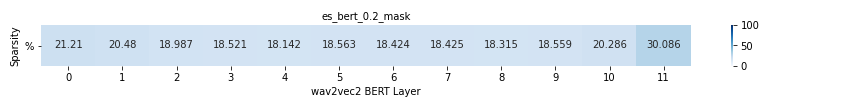}
    \centering
    \caption{Sparsity over layers for {\tt wav2vec-base} finetuned for Spanish \textit{es} at 20\% sparsity.}
    \label{fig:weight_dist_es_0.2}
    \end{figure*}
    \begin{figure*} [!h]
    \includegraphics[width=\linewidth]{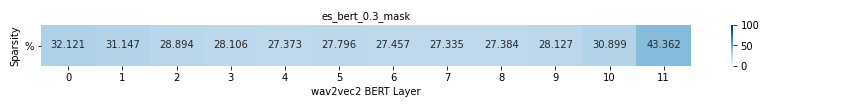}
    \centering
    \caption{Sparsity over layers for {\tt wav2vec-base} finetuned for Spanish \textit{es} at 30\% sparsity.}
    \label{fig:weight_dist_es_0.3}
    \end{figure*}
    \begin{figure*} [!h]
    \includegraphics[width=\linewidth]{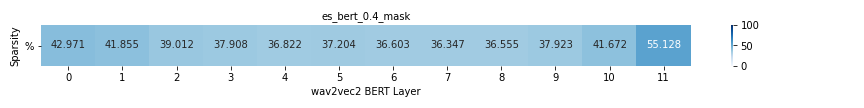}
    \centering
    \caption{Sparsity over layers for {\tt wav2vec-base} finetuned for Spanish \textit{es} at 40\% sparsity.}
    \label{fig:weight_dist_es_0.4}
    \end{figure*}
    \begin{figure*} [!h]
    \includegraphics[width=\linewidth]{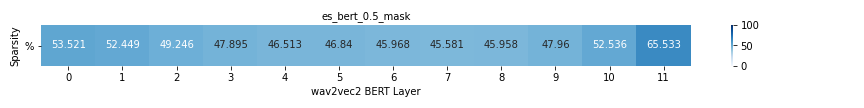}
    \centering
    \caption{Sparsity over layers for {\tt wav2vec-base} finetuned for Spanish \textit{es} at 50\% sparsity.}
    \label{fig:weight_dist_es_0.5}
    \end{figure*}
    \begin{figure*} [!h]
    \includegraphics[width=\linewidth]{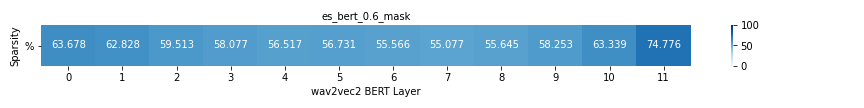}
    \centering
    \caption{Sparsity over layers for {\tt wav2vec-base} finetuned for Spanish \textit{es} at 60\% sparsity.}
    \label{fig:weight_dist_es_0.6}
    \end{figure*}
    \begin{figure*} [!h]
    \includegraphics[width=\linewidth]{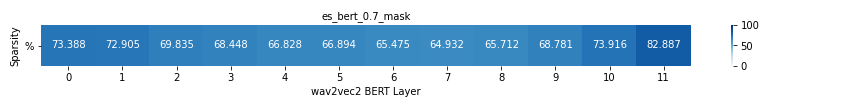}
    \centering
    \caption{Sparsity over layers for {\tt wav2vec-base} finetuned for Spanish \textit{es} at 70\% sparsity.}
    \label{fig:weight_dist_es_0.7}
    \end{figure*}
    \begin{figure*} [!h]
    \includegraphics[width=\linewidth]{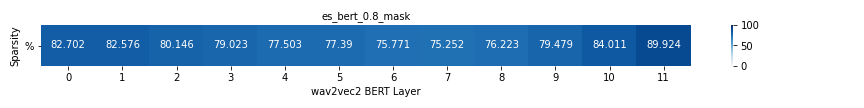}
    \centering
    \caption{Sparsity over layers for {\tt wav2vec-base} finetuned for Spanish \textit{es} at 80\% sparsity.}
    \label{fig:weight_dist_es_0.8}
    \end{figure*}
    \begin{figure*} [!h]
    \includegraphics[width=\linewidth]{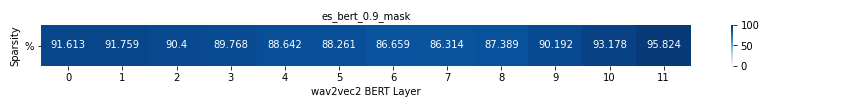}
    \centering
    \caption{Sparsity over layers for {\tt wav2vec-base} finetuned for Spanish \textit{es} at 90\% sparsity.}
    \label{fig:weight_dist_es_0.9}
    \end{figure*}

    \begin{figure*} [!h]
    \includegraphics[width=\linewidth]{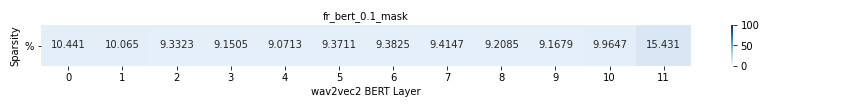}
    \centering
    \caption{Sparsity over layers for {\tt wav2vec-base} finetuned for French \textit{fr} at 10\% sparsity.}
    \label{fig:weight_dist_fr_0.1}
    \end{figure*}
    \begin{figure*} [!h]
    \includegraphics[width=\linewidth]{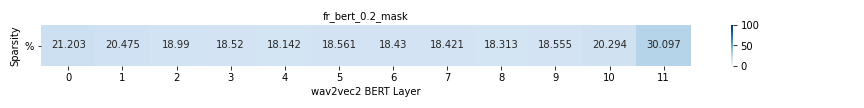}
    \centering
    \caption{Sparsity over layers for {\tt wav2vec-base} finetuned for French \textit{fr} at 20\% sparsity.}
    \label{fig:weight_dist_fr_0.2}
    \end{figure*}
    \begin{figure*} [!h]
    \includegraphics[width=\linewidth]{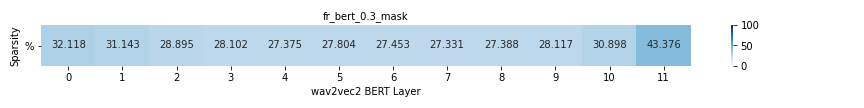}
    \centering
    \caption{Sparsity over layers for {\tt wav2vec-base} finetuned for French \textit{fr} at 30\% sparsity.}
    \label{fig:weight_dist_fr_0.3}
    \end{figure*}
    \begin{figure*} [!h]
    \includegraphics[width=\linewidth]{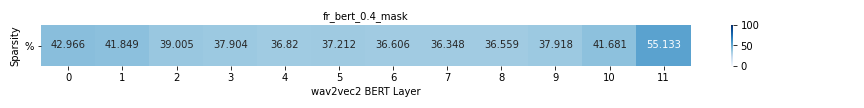}
    \centering
    \caption{Sparsity over layers for {\tt wav2vec-base} finetuned for French \textit{fr} at 40\% sparsity.}
    \label{fig:weight_dist_fr_0.4}
    \end{figure*}
    \begin{figure*} [!h]
    \includegraphics[width=\linewidth]{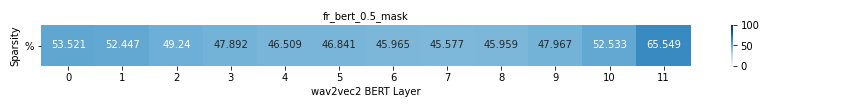}
    \centering
    \caption{Sparsity over layers for {\tt wav2vec-base} finetuned for French \textit{fr} at 50\% sparsity.}
    \label{fig:weight_dist_fr_0.5}
    \end{figure*}
    \begin{figure*} [!h]
    \includegraphics[width=\linewidth]{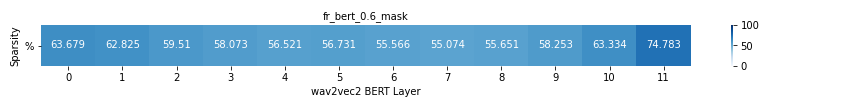}
    \centering
    \caption{Sparsity over layers for {\tt wav2vec-base} finetuned for French \textit{fr} at 60\% sparsity.}
    \label{fig:weight_dist_fr_0.6}
    \end{figure*}
    \begin{figure*} [!h]
    \includegraphics[width=\linewidth]{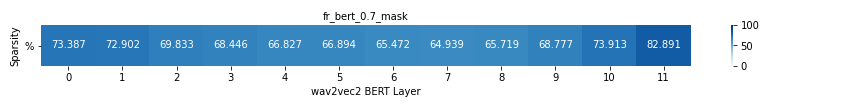}
    \centering
    \caption{Sparsity over layers for {\tt wav2vec-base} finetuned for French \textit{fr} at 70\% sparsity.}
    \label{fig:weight_dist_fr_0.7}
    \end{figure*}
    \begin{figure*} [!h]
    \includegraphics[width=\linewidth]{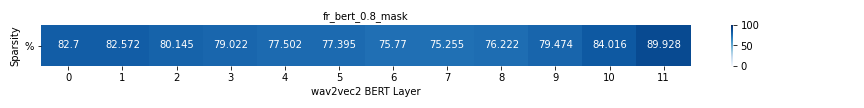}
    \centering
    \caption{Sparsity over layers for {\tt wav2vec-base} finetuned for French \textit{fr} at 80\% sparsity.}
    \label{fig:weight_dist_fr_0.8}
    \end{figure*}
    \begin{figure*} [!h]
    \includegraphics[width=\linewidth]{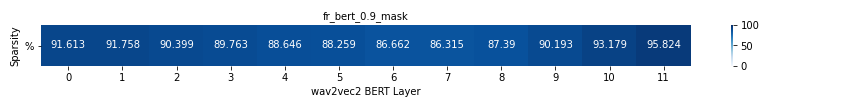}
    \centering
    \caption{Sparsity over layers for {\tt wav2vec-base} finetuned for French \textit{fr} at 90\% sparsity.}
    \label{fig:weight_dist_fr_0.9}
    \end{figure*}

    \begin{figure*} [!h]
    \includegraphics[width=\linewidth]{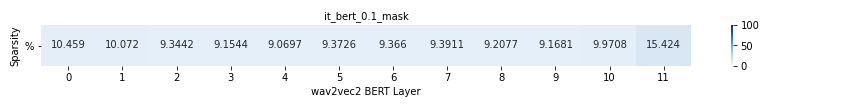}
    \centering
    \caption{Sparsity over layers for {\tt wav2vec-base} finetuned for Italian \textit{it} at 10\% sparsity.}
    \label{fig:weight_dist_it_0.1}
    \end{figure*}
    \begin{figure*} [!h]
    \includegraphics[width=\linewidth]{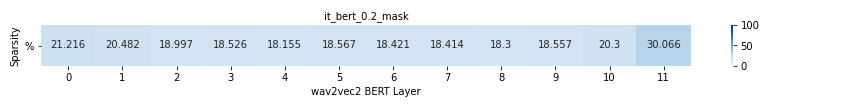}
    \centering
    \caption{Sparsity over layers for {\tt wav2vec-base} finetuned for Italian \textit{it} at 20\% sparsity.}
    \label{fig:weight_dist_it_0.2}
    \end{figure*}
    \begin{figure*} [!h]
    \includegraphics[width=\linewidth]{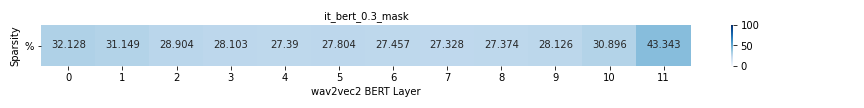}
    \centering
    \caption{Sparsity over layers for {\tt wav2vec-base} finetuned for Italian \textit{it} at 30\% sparsity.}
    \label{fig:weight_dist_it_0.3}
    \end{figure*}
    \begin{figure*} [!h]
    \includegraphics[width=\linewidth]{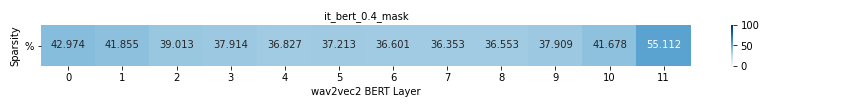}
    \centering
    \caption{Sparsity over layers for {\tt wav2vec-base} finetuned for Italian \textit{it} at 40\% sparsity.}
    \label{fig:weight_dist_it_0.4}
    \end{figure*}
    \begin{figure*} [!h]
    \includegraphics[width=\linewidth]{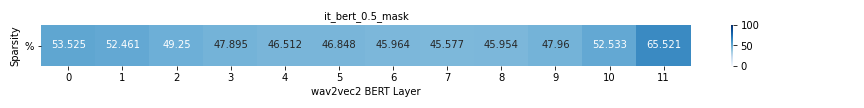}
    \centering
    \caption{Sparsity over layers for {\tt wav2vec-base} finetuned for Italian \textit{it} at 50\% sparsity.}
    \label{fig:weight_dist_it_0.5}
    \end{figure*}
    \begin{figure*} [!h]
    \includegraphics[width=\linewidth]{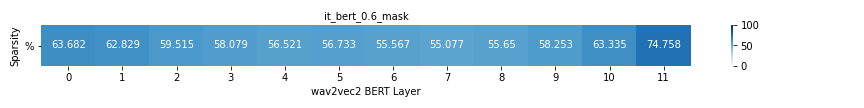}
    \centering
    \caption{Sparsity over layers for {\tt wav2vec-base} finetuned for Italian \textit{it} at 60\% sparsity.}
    \label{fig:weight_dist_it_0.6}
    \end{figure*}
    \begin{figure*} [!h]
    \includegraphics[width=\linewidth]{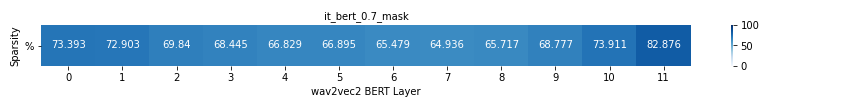}
    \centering
    \caption{Sparsity over layers for {\tt wav2vec-base} finetuned for Italian \textit{it} at 70\% sparsity.}
    \label{fig:weight_dist_it_0.7}
    \end{figure*}
    \begin{figure*} [!h]
    \includegraphics[width=\linewidth]{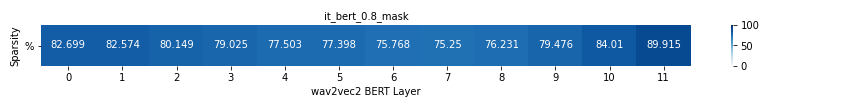}
    \centering
    \caption{Sparsity over layers for {\tt wav2vec-base} finetuned for Italian \textit{it} at 80\% sparsity.}
    \label{fig:weight_dist_it_0.8}
    \end{figure*}
    \begin{figure*} [!h]
    \includegraphics[width=\linewidth]{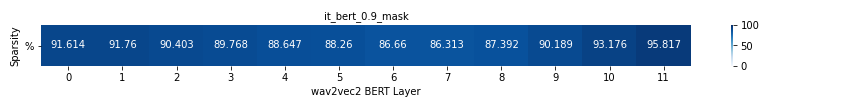}
    \centering
    \caption{Sparsity over layers for {\tt wav2vec-base} finetuned for Italian \textit{it} at 90\% sparsity.}
    \label{fig:weight_dist_it_0.9}
    \end{figure*}

    \begin{figure*} [!h]
    \includegraphics[width=\linewidth]{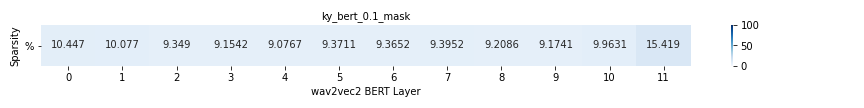}
    \centering
    \caption{Sparsity over layers for {\tt wav2vec-base} finetuned for Kyrgyz \textit{ky} at 10\% sparsity.}
    \label{fig:weight_dist_ky_0.1}
    \end{figure*}
    \begin{figure*} [!h]
    \includegraphics[width=\linewidth]{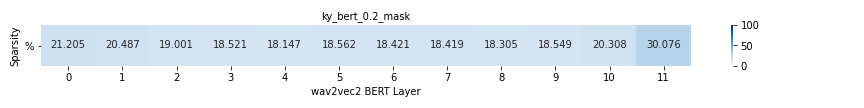}
    \centering
    \caption{Sparsity over layers for {\tt wav2vec-base} finetuned for Kyrgyz \textit{ky} at 20\% sparsity.}
    \label{fig:weight_dist_ky_0.2}
    \end{figure*}
    \begin{figure*} [!h]
    \includegraphics[width=\linewidth]{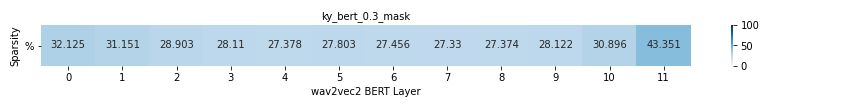}
    \centering
    \caption{Sparsity over layers for {\tt wav2vec-base} finetuned for Kyrgyz \textit{ky} at 30\% sparsity.}
    \label{fig:weight_dist_ky_0.3}
    \end{figure*}
    \begin{figure*} [!h]
    \includegraphics[width=\linewidth]{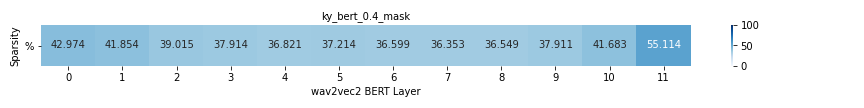}
    \centering
    \caption{Sparsity over layers for {\tt wav2vec-base} finetuned for Kyrgyz \textit{ky} at 40\% sparsity.}
    \label{fig:weight_dist_ky_0.4}
    \end{figure*}
    \begin{figure*} [!h]
    \includegraphics[width=\linewidth]{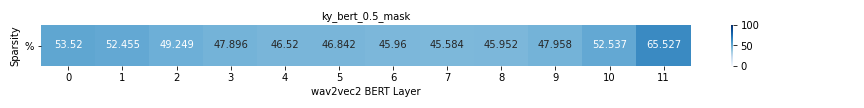}
    \centering
    \caption{Sparsity over layers for {\tt wav2vec-base} finetuned for Kyrgyz \textit{ky} at 50\% sparsity.}
    \label{fig:weight_dist_ky_0.5}
    \end{figure*}
    \begin{figure*} [!h]
    \includegraphics[width=\linewidth]{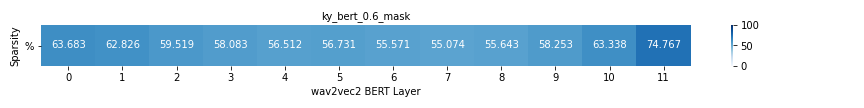}
    \centering
    \caption{Sparsity over layers for {\tt wav2vec-base} finetuned for Kyrgyz \textit{ky} at 60\% sparsity.}
    \label{fig:weight_dist_ky_0.6}
    \end{figure*}
    \begin{figure*} [!h]
    \includegraphics[width=\linewidth]{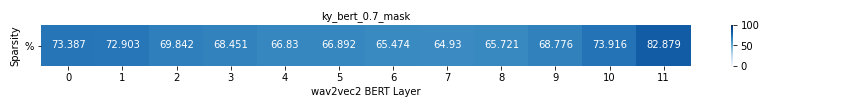}
    \centering
    \caption{Sparsity over layers for {\tt wav2vec-base} finetuned for Kyrgyz \textit{ky} at 70\% sparsity.}
    \label{fig:weight_dist_ky_0.7}
    \end{figure*}
    \begin{figure*} [!h]
    \includegraphics[width=\linewidth]{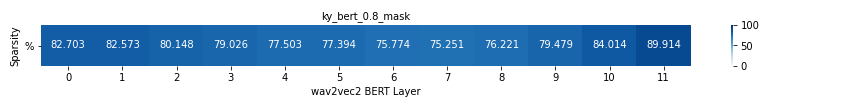}
    \centering
    \caption{Sparsity over layers for {\tt wav2vec-base} finetuned for Kyrgyz \textit{ky} at 80\% sparsity.}
    \label{fig:weight_dist_ky_0.8}
    \end{figure*}
    \begin{figure*} [!h]
    \includegraphics[width=\linewidth]{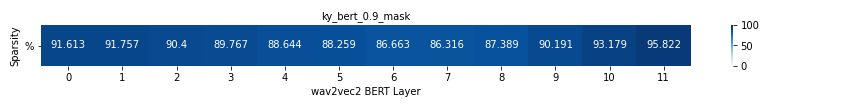}
    \centering
    \caption{Sparsity over layers for {\tt wav2vec-base} finetuned for Kyrgyz \textit{ky} at 90\% sparsity.}
    \label{fig:weight_dist_ky_0.9}
    \end{figure*}
    
    \begin{figure*} [!h]
    \includegraphics[width=\linewidth]{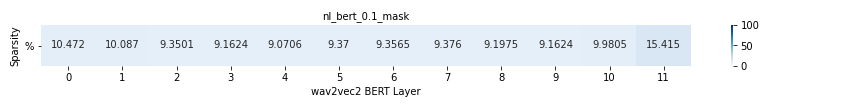}
    \centering
    \caption{Sparsity over layers for {\tt wav2vec-base} finetuned for Dutch \textit{nl} at 10\% sparsity.}
    \label{fig:weight_dist_nl_0.1}
    \end{figure*}
    \begin{figure*} [!h]
    \includegraphics[width=\linewidth]{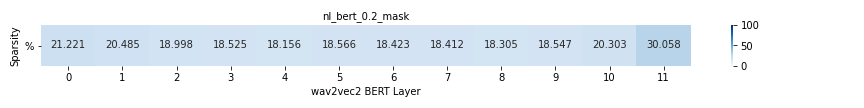}
    \centering
    \caption{Sparsity over layers for {\tt wav2vec-base} finetuned for Dutch \textit{nl} at 20\% sparsity.}
    \label{fig:weight_dist_nl_0.2}
    \end{figure*}
    \begin{figure*} [!h]
    \includegraphics[width=\linewidth]{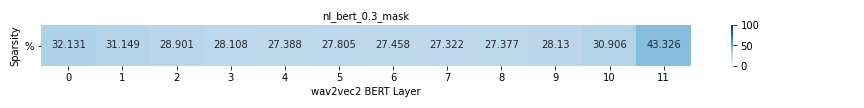}
    \centering
    \caption{Sparsity over layers for {\tt wav2vec-base} finetuned for Dutch \textit{nl} at 30\% sparsity.}
    \label{fig:weight_dist_nl_0.3}
    \end{figure*}
    \begin{figure*} [!h]
    \includegraphics[width=\linewidth]{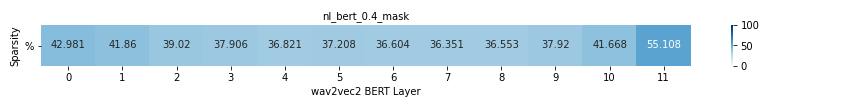}
    \centering
    \caption{Sparsity over layers for {\tt wav2vec-base} finetuned for Dutch \textit{nl} at 40\% sparsity.}
    \label{fig:weight_dist_nl_0.4}
    \end{figure*}
    \begin{figure*} [!h]
    \includegraphics[width=\linewidth]{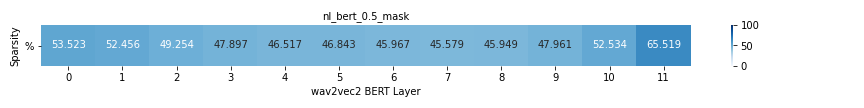}
    \centering
    \caption{Sparsity over layers for {\tt wav2vec-base} finetuned for Dutch \textit{nl} at 50\% sparsity.}
    \label{fig:weight_dist_nl_0.5}
    \end{figure*}
    \begin{figure*} [!h]
    \includegraphics[width=\linewidth]{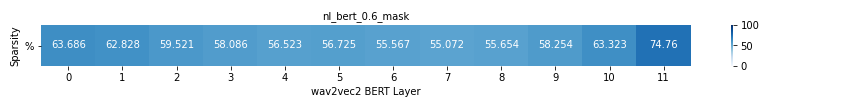}
    \centering
    \caption{Sparsity over layers for {\tt wav2vec-base} finetuned for Dutch \textit{nl} at 60\% sparsity.}
    \label{fig:weight_dist_nl_0.6}
    \end{figure*}
    \begin{figure*} [!h]
    \includegraphics[width=\linewidth]{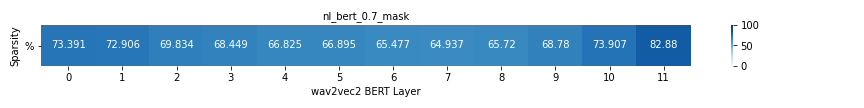}
    \centering
    \caption{Sparsity over layers for {\tt wav2vec-base} finetuned for Dutch \textit{nl} at 70\% sparsity.}
    \label{fig:weight_dist_nl_0.7}
    \end{figure*}
    \begin{figure*} [!h]
    \includegraphics[width=\linewidth]{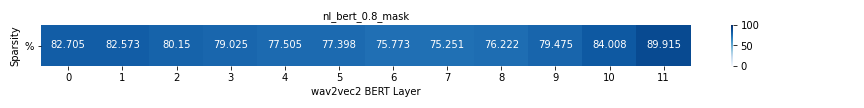}
    \centering
    \caption{Sparsity over layers for {\tt wav2vec-base} finetuned for Dutch \textit{nl} at 80\% sparsity.}
    \label{fig:weight_dist_nl_0.8}
    \end{figure*}
    \begin{figure*} [!h]
    \includegraphics[width=\linewidth]{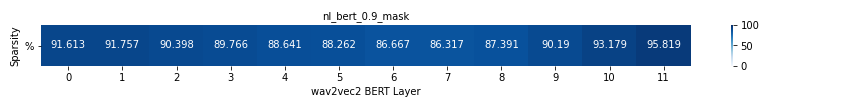}
    \centering
    \caption{Sparsity over layers for {\tt wav2vec-base} finetuned for Dutch \textit{nl} at 90\% sparsity.}
    \label{fig:weight_dist_nl_0.9}
    \end{figure*}
    
    \begin{figure*} [!h]
    \includegraphics[width=\linewidth]{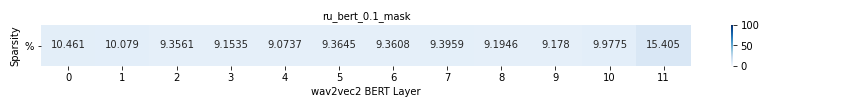}
    \centering
    \caption{Sparsity over layers for {\tt wav2vec-base} finetuned for Russian \textit{ru} at 10\% sparsity.}
    \label{fig:weight_dist_ru_0.1}
    \end{figure*}
    \begin{figure*} [!h]
    \includegraphics[width=\linewidth]{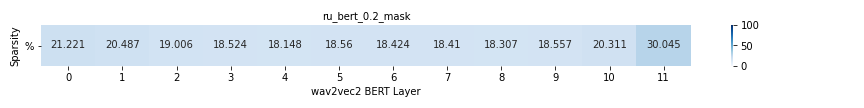}
    \centering
    \caption{Sparsity over layers for {\tt wav2vec-base} finetuned for Russian \textit{ru} at 20\% sparsity.}
    \label{fig:weight_dist_ru_0.2}
    \end{figure*}
    \begin{figure*} [!h]
    \includegraphics[width=\linewidth]{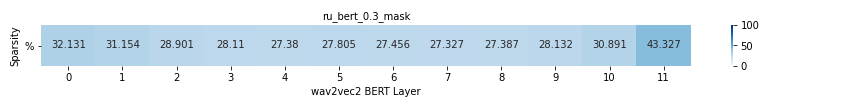}
    \centering
    \caption{Sparsity over layers for {\tt wav2vec-base} finetuned for Russian \textit{ru} at 30\% sparsity.}
    \label{fig:weight_dist_ru_0.3}
    \end{figure*}
    \begin{figure*} [!h]
    \includegraphics[width=\linewidth]{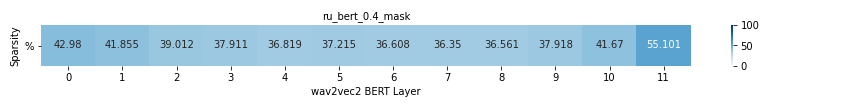}
    \centering
    \caption{Sparsity over layers for {\tt wav2vec-base} finetuned for Russian \textit{ru} at 40\% sparsity.}
    \label{fig:weight_dist_ru_0.4}
    \end{figure*}
    \begin{figure*} [!h]
    \includegraphics[width=\linewidth]{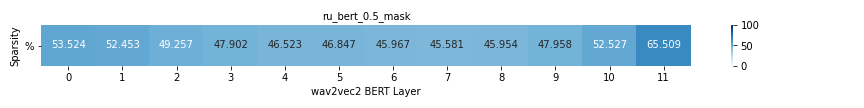}
    \centering
    \caption{Sparsity over layers for {\tt wav2vec-base} finetuned for Russian \textit{ru} at 50\% sparsity.}
    \label{fig:weight_dist_ru_0.5}
    \end{figure*}
    \begin{figure*} [!h]
    \includegraphics[width=\linewidth]{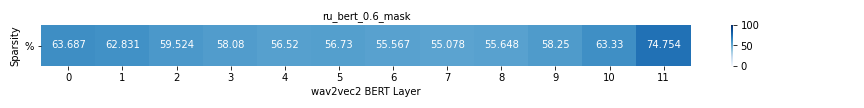}
    \centering
    \caption{Sparsity over layers for {\tt wav2vec-base} finetuned for Russian \textit{ru} at 60\% sparsity.}
    \label{fig:weight_dist_ru_0.6}
    \end{figure*}
    \begin{figure*} [!h]
    \includegraphics[width=\linewidth]{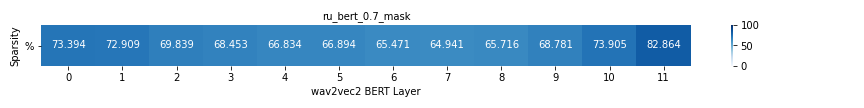}
    \centering
    \caption{Sparsity over layers for {\tt wav2vec-base} finetuned for Russian \textit{ru} at 70\% sparsity.}
    \label{fig:weight_dist_ru_0.7}
    \end{figure*}
    \begin{figure*} [!h]
    \includegraphics[width=\linewidth]{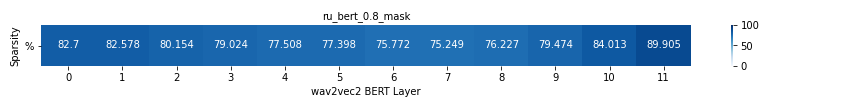}
    \centering
    \caption{Sparsity over layers for {\tt wav2vec-base} finetuned for Russian \textit{ru} at 80\% sparsity.}
    \label{fig:weight_dist_ru_0.8}
    \end{figure*}
    \begin{figure*} [!h]
    \includegraphics[width=\linewidth]{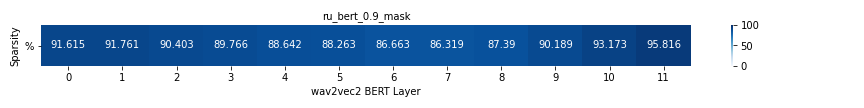}
    \centering
    \caption{Sparsity over layers for {\tt wav2vec-base} finetuned for Russian \textit{ru} at 90\% sparsity.}
    \label{fig:weight_dist_ru_0.9}
    \end{figure*}
    
    \begin{figure*} [!h]
    \includegraphics[width=\linewidth]{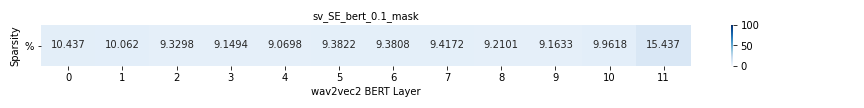}
    \centering
    \caption{Sparsity over layers for {\tt wav2vec-base} finetuned for Swedish \textit{sv-SE} at 10\% sparsity.}
    \label{fig:weight_dist_sv_SE_0.1}
    \end{figure*}
    \begin{figure*} [!h]
    \includegraphics[width=\linewidth]{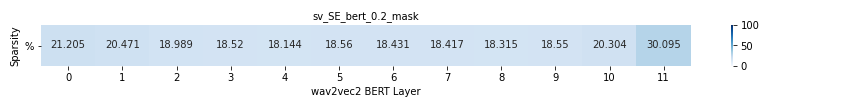}
    \centering
    \caption{Sparsity over layers for {\tt wav2vec-base} finetuned for Swedish \textit{sv-SE} at 20\% sparsity.}
    \label{fig:weight_dist_sv_SE_0.2}
    \end{figure*}
    \begin{figure*} [!h]
    \includegraphics[width=\linewidth]{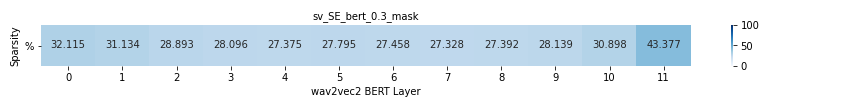}
    \centering
    \caption{Sparsity over layers for {\tt wav2vec-base} finetuned for Swedish \textit{sv-SE} at 30\% sparsity.}
    \label{fig:weight_dist_sv_SE_0.3}
    \end{figure*}
    \begin{figure*} [!h]
    \includegraphics[width=\linewidth]{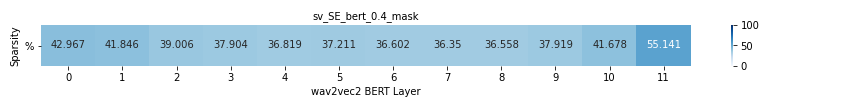}
    \centering
    \caption{Sparsity over layers for {\tt wav2vec-base} finetuned for Swedish \textit{sv-SE} at 40\% sparsity.}
    \label{fig:weight_dist_sv_SE_0.4}
    \end{figure*}
    \begin{figure*} [!h]
    \includegraphics[width=\linewidth]{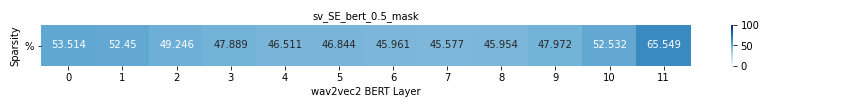}
    \centering
    \caption{Sparsity over layers for {\tt wav2vec-base} finetuned for Swedish \textit{sv-SE} at 50\% sparsity.}
    \label{fig:weight_dist_sv_SE_0.5}
    \end{figure*}
    \begin{figure*} [!h]
    \includegraphics[width=\linewidth]{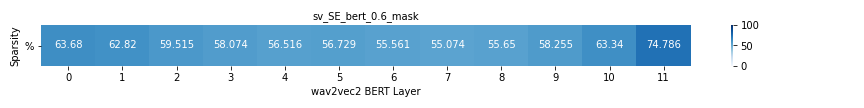}
    \centering
    \caption{Sparsity over layers for {\tt wav2vec-base} finetuned for Swedish \textit{sv-SE} at 60\% sparsity.}
    \label{fig:weight_dist_sv_SE_0.6}
    \end{figure*}
    \begin{figure*} [!h]
    \includegraphics[width=\linewidth]{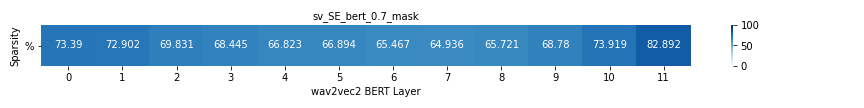}
    \centering
    \caption{Sparsity over layers for {\tt wav2vec-base} finetuned for Swedish \textit{sv-SE} at 70\% sparsity.}
    \label{fig:weight_dist_sv_SE_0.7}
    \end{figure*}
    \begin{figure*} [!h]
    \includegraphics[width=\linewidth]{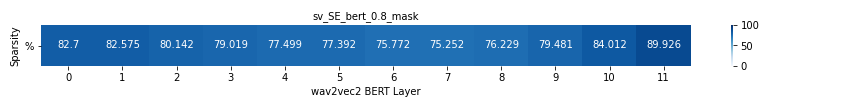}
    \centering
    \caption{Sparsity over layers for {\tt wav2vec-base} finetuned for Swedish \textit{sv-SE} at 80\% sparsity.}
    \label{fig:weight_dist_sv_SE_0.8}
    \end{figure*}
    \begin{figure*} [!h]
    \includegraphics[width=\linewidth]{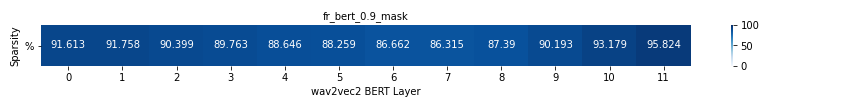}
    \centering
    \caption{Sparsity over layers for {\tt wav2vec-base} finetuned for Swedish \textit{sv-SE} at 90\% sparsity.}
    \label{fig:weight_dist_sv_SE_0.9}
    \end{figure*}
    
    \begin{figure*} [!h]
    \includegraphics[width=\linewidth]{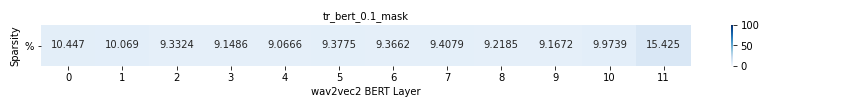}
    \centering
    \caption{Sparsity over layers for {\tt wav2vec-base} finetuned for Turkish \textit{tr} at 10\% sparsity.}
    \label{fig:weight_dist_tr_0.1}
    \end{figure*}
    \begin{figure*} [!h]
    \includegraphics[width=\linewidth]{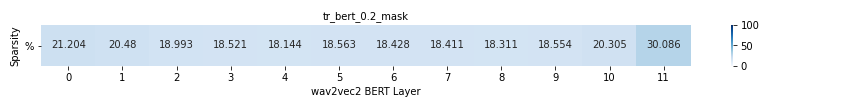}
    \centering
    \caption{Sparsity over layers for {\tt wav2vec-base} finetuned for Turkish \textit{tr} at 20\% sparsity.}
    \label{fig:weight_dist_tr_0.2}
    \end{figure*}
    \begin{figure*} [!h]
    \includegraphics[width=\linewidth]{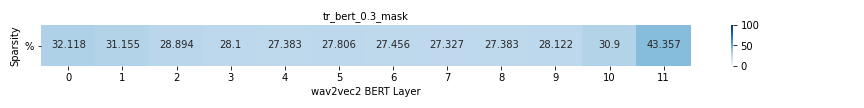}
    \centering
    \caption{Sparsity over layers for {\tt wav2vec-base} finetuned for Turkish \textit{tr} at 30\% sparsity.}
    \label{fig:weight_dist_tr_0.3}
    \end{figure*}
    \begin{figure*} [!h]
    \includegraphics[width=\linewidth]{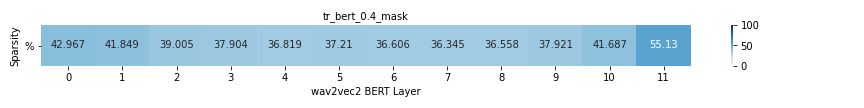}
    \centering
    \caption{Sparsity over layers for {\tt wav2vec-base} finetuned for Turkish \textit{tr} at 40\% sparsity.}
    \label{fig:weight_dist_tr_0.4}
    \end{figure*}
    \begin{figure*} [!h]
    \includegraphics[width=\linewidth]{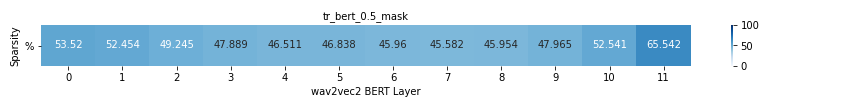}
    \centering
    \caption{Sparsity over layers for {\tt wav2vec-base} finetuned for Turkish \textit{tr} at 50\% sparsity.}
    \label{fig:weight_dist_tr_0.5}
    \end{figure*}
    \begin{figure*} [!h]
    \includegraphics[width=\linewidth]{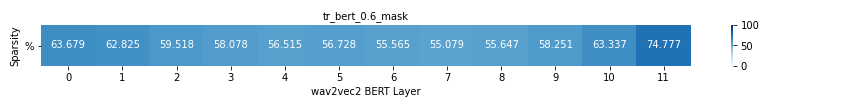}
    \centering
    \caption{Sparsity over layers for {\tt wav2vec-base} finetuned for Turkish \textit{tr} at 60\% sparsity.}
    \label{fig:weight_dist_tr_0.6}
    \end{figure*}
    \begin{figure*} [!h]
    \includegraphics[width=\linewidth]{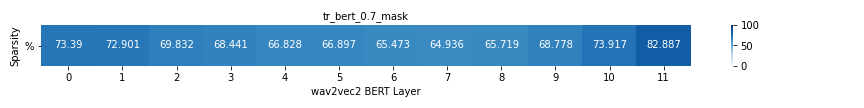}
    \centering
    \caption{Sparsity over layers for {\tt wav2vec-base} finetuned for Turkish \textit{tr} at 70\% sparsity.}
    \label{fig:weight_dist_tr_0.7}
    \end{figure*}
    \begin{figure*} [!h]
    \includegraphics[width=\linewidth]{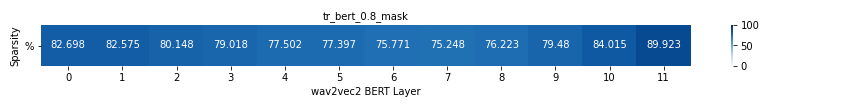}
    \centering
    \caption{Sparsity over layers for {\tt wav2vec-base} finetuned for Turkish \textit{tr} at 80\% sparsity.}
    \label{fig:weight_dist_tr_0.8}
    \end{figure*}
    \begin{figure*} [!h]
    \includegraphics[width=\linewidth]{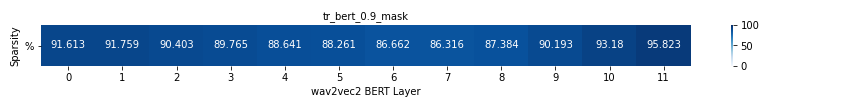}
    \centering
    \caption{Sparsity over layers for {\tt wav2vec-base} finetuned for Turkish \textit{tr} at 90\% sparsity.}
    \label{fig:weight_dist_tr_0.9}
    \end{figure*}
    
    \begin{figure*} [!h]
    \includegraphics[width=\linewidth]{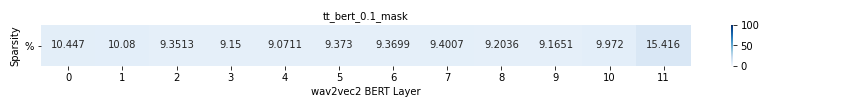}
    \centering
    \caption{Sparsity over layers for {\tt wav2vec-base} finetuned for Tatar \textit{tt} at 10\% sparsity.}
    \label{fig:weight_dist_tt_0.1}
    \end{figure*}
    \begin{figure*} [!h]
    \includegraphics[width=\linewidth]{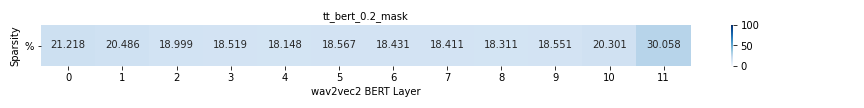}
    \centering
    \caption{Sparsity over layers for {\tt wav2vec-base} finetuned for Tatar \textit{tt} at 20\% sparsity.}
    \label{fig:weight_dist_tt_0.2}
    \end{figure*}
    \begin{figure*} [!h]
    \includegraphics[width=\linewidth]{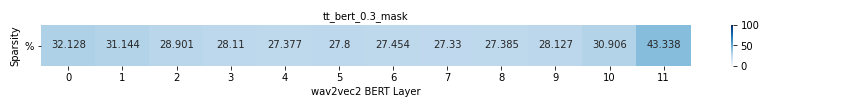}
    \centering
    \caption{Sparsity over layers for {\tt wav2vec-base} finetuned for Tatar \textit{tt} at 30\% sparsity.}
    \label{fig:weight_dist_tt_0.3}
    \end{figure*}
    \begin{figure*} [!h]
    \includegraphics[width=\linewidth]{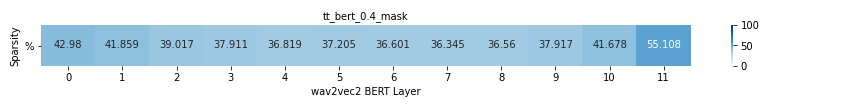}
    \centering
    \caption{Sparsity over layers for {\tt wav2vec-base} finetuned for Tatar \textit{tt} at 40\% sparsity.}
    \label{fig:weight_dist_tt_0.4}
    \end{figure*}
    \begin{figure*} [!h]
    \includegraphics[width=\linewidth]{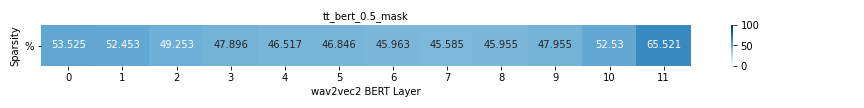}
    \centering
    \caption{Sparsity over layers for {\tt wav2vec-base} finetuned for Tatar \textit{tt} at 50\% sparsity.}
    \label{fig:weight_dist_tt_0.5}
    \end{figure*}
    \begin{figure*} [!h]
    \includegraphics[width=\linewidth]{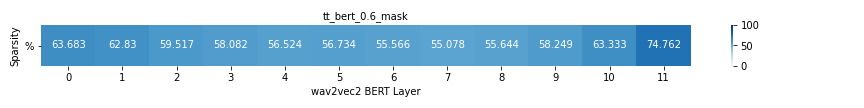}
    \centering
    \caption{Sparsity over layers for {\tt wav2vec-base} finetuned for Tatar \textit{tt} at 60\% sparsity.}
    \label{fig:weight_dist_tt_0.6}
    \end{figure*}
    \begin{figure*} [!h]
    \includegraphics[width=\linewidth]{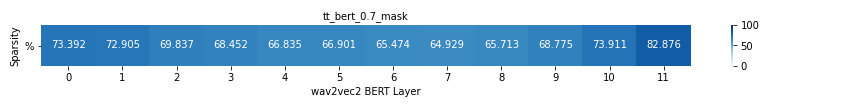}
    \centering
    \caption{Sparsity over layers for {\tt wav2vec-base} finetuned for Tatar \textit{tt} at 70\% sparsity.}
    \label{fig:weight_dist_tt_0.7}
    \end{figure*}
    \begin{figure*} [!h]
    \includegraphics[width=\linewidth]{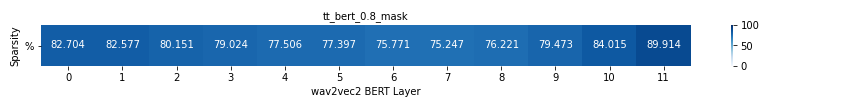}
    \centering
    \caption{Sparsity over layers for {\tt wav2vec-base} finetuned for Tatar \textit{tt} at 80\% sparsity.}
    \label{fig:weight_dist_tt_0.8}
    \end{figure*}
    \begin{figure*} [!h]
    \includegraphics[width=\linewidth]{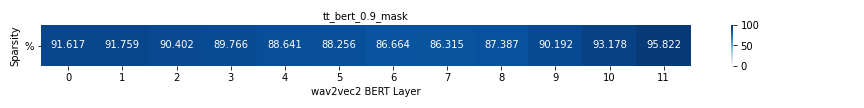}
    \centering
    \caption{Sparsity over layers for {\tt wav2vec-base} finetuned for Tatar \textit{tt} at 90\% sparsity.}
    \label{fig:weight_dist_tt_0.9}
    \end{figure*}

    \begin{figure*} [!h]
    \includegraphics[width=\linewidth]{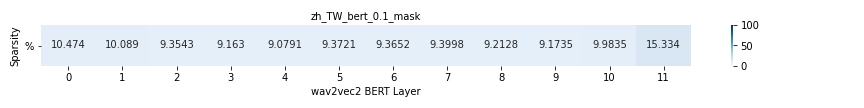}
    \centering
    \caption{Sparsity over layers for {\tt wav2vec-base} finetuned for Mandarin \textit{zh-TW} at 10\% sparsity.}
    \label{fig:weight_dist_zh_TW_0.1}
    \end{figure*}
    \begin{figure*} [!h]
    \includegraphics[width=\linewidth]{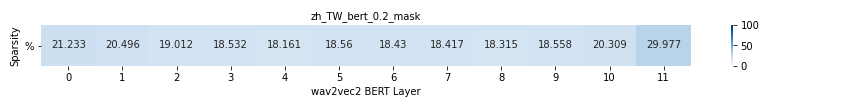}
    \centering
    \caption{Sparsity over layers for {\tt wav2vec-base} finetuned for Mandarin \textit{zh-TW} at 20\% sparsity.}
    \label{fig:weight_dist_zh_TW_0.2}
    \end{figure*}
    \begin{figure*} [!h]
    \includegraphics[width=\linewidth]{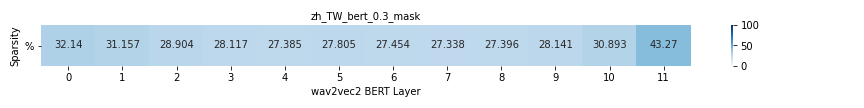}
    \centering
    \caption{Sparsity over layers for {\tt wav2vec-base} finetuned for Mandarin \textit{zh-TW} at 30\% sparsity.}
    \label{fig:weight_dist_zh_TW_0.3}
    \end{figure*}
    \begin{figure*} [!h]
    \includegraphics[width=\linewidth]{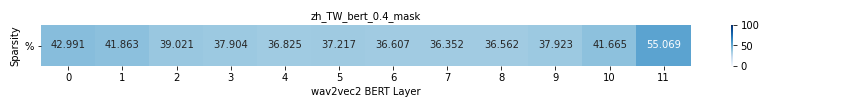}
    \centering
    \caption{Sparsity over layers for {\tt wav2vec-base} finetuned for Mandarin \textit{zh-TW} at 40\% sparsity.}
    \label{fig:weight_dist_zh_TW_0.4}
    \end{figure*}
    \begin{figure*} [!h]
    \includegraphics[width=\linewidth]{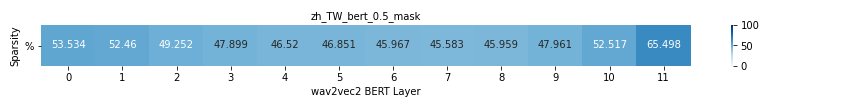}
    \centering
    \caption{Sparsity over layers for {\tt wav2vec-base} finetuned for Mandarin \textit{zh-TW} at 50\% sparsity.}
    \label{fig:weight_dist_zh_TW_0.5}
    \end{figure*}
    \begin{figure*} [!h]
    \includegraphics[width=\linewidth]{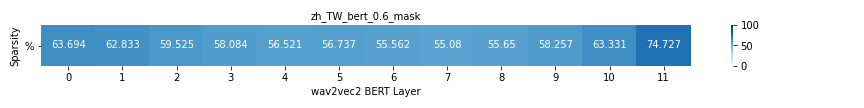}
    \centering
    \caption{Sparsity over layers for {\tt wav2vec-base} finetuned for Mandarin \textit{zh-TW} at 60\% sparsity.}
    \label{fig:weight_dist_zh_TW_0.6}
    \end{figure*}
    \begin{figure*} [!h]
    \includegraphics[width=\linewidth]{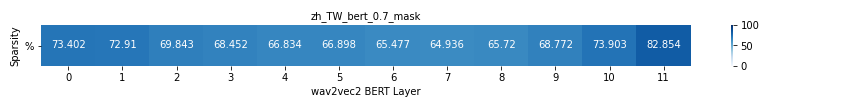}
    \centering
    \caption{Sparsity over layers for {\tt wav2vec-base} finetuned for Mandarin \textit{zh-TW} at 70\% sparsity.}
    \label{fig:weight_dist_zh_TW_0.7}
    \end{figure*}
    \begin{figure*} [!h]
    \includegraphics[width=\linewidth]{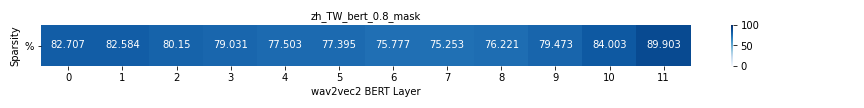}
    \centering
    \caption{Sparsity over layers for {\tt wav2vec-base} finetuned for Mandarin \textit{zh-TW} at 80\% sparsity.}
    \label{fig:weight_dist_zh_TW_0.8}
    \end{figure*}
    \begin{figure*} [!h]
    \includegraphics[width=\linewidth]{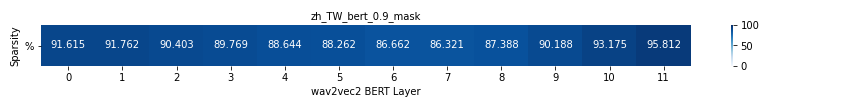}
    \centering
    \caption{Sparsity over layers for {\tt wav2vec-base} finetuned for Mandarin \textit{zh-TW} at 90\% sparsity.}
    \label{fig:weight_dist_zh_TW_0.9}
    \end{figure*}

\newpage~\newpage

\section{Experimental Limitations}
\label{app:limitations}
    Below, we list several limiting factors of our experimental designs:  
    \begin{enumerate}
        \item Experiments are on contrastive pre-trained models only. It is unclear whether the results would generalize to pre-trained models with other objectives, such as mask prediction ({\tt HuBERT}) or autoregressive prediction ({\tt APC}), etc. 
        \item Although standard, our experiments are on relatively large pre-trained models (number of parameter is 90M for {\tt wav2vec2-base} and 315M for {\tt wav2vec2-large} and {\tt xlsr}. It would be interesting to investigate if small pre-trained models can also be pruned and whether Observation~\ref{observation:similarity} holds for them. 
        \item Our {\tt wav2vec2-base} and {\tt wav2vec2-large} are both pre-trained on Librispeech 960 hours. Another lack of study is the effect of pre-training data selections -- what happens if pre-training and fine-tuning data are from different sources? 
        \item Our fine-tuning dataset (Librispeech and CommonVoice) are both read speech. Experiments on conversational (e.g. telephone) speech should be investigated.  
        \item In addition, though opposite to our motivation, it is unclear is the results hold for high-resource languages (e.g. 100h$\sim$1000h of fine-tuning data). 
        \item Our ASR experiments are based on self-supervised pre-trained models. It remains to be studied on applying {\tt PARP} to E2E ASR without self-supervised pre-training. 
        \item Lastly, we note that this study is scientific by nature. Observation~\ref{observation:similarity} emerges after our initial pilot study, and it motivates the central idea of {\tt PARP}. We will leave it to follow-up work to test whether such pruning method is effective in more realistic settings (e.g. noisy data, limited bandwidth, etc). 
        \end{enumerate}  


\end{document}